\documentclass[sigconf,screen]{acmart}

\usepackage{enumitem}
\usepackage[captionskip=-1pt]{subfig}
\usepackage{graphicx}
\usepackage{multirow}
\usepackage{makecell}
\usepackage{algorithm,algorithmic}
\usepackage{pifont}
\usepackage{cleveref}

\theoremstyle{plain}

\makeatletter
\newcommand*{\Rom}[1]{\expandafter\@slowromancap\romannumeral #1@}
\newcommand*{\rom}[1]{\romannumeral#1\relax}
\makeatother

\allowdisplaybreaks

\AtBeginDocument{%
  }

\copyrightyear{2025}
\acmYear{2025}
\setcopyright{cc}
\setcctype{by}
\acmConference[KDD '25]{Proceedings of the 31st ACM SIGKDD Conference on Knowledge Discovery and Data Mining V.2}{August 3--7, 2025}{Toronto, ON, Canada}
\acmBooktitle{Proceedings of the 31st ACM SIGKDD Conference on Knowledge Discovery and Data Mining V.2 (KDD '25), August 3--7, 2025, Toronto, ON, Canada}
\acmDOI{10.1145/3711896.3737075}
\acmISBN{979-8-4007-1454-2/2025/08}

\begin{document}

\title{Partition-wise Graph Filtering: A Unified Perspective Through the Lens of Graph Coarsening}


\author{Guoming Li}
\affiliation{%
  \institution{Mohamed bin Zayed University of Artificial Intelligence}
  \city{Abu Dhabi}
  \country{United Arab Emirates}
}
\email{paskardli@outlook.com}

\author{Jian Yang}
\affiliation{%
  \institution{University of Chinese Academy of Sciences}
  \city{Beijing}
  \country{China}
}
\email{jianyang0227@gmail.com}

\author{Yifan Chen}
\authornote{Corresponding Authors}
\affiliation{%
  \institution{Hong Kong Baptist University}
  \city{Hong Kong SAR}
  \country{China}
}
\email{yifanc@hkbu.edu.hk}


\renewcommand{\shortauthors}{Guoming Li, Jian Yang, and Yifan Chen}

\begin{abstract}
Filtering-based graph neural networks (GNNs) constitute a distinct class of GNNs that employ graph filters to handle graph-structured data, achieving notable success in various graph-related tasks. 
Conventional methods adopt a graph-wise filtering paradigm, imposing a uniform filter across all nodes, yet recent findings suggest that this rigid paradigm struggles with heterophilic graphs. 
To overcome this, recent works have introduced node-wise filtering, which assigns distinct filters to individual nodes, offering enhanced adaptability. 
However, a fundamental gap remains: a comprehensive framework unifying these two strategies is still absent, limiting theoretical insights into the filtering paradigms. 
Moreover, through the lens of \emph{Contextual Stochastic Block Model}, we reveal that a synthesis of graph-wise and node-wise filtering provides a sufficient solution for classification on graphs exhibiting both homophily and heterophily, suggesting the risk of excessive parameterization and potential overfitting with node-wise filtering. 
To address the limitations, this paper introduces {\bf C}oarsening-guided {\bf P}artition-wise {\bf F}iltering (CPF). 
CPF innovates by performing filtering on node partitions. 
The method begins with \underline{structure-aware} partition-wise filtering, which filters node partitions obtained via graph coarsening algorithms, and then performs \underline{feature-aware} partition-wise filtering, refining node embeddings via filtering on clusters produced by $k$-means clustering over features. 
In-depth analysis is conducted for each phase of CPF, showing its superiority over other paradigms. 
Finally, benchmark node classification experiments, along with a real-world graph anomaly detection application, validate CPF's efficacy and practical utility.
Code is available with the Github repository:~\url{https://github.com/vasile-paskardlgm/CPF}.
\end{abstract}

\ccsdesc[500]{Computing methodologies~Machine learning}

\keywords{Graph Filtering, Graph Coarsening, Node Classification, Heterophily}



\maketitle

\newcommand\kddavailabilityurl{https://doi.org/10.5281/zenodo.15476138}

\ifdefempty{\kddavailabilityurl}{}{
\begingroup\small\noindent\raggedright\textbf{KDD Availability Link:}\\
The source code of this paper has been made publicly available at \url{\kddavailabilityurl}.
\endgroup
}


\section{Introduction}

Graph neural networks (GNNs; \cite{comprehensivegnn,gnn-survey}) have emerged as potent tools to capture structural information from graph data, 
achieving prominent performance across numerous graph-based web applications, such as web search~\cite{websearch_1,websearch_2}, recommendation systems~\cite{recsys_1,recsys_2}, social network analysis~\cite{social_2}, anomaly detection~\cite{GAnoDet-1-CARE-GNN,GAnoDet-2-PC-GNN,GAnoDet-3-GDN}, etc.

Among GNN varieties, filtering-based GNNs stand out for their  use of graph filters to conduct filtering operations on graph data, receiving great attention due to their impressive efficacy. 
Conventional filtering-based GNNs follow a {\it graph-wise filtering} paradigm, wherein a uniform filter is applied across all nodes. 
This approach has yielded promising results in node classification tasks~\cite{GPRGNN,BernNet-GNN-narrowbandresults-1,chebnet2d,ChebNetII,OptBasisGNN,JacobiConv,decoupled-UniFilter,decoupled-PCConv,decoupled-AdaptKry,decoupled-TFGNN}. 
However, recent studies provide both theoretical and empirical evidence that this paradigm lacks flexibility and struggles with heterophilic graphs~\cite{heterophily-gnn-survey,heterophily-gnn-survey-2,heterophily-gnn-survey-3}. 
To address this, {\it node-wise filtering} has emerged as a novel paradigm that assigns distinct filters to individual nodes, yielding significant performance gains, especially in heterophilic graphs~\cite{decoupled-NFGNN-nodewise,nodewise-1,nodewise-2,nodewise-3}.

Despite the success of node-wise filtering, however, a critical research gap persists: the absence of a unifying formal framework that integrates both strategies impedes a deeper understanding of filtering paradigms. 
Moreover, in Section~\ref{section-CPF-motivation}, we leverage the Contextual Stochastic Block Model (CSBM; \cite{CSBM}) to reveal that a synthesis of graph-wise and node-wise filtering provides a sufficient solution for classification on graphs with mixed homophily and heterophily—characteristics prevalent in real-world graph data~\cite{dataset5-ogb,dataset6-large-hetero}. 
This analysis highlights a potential risk with node-wise filtering: its increased complexity and parameterization might lead to overfitting, ultimately diminishing model performance.

To tackle the challenges and advance filtering-based GNNs, in this paper~\footnote{Full paper available on arXiv at: ~\url{https://arxiv.org/abs/2505.14033}} we introduce {\bf C}oarsening-guided {\bf P}artition-wise {\bf F}iltering (CPF), an innovative filtering approach that operates on node partitions. 
CPF begins with \underline{structure-aware} partition-wise filtering, which filters node partitions obtained via graph coarsening algorithms~\cite{Gcoarse_survey}, capturing task-agnostic graph information. 
It then progresses to \underline{feature-aware} partition-wise filtering, refining node embeddings via filtering on clusters produced by $k$-means clustering~\cite{kmeans}, thus enabling better alignment to downstream tasks. 
Each phase of CPF is supported by fine-grained analysis, ensuring its effectiveness. 
Finally, we assess CPF on benchmark node classification tasks, along with a graph anomaly detection task to validate its practical utility. 
We list our contributions below:
\begin{itemize}[leftmargin=15pt,parsep=0pt,itemsep=0pt,topsep=1pt]
\item \textit{Identifying research problem.} We examine the limitations of node-wise filtering through the lens of the CSBM model, revealing promising pathways for advancing filtering-based GNNs.
\item \textit{Proposing new method.} We present CPF, a pioneering filtering-based GNN that operates on node partitions. 
Specifically, CPF utilizes partition-wise filtering for both graph structure and node features, extracting task-agnostic structural patterns while adaptively incorporating task-specific feature information.
\item \textit{Providing in-depth analysis.} We offer in-depth analysis for each phase of CPF, confirming its effectiveness and offering fresh insights into the research on filtering-based GNNs.
\item \textit{Conducting extensive evaluations.} We assess CPF through benchmark node classification tasks and a real-world graph anomaly detection task, validating its superiority over state-of-the-art methods and establishing it as an effective mechanism for filtering-based GNN design.
\end{itemize}


\begin{figure*}[!th]
    \centering
    \includegraphics[width=\linewidth]{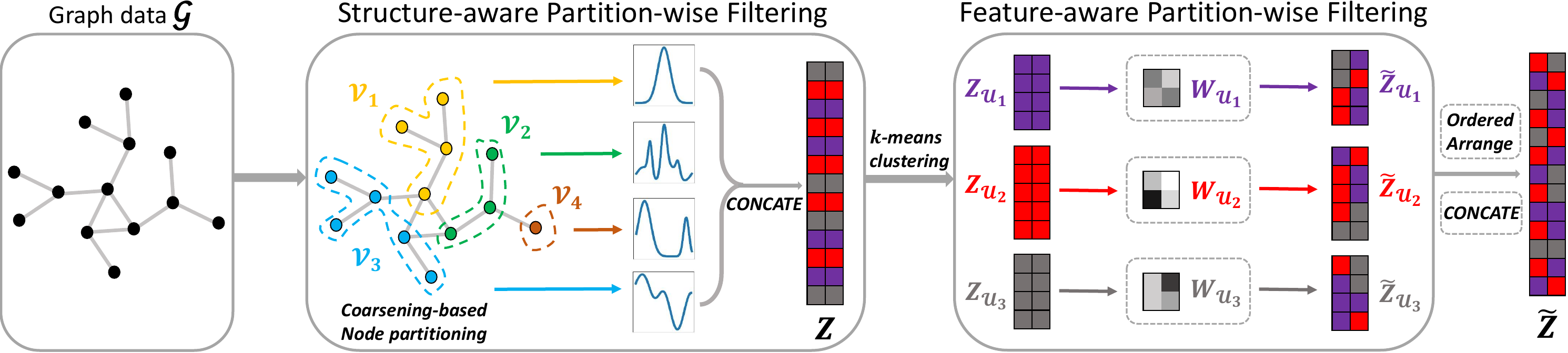}
    \caption{
    Overview of CPF’s filtering procedure. 
    It employs the partition-wise graph filtering in two aspects: structure, where partitions are obtained via graph coarsening, and features, where partitions are derived through $k$-means clustering on features.
    }
    \label{fig:CPFGNN}
\end{figure*}


\section{Preliminaries}
\label{section-preliminaries}

This section provides a foundational overview of graph coarsening, graph heterophily, and filtering-based GNNs as a streamlined background for this paper. 
Discussion covering a wider range of related works is available in Section~\ref{section:related-works}.

\subsubsection*{\bf Graph notations} Let $\mathcal{G}=(\boldsymbol{A}, \boldsymbol{X})$ be an undirected, unweighted graph with adjacency matrix $\boldsymbol{A}\in\{0,1\}^{n\times n}$ and node feature $\boldsymbol{X}\in\mathbb{R}^{n\times d}$. 
In addition, $\boldsymbol{L}=\boldsymbol{I}-\boldsymbol{D}^{-\frac{1}{2}}\boldsymbol{A}\boldsymbol{D}^{-\frac{1}{2}}$ is the normalized graph Laplacian, with $\boldsymbol{I}$, $\boldsymbol{D}$ being the identity matrix and the degree matrix.

\subsubsection*{\bf Graph coarsening} The graph coarsening techniques~\cite{Gcoarse_survey,Gcoarse_survey-2,Gcoarse-1-algebraic-distance,rsa-Gcoarse-2,rsa-Gcoarse-3-Local-Variation,rsa-Gcoarse-4-MGC-SGC,rsa-Gcoarse-5,rsa-Gcoarse-6,rsa-Gcoarse-7,rsa-Gcoarse-8-FGC} aims to reduce the size of the large graph while preserving the spectral properties~\cite{spectralgraphtheory}. 
The coarsened graph $\mathcal{G}^{\prime}$ with $n^{\prime}<n$ nodes is obtained by merging nodes from the original graph $\mathcal{G}$ into \textit{supernodes}, with $r = 1 - {n^{\prime}}/{n}$ being the \textit{coarsening ratio}. 
The coarsening process is represented via a \textit{coarsening matrix} $\boldsymbol{C}\in\mathbb{R}^{n^{\prime}\times n}$, where $C_{ij}\neq0$ if and only if $j$-th node of $\mathcal{G}$ is merged into the $i$-th supernode of $\mathcal{G}^{\prime}$. 
In general, let $m_i$ be the number of nodes mapped to the $i$-th supernode, the nonzero entries $C_{ij}$ are adopted as $C_{ij} = {1}/{m_i}$, ensuring that $\sum_{j=1}^{n}C_{ij}=1$ holds for all $i$. 

The majority of the literature evaluates the quality of a coarsening by measuring the similarity between the spectrum of the original and coarsened graphs, typically represented by the graph Laplacian $\boldsymbol{L}$. 
A widely adopted measurement for spectral similarity is the \textit{restricted spectral approximation} (RSA) constant~\cite{rsa-Gcoarse-3-Local-Variation,rsa-Gcoarse-7}:
\begin{definition}
\label{def:RSA-constant}
(\textit{RSA constant}). Consider a subspace $\mathcal{R}\subset\mathbb{R}^{n}$, a graph Laplacian $\boldsymbol{L}$, a graph signal $\boldsymbol{x}\in\mathbb{R}^{n}$, a coarsening matrix $\boldsymbol{C}$ and its pseudo-inverse $\boldsymbol{C}^{+}$. 
The RSA constant $\epsilon_{\boldsymbol{L}, \boldsymbol{C}, \mathcal{R}}$ is defined as:
\begin{align}
\label{eq:RSA}
\epsilon_{\boldsymbol{L}, \boldsymbol{C}, \mathcal{R}} = \mathop{\text{sup}}\limits_{\boldsymbol{x}\in\mathcal{R},\Vert x\Vert_{\boldsymbol{L}}=1}\Vert \boldsymbol{x}-\boldsymbol{C}^{+}\boldsymbol{C}\boldsymbol{x} \Vert_{\boldsymbol{L}}\ ,
\end{align}
where $\Vert \cdot\Vert_{\boldsymbol{L}}$ is the induced semi-norm defined as $\Vert \boldsymbol{x} \Vert_{\boldsymbol{L}}=\sqrt{\boldsymbol{x}^{\top}\boldsymbol{L}\boldsymbol{x}}$.
\end{definition}
The RSA constant is a generalization of the commonly used \textit{$\epsilon$-similarity}~\cite{rsa-Gcoarse-2,rsa-Gcoarse-3-Local-Variation,rsa-Gcoarse-4-MGC-SGC,rsa-Gcoarse-5,rsa-Gcoarse-7}, capturing how well graph signals $\boldsymbol{x} \in \mathcal{R}$ are retained through coarsening and reconstruction, expressed by $\boldsymbol{C}^{+}\boldsymbol{C}\boldsymbol{x}$, under the norm $\Vert \cdot \Vert_{\boldsymbol{L}}$. 
Thus, the goal of a coarsening algorithm is to find the matrix $\boldsymbol{C}$ that minimizes the $\epsilon_{\boldsymbol{L}, \boldsymbol{C}, \mathcal{R}}$. 

\subsubsection*{\bf Graph heterophily} Recently, a novel class of labeled graphs exhibiting {\it hterophily} has gained increasing attention in the graph learning domain~\cite{heterophily-gnn-survey}. 
Unlike conventional {\it homophily} graphs, where connected nodes often share the same labels, heterophilic graphs defy this assumption by featuring diverse label distributions within local substructures. 
Various metrics have been proposed to measure heterophily in a graph, with node- and edge-level metrics being the most widely used. 
Readers interested in a more comprehensive understanding can refer to recent surveys~\cite{heterophily-gnn-survey,heterophily-gnn-survey-2,heterophily-gnn-survey-3}. 
In summary, this foundational distinction presents significant challenges for applying GNNs to node classification tasks.

Among the existing works on designing advanced GNNs to address heterophily, this paper focuses on a distinct branch that constructs critical graph filters to extract key information, leading to the class of filtering-based GNNs~\cite{GPRGNN,BernNet-GNN-narrowbandresults-1,chebnet2d,ChebNetII,OptBasisGNN,JacobiConv,ERGNN,decoupled-UniFilter,decoupled-PCConv,decoupled-AdaptKry,decoupled-TFGNN}.

\subsubsection*{\bf Filtering on graphs} Let $\boldsymbol{x}\in\mathbb{R}^{n}$ be a signal on graph $\mathcal{G}$, namely a single \textbf{column} of the node feature matrix $\boldsymbol{X}$. 
Let $\boldsymbol{L}=\boldsymbol{U}\mathrm{diag}(\boldsymbol{\lambda})\boldsymbol{U}^{\top}$ be the eigendecomposition of graph Laplacian, with $\boldsymbol{U}$, $\boldsymbol{\lambda}$ being the eigenvectors and eigenvalues, respectively. 
A spectral graph filter is a point-wise mapping $f: \mathbb{R}\mapsto\mathbb{R}$ applied to $\boldsymbol{\lambda}$~\cite{GraphSignalProcessingOverviewChallengesandApplications,GraphSignalProcessingforMachineLearningAReviewandNewPerspectives,DiscreteSignalProcessingonGraphs,DiscretesignalprocessingongraphsGraphfouriertransform,DiscretesignalprocessingongraphsGraphfilters}. 
Thus, the filtering operation on $\boldsymbol{x}$ with a filter $f(\cdot)$ is defined as:
\begin{align}
\boldsymbol{z} \triangleq \boldsymbol{U}\mathrm{diag}(f(\boldsymbol{\lambda}))\boldsymbol{U}^{\top}\boldsymbol{x}\ ,
\label{eq:graph-filtering}
\end{align}
where $\boldsymbol{z}\in\mathbb{R}^{n}$ is the filtered output. 
Due to the intensive computation cost of eigendecomposition, the mapping $f$ is typically implemented via an order-$K$ polynomial approximation in practice~\cite{GPRGNN,BernNet-GNN-narrowbandresults-1,chebnet2d,ChebNetII,OptBasisGNN,JacobiConv,decoupled-UniFilter,decoupled-PCConv,decoupled-AdaptKry,decoupled-TFGNN}, which rewrites Eq.~\eqref{eq:graph-filtering} as: 
\begin{align}
\label{eq:graph-filtering-polynomial-graphwise}
\boldsymbol{z} = \boldsymbol{U}\mathrm{diag}\left(\sum_{k=0}^{K}\theta_{k}\mathbf{T}_{k}(\boldsymbol{\lambda})\right)\boldsymbol{U}^{\top}\boldsymbol{x} = \sum_{k=0}^{K}\theta_{k}\mathbf{T}_{k}(\boldsymbol{L})\boldsymbol{x}\ ,
\end{align}
where $\mathbf{T}_{k}$ is reloaded both as an element-wise function for a vector input $\boldsymbol{\lambda}$ and a matrix polynomial for matrix inputs, denoting the $k$-th term of a specific polynomial; $\theta_{k}$ is a (trainable) coefficient. 
Eq.~\eqref{eq:graph-filtering-polynomial-graphwise} represents \textit{graph-wise filtering}, indicating that a single operator $\sum_{k=0}^{K}\theta_{k}\mathbf{T}_{k}(\boldsymbol{L})$ is applied to process the features across all the nodes. 

In contrast, the \textit{node-wise filtering} can be formally expressed as:
\begin{align}
\label{eq:graph-filtering-polynomial-nodewise}
\boldsymbol{z}_{i} = \boldsymbol{\delta}_{i}\cdot\sum_{k=0}^{K}\theta_{ik}\mathbf{T}_{k}(\boldsymbol{L})\boldsymbol{x}\ ,\ i=1,2,...,n\ .
\end{align}
The one-hot vector $\boldsymbol{\delta}_{i}=\left[0,0,...,0,1,0,...,0\right] \in \{0, 1\}^{n}$ serves as a sifting vector, designed to extract information for the $i$-th node. 
As shown in Eq.~\eqref{eq:graph-filtering-polynomial-nodewise}, node-wise filtering, in contrast to graph-wise filtering of Eq.~\eqref{eq:graph-filtering-polynomial-graphwise}, assigns a distinct operator $\sum_{k=0}^{K}\theta_{ik}\mathbf{T}_{k}(\boldsymbol{L})$ to each node $i$. 
This strategy improves model performance but comes at the cost of higher computational complexity~\cite{decoupled-NFGNN-nodewise,decoupled-NFGNN-nodewise,nodewise-1,nodewise-3}. 


\section{Methodology}
\label{section-CPF}
This section provides a detailed introduction to the proposed CPF method. 
We start by presenting an intuitive motivation, supported by theoretical insights to provide clarity. 
Next, we formally present CPF and illustrate its components in detail.


\subsection{Motivation}
\label{section-CPF-motivation}

Conventional graph-wise filtering methods have shown promising performance in node classification tasks~\cite{GPRGNN,ChebNetII,JacobiConv,OptBasisGNN,decoupled-AdaptKry}. 
However, recent studies reveal that these methods lack flexibility by enforcing a uniform filter across all nodes, limiting their effectiveness on heterophilic graphs~\cite{decoupled-NFGNN-nodewise,nodewise-1,nodewise-3}. 
To address this, the recent research introduces node-wise filtering that assigns unique filters to individual nodes, supported by both theoretical and empirical evidence validating its superiority over graph-wise filtering.

While node-wise filtering approaches provide greater flexibility, they typically enforce each node to use a unique filter. 
This configuration, however, lacks deeper consideration of the actual filtering needs among nodes, prompting us to ask: {\bf Must the required filters always differ among nodes?} 
Drawing inspiration from recent research that explores node-wise filtering in heterophilic graphs via the Contextual Stochastic Block Model (CSBM; \cite{CSBM}), we establish the following Proposition~\ref{proposition:optimal-is-hybrid}. 
It suggests that the theoretically optimal filtering paradigm—one that correctly classifies nodes while minimizing the number of filters (parameters)—is a hybrid of graph-wise and node-wise filtering.

\begin{proposition}
\label{proposition:optimal-is-hybrid}
Consider a binary-class graph $\mathcal{G}$ generated from the distribution $CSBM(n, \boldsymbol{\mu}, \boldsymbol{\nu}, (p_0, q_0), (p_1, q_1), P)$, as defined in~\cite{nodewise-3}. Here, $\boldsymbol{\mu}$ and $\boldsymbol{\nu}$ define Gaussian distributions for the generation of random node features, while $(p_0, q_0)$, $(p_1, q_1)$, and $P$ control the graph structure’s statistical properties.

Define $\mathcal{V}_{\text{homo}}$ and $\mathcal{V}_{\text{hete}}$ as the homophilic and heterophilic node subsets, respectively. A minimal-parameter solution for achieving linear separability across all nodes is attainable using a uniform (graph-wise) filter for $\mathcal{V}_{\text{homo}}$ and node-specific filters for $\mathcal{V}_{\text{hete}}$.
\end{proposition}

\subsubsection*{\bf Remark} The proof can be found in Appendix~\ref{appendix:proposition:optimal-is-hybrid}. 
The proposition above underscores that a {\bf synthesis of graph-wise and node-wise filtering} suffices for classification on graphs exhibiting both homophilic and heterophilic properties—characteristics prevalent in real-world datasets such as ogb~\cite{dataset5-ogb} and LINKX~\cite{dataset6-large-hetero}. 
Given this insight, node-wise filtering, which applies a unique filter to each node, risks excessive parameterization. 
The heightened model complexity can introduce adverse effects, including an increased risk of overfitting. 
Motivated by the observed drawback in node-wise filtering, we aim to establish a more advanced paradigm by introducing a unified framework that integrates both graph and node-wise filtering, i.e., the proposed {\bf partition-wise filtering}.

\subsection{The Proposed CPF Framework: An Overview}
\label{section-CPF-overview}

This section provides an overview of CPF, our proposed model. 
As depicted in Figure~\ref{fig:CPFGNN}, CPF operates through two successive filtering steps: {\bf structure-aware partition-wise filtering} and {\bf feature-aware partition-wise filtering}. 
Given an input graph $\mathcal{G}$, CPF first performs partition-wise filtering on node partitions generated via task-agnostic graph coarsening, extracting structural information. 
Subsequently, CPF applies filtering on partitions formed through feature coarsening, achieved via $k$-means clustering, to refine node embeddings for task-specific adaptation.
The final embeddings, $\tilde{\boldsymbol{Z}}$, serve as inputs for downstream tasks such as loss computation. 
The following subsections elaborate on these steps.

\subsection{Structure-aware Partition-wise Filtering}
\label{section-CPF-structure-partition-filtering}

CPF first employs a structure-aware partition-wise filtering, utilizing the theoretical framework of structural (graph) coarsening~\cite{Gcoarse_survey}. 
This process involves two sequential steps: \ding{182} {\it graph coarsening-guided partitioning} and \ding{183} {\it filtering on coarsening-guided partitions}. 
The following sections provides an in-depth examination of each.

\subsubsection{\bf \ding{182} Graph coarsening-guided partitioning}\label{section-CPF-structure-partition-filtering:partitioning} 
CPF performs node partitioning by utilizing a preselected graph coarsening algorithm. 
This algorithm can be any existing method designed to preserve the spectral properties of the graph, typically by leveraging the RSA constant or more advanced graph coarsening metrics~\cite{rsa-Gcoarse-2,rsa-Gcoarse-3-Local-Variation,rsa-Gcoarse-4-MGC-SGC,rsa-Gcoarse-5,rsa-Gcoarse-6,rsa-Gcoarse-7,rsa-Gcoarse-8-FGC}. 
With a preset coarsening ratio $r$, $0\leq r \leq\frac{n-1}{n}$, the coarsening algorithm applied to the input graph $\mathcal{G}$ generates a coarsening matrix $\boldsymbol{C}$, which groups the nodes into $n^{\prime}=\lfloor (1-r) n \rfloor$ disjoint clusters (supernodes), represented as $\mathcal{V}=\{\mathcal{V}_{1},\mathcal{V}_{2},...,\mathcal{V}_{n^{\prime}}\}$. 
Each node cluster $\mathcal{V}_{i}$ represents a partition in the structural aspect, directing the subsequent structure-aspect partition-wise filtering operation.

The node partitioning induced by graph coarsening uncovers critical structural insights and extracts core graph information that is {\bf task-agnostic}. 
We summarize the essences to a Theorem~\ref{theorem:message-passing-guarantee}:

\begin{theorem}
\label{theorem:message-passing-guarantee}
Let $\boldsymbol{\Delta}=\boldsymbol{L}-\boldsymbol{I}$ and $\boldsymbol{\Pi}=\boldsymbol{C}^{+}\boldsymbol{C}$. 
For the $k$-th order term of the polynomial-based operator, denoted as $\boldsymbol{\Delta}^{k}$, the filtered signal $\boldsymbol{\Delta}^{k}\boldsymbol{x}$ satisfies the following inequality:
\begin{equation}
\label{eq:message-passing-generalized-1}
\lVert \boldsymbol{\Delta}^{k}\boldsymbol{x} - \boldsymbol{\Pi}\boldsymbol{\Delta}\boldsymbol{\Pi}\boldsymbol{\Delta}^{k-1}\boldsymbol{x} \rVert_{\boldsymbol{L}}  \leq \epsilon_{\boldsymbol{L}, \boldsymbol{C}, \mathcal{R}} \Vert \boldsymbol{x} \Vert_{\boldsymbol{L}} \left( \lVert \boldsymbol{\Delta} \rVert_{\boldsymbol{L}} + \lVert \boldsymbol{\Pi}\boldsymbol{\Delta} \rVert_{\boldsymbol{L}} \right)\ .
\end{equation}
Furthermore, when the coarsening algorithm achieves a sufficiently small $\epsilon_{\boldsymbol{L}, \boldsymbol{C}, \mathcal{R}} \rightarrow 0$, the following approximation result holds:
\begin{equation}
\label{eq:message-passing-generalized-2}
\lVert \boldsymbol{\Delta}^{k}\boldsymbol{x} - (\boldsymbol{\Pi}\boldsymbol{\Delta}\boldsymbol{\Pi})^{k}\boldsymbol{x} \rVert_{\boldsymbol{L}} \rightarrow 0.
\end{equation}
\end{theorem}

\subsubsection*{\bf Remark} The proof can be found in Appendix~\ref{appendix:theorem:message-passing-guarantee}. 
As demonstrated by the above theorem, the core operation in filtering-based GNNs, graph propagation $\boldsymbol{L}\boldsymbol{x}$, is consistently approximated by the term $\boldsymbol{\Pi}\boldsymbol{\Delta}\boldsymbol{\Pi}\boldsymbol{\Delta}\boldsymbol{x}$, which corresponds to the propagation on coarsened graph. 
This means that message propagation across nodes in $\mathcal{G}$ is structurally equivalent to the propagation between clusters $\mathcal{V}_{1},\mathcal{V}_{2},...,\mathcal{V}_{n^{\prime}}$ derived from graph coarsening. 
Such equivalence emphasizes the role of node partitions in capturing task-agnostic graph information, further underscoring the rationale for partition-based filtering, which will be detailed in the following section.

Additionally, graph coarsening methods are typically unsupervised algorithms that rely solely on the graph structure, enabling node partitioning to be \textbf{precomputed} and treated as a fixed input for subsequent operations. 
This scheme separates node partitioning from the model training, promoting the efficiency of the CPF method. 
In this paper, we adopt the \textit{Local Variation} method~\cite{rsa-Gcoarse-3-Local-Variation}, which coarsens graphs by optimizing RSA constant, for our main experiments, while other methods are explored in ablation studies.

\subsubsection{\bf \ding{183} Filtering on coarsening-guided partitions}\label{section-CPF-structure-partition-filtering:filtering} 
In analogy to node-wise filtering, partition-wise filtering performs graph filtering operation to each partition $\mathcal{V}_{i}$, $i=1,2,...,n^{\prime}$, through an independent polynomial-based graph filter, $\sum_{k=0}^{K}\theta_{ik}\mathbf{T}_{k}(\boldsymbol{L})$. 
The resulted output for partition $\mathcal{V}_{i}$, namely $\boldsymbol{Z}_{\mathcal{V}_{i}}$, is formulated below:
\begin{equation}
\label{eq:partion-filtering}
\boldsymbol{Z}_{\mathcal{V}_{i}}=
\frac{1}{|\mathcal{V}_{i}|} \sum_{m\in\mathcal{V}_{i}} \mathrm{diag}(\boldsymbol{\delta}_{m}) \sum_{k=0}^{K}\theta_{ik}\mathbf{T}_{k}(\boldsymbol{L})\boldsymbol{X}\ ,\ i=1,2,...,n^{\prime}\ .
\end{equation}
The design guarantees that nodes within the same partition are subjected to a unified filter, facilitating the extraction of structural information discussed in Section~\ref{section-CPF-structure-partition-filtering:partitioning}. 
As a result, the node embeddings $\boldsymbol{Z}$ are obtained by concatenating all $\boldsymbol{Z}_{\mathcal{V}_{i}}$ in their node order, which will be utilized and optimized in downstream tasks.

Furthermore, to facilitate a streamlined implementation, we present a {\bf unified and equivalent formulation} of \Cref{eq:partion-filtering}, with a detailed derivation provided in Appendix~\ref{appendix:derivation}.

\begin{equation}
\label{eq:partion-filtering-unified}
\boldsymbol{Z} = \sum_{k=0}^{K} \mathrm{diag}(\boldsymbol{C}^{+}\boldsymbol{\Theta_{:k+1}}) \mathbf{T}_{k}(\boldsymbol{L})\boldsymbol{X}\ .
\end{equation}

Here, the parameter matrix $\boldsymbol{\Theta} \in \mathbb{R}^{n^{\prime} \times (K+1)}$ consolidates the coefficient $\theta_{ik} / |\mathcal{V}_{i}|$ into a unified representation, enabling a streamlined and unified implementation of the filtering process. 
This formulation offers deeper insight into the interplay between graph-wise and node-wise filtering introduced in Section~\ref{section-preliminaries}, revealing their unification within our CPF, as stated in the proposition below.

\begin{proposition}
\label{proposition:unifying}
Recall that the coarsening ratio $r$ spans the range $\left[0,\frac{n-1}{n}\right]$.
When the coarsening ratio $r=\frac{n-1}{n}$, Eq.~\ref{eq:partion-filtering-unified} performs graph-wise filtering; when $r=0$, Eq.~\ref{eq:partion-filtering-unified} performs node-wise filtering.
\end{proposition}

\subsubsection*{\bf Remark} The proof can be found in Appendix~\ref{appendix:proposition:unifying}. 
This proposition asserts that the CPF framework encapsulates both graph-wise and node-wise filtering as extreme cases. 
Beyond the two strategies, when $r$ is set within the open interval $\left(0, {(n-1)}/{n}\right)$, the resulting node partitions may include both multiple nodes and single node; 
this scenario suggests partition-wise filtering can be viewed as a synthesis between the two strategies: 
increasing $r$ moves partition-wise filtering to the graph-wise case, whereas with decreasing $r$ CPF leans to node-wise case. 
Hence, CPF offers an advanced paradigm that retains the flexibility of node-wise filtering but also integrates graph-wise filtering to reduce the parameter redundancy, providing a viable remedy for the limitations mentioned in Section~\ref{section-CPF-motivation}.

\subsection{Feature-aware Partition-wise Filtering}
\label{section-CPF-feature-partition-filtering}

Since structure-aware partition-wise filtering operates on node partitions that are task-agnostic, the obtained embeddings $\boldsymbol{Z}$ may not be optimally aligned with downstream tasks, even when filter coefficients $\boldsymbol{\Theta}$ (Eq.~\ref{eq:partion-filtering-unified}) are optimized. 
To address this limitation, we propose feature-aware partition-wise filtering, which enhances $\boldsymbol{Z}$ for more task-relevant representation learning. 
This refinement consists of two steps: \ding{182} {\it feature coarsening-guided node partitioning} and \ding{183} {\it class-wise feature filtering}, both of which are detailed below.

\subsubsection{\bf \ding{182} Feature coarsening-guided node partitioning}\label{section-CPF-feature-partition-filtering:partitioning} 

Unlike the partitioning approach discussed in Section~\ref{section-CPF-structure-partition-filtering:partitioning}, which is based solely on structural information, we incorporate feature considerations by partitioning nodes based on feature-driven ``coarsening'', specifically by performing clustering~\cite{clustering-survey} on node embeddings. 
The number of partitions is set to the number of node classes. 
This approach is motivated by the role of node embeddings $\boldsymbol{Z}$, which serve as predictions for node labels in classification tasks. 
Hence, clustering nodes with similar embeddings mirrors the partitioning of nodes into distinct categories.

Specifically, we, rather than employing more complex algorithms, apply the well-established {\it $k$-means} algorithm~\cite{kmeans}, which continues to be widely adopted for its proven stability and effectiveness~\cite{kmeans-stability}. 
Thus, using the $k$-means algorithm, we can divide $\boldsymbol{Z}$ into $c$ partitions: $\boldsymbol{Z}_{\mathcal{U}_{1}}, \boldsymbol{Z}_{\mathcal{U}_{2}},..., \boldsymbol{Z}_{\mathcal{U}_{c}}$, with $c$ representing the number of node categories and $\mathcal{U}_{j}$, $j=1,2,...,c$, referring to the node partitions.

\subsubsection{\bf \ding{183} Class-wise feature filtering}\label{section-CPF-feature-partition-filtering:filtering} 

With the partitions of node embeddings $\boldsymbol{Z}_{\mathcal{U}_{1}}, \boldsymbol{Z}_{\mathcal{U}_{2}},..., \boldsymbol{Z}_{\mathcal{U}_{c}}$ obtained, we propose class-wise feature filtering, which associates a distinct linear transformation $\boldsymbol{W}_{\mathcal{U}_{j}}$ to each partition, as outlined below:

\begin{equation}
\label{eq:class-feature-filtering}
\tilde{\boldsymbol{Z}}_{\mathcal{U}_{j}} = \boldsymbol{Z}_{\mathcal{U}_{j}} \boldsymbol{W}_{\mathcal{U}_{j}}\ ,\ j=1,2,...,c\ ,
\end{equation}

where $\tilde{\boldsymbol{Z}}_{\mathcal{U}_{j}}$ are consolidated into the final output for downstream tasks. 
The class-wise filtering operates as a task-relevant refinement to the embeddings $\boldsymbol{Z}$, as outlined in the following proposition:

\begin{proposition}
\label{proposition:feature-adjustment}
Let $\boldsymbol{z}_{1}, \boldsymbol{z}_{2}, \boldsymbol{z}_{3} \in \mathbb{R}^c$ and define $d_{12} = \|\boldsymbol{z}_{1} - \boldsymbol{z}_{2}\|$ and $d_{13} = \|\boldsymbol{z}_{1} - \boldsymbol{z}_{3}\|$. Assume that $d_{12} < d_{13}$. Then:
\begin{enumerate}[leftmargin=*,parsep=0pt,itemsep=0pt,topsep=1pt]
    \item There exists $\boldsymbol{W} \in \mathbb{R}^{c \times c}$, such that $\|\boldsymbol{W}\boldsymbol{z}_{1} - \boldsymbol{W}\boldsymbol{z}_{2}\| > \|\boldsymbol{W}\boldsymbol{z}_{1} - \boldsymbol{W}\boldsymbol{z}_{3}\|$
    \item There exist $\boldsymbol{W}_1$, $\boldsymbol{W}_2 \in \mathbb{R}^{c \times c}$, such that $\|\boldsymbol{W}_1\boldsymbol{z}_{1} - \boldsymbol{W}_1\boldsymbol{z}_{2}\| > \|\boldsymbol{W}_1\boldsymbol{z}_{1} - \boldsymbol{W}_2\boldsymbol{z}_{3}\|$
\end{enumerate}
\end{proposition}

\subsubsection*{\bf Remark} Proof can be found in Appendix~\ref{appendix:proposition:feature-adjustment}. 
This proposition suggests that class-wise filtering facilitates the refinement of both intra-cluster and inter-cluster relationships in node embeddings, enabling their adaptive alignment with true labels in downstream tasks. 
For instance, nodes within the same cluster may belong to different classes, yet their embeddings in $\boldsymbol{Z}$ might be overly similar. 
To reduce misclassification, class-wise filtering increases the distance between correctly classified and misclassified nodes while simultaneously adjusting the latter toward their true class. 
This acts as a corrective step that enhances classification accuracy.

\subsection{Enhancing Scalability with Decoupled Architecture}
\label{section-CPF-decoupled-gnn}

CPF adopts a decoupled GNN architecture that separates graph filtering from feature transformation. 
This design principle, first proposed by~\cite{decoupled-advantages-3-coupled-disadvantages-2-APPNP}, has become a de facto choice in modern \textit{spectral GNNs} for its significant efficacy and computational efficiency~\cite{GPRGNN,BernNet-GNN-narrowbandresults-1,chebnet2d,ChebNetII,OptBasisGNN,JacobiConv,decoupled-UniFilter,decoupled-PCConv,decoupled-AdaptKry}, and even stands out as a promising solution for \textit{scalable GNNs}~\cite{decoupled-LD2,decoupled-SCARA}. 
In the following, we present two versions of CPF to accommodate different graph sizes. 

\ding{182}~Specifically, in the case of \textbf{medium-to-large} graphs like Cora~\cite{dataset1-cora} and Arxiv~\cite{dataset5-ogb}, CPF operates as described below:
\begin{equation}
\label{eq:decoupled-CPF-medium}
\boldsymbol{H}=\mathrm{MLP}(\boldsymbol{X})\ ,\ \tilde{\boldsymbol{Z}}=\mathrm{FILTERING}(\boldsymbol{H};\mathcal{V},\mathcal{G})\ .
\end{equation}
$\mathrm{MLP}(\cdot)$ denotes the multi-layer perceptron for feature transformation, and $\mathrm{FILTERING}(\cdot\ ;\mathcal{V},\mathcal{G})$ refers to the partition-wise filtering.

\ding{183}~In the case of \textbf{exceptionally large} graphs, such as Wiki~\cite{dataset6-large-hetero} and Papers100M~\cite{dataset5-ogb}, CPF is implemented as follows:
\begin{equation}
\label{eq:decoupled-CPF-large}
\boldsymbol{H}=\mathrm{FILTERING}(\boldsymbol{X};\mathcal{V},\mathcal{G})\ ,\ \tilde{\boldsymbol{Z}}=\mathrm{MLP}(\boldsymbol{H})\ .
\end{equation}
\subsubsection*{\bf Remark} The different implementations of CPF come from hardware constraints and introduce notable benefits: 
(\rom{1})~for medium-to-large graphs, the graph data can be fully stored on GPUs; 
therefore, by simply reducing the feature dimensions with MLP, the subsequent filtering process could achieves high efficiency; 
(\rom{2})~for exceptionally large graphs, where GPU memory limitations become a very issue, CPF precomputes features, $\boldsymbol{L}^{k}\boldsymbol{X}$, $k=1,2,...K$, and stores them as static data files. 
This allows for efficient filtering operation through repeated reads of precomputed features, mitigating the intense computational complexity associated with GNN training; 
an efficient MLP is applied later.

\subsection{Complexity Analysis of CPF}
\label{section-CPF-complexity}

This subsection is devoted to the complexity analysis of CPF. 
Specifically, we will focus on filtering operations since, as we mentioned in \Cref{section-CPF-decoupled-gnn}, 
modern filtering-based GNNs typically adopt a decoupled architecture and the feature transformation parts are usually the same~\cite{GPRGNN,BernNet-GNN-narrowbandresults-1,chebnet2d,ChebNetII,OptBasisGNN,JacobiConv,decoupled-UniFilter,decoupled-TFGNN,decoupled-PCConv,decoupled-AdaptKry}.

To start, we consider a graph $\mathcal{G}$ with $n$ nodes, $E$ edges, and $c$ node classes. 
Across all filtering-based GNNs, the maximum order of polynomial filters is set to $K$. 
Below, we outline the computational complexity of each filtering paradigm.

\begin{itemize}[leftmargin=*]
    \item {\bf Graph-wise filtering} adopts a complexity of $\mathcal{O}(EcK)$.
    \item {\bf Node-wise filtering} adopts a complexity of $\mathcal{O}(nEcK)$.
    \item {\bf Our CPF method (partition-wise filtering)} adopts a total complexity of $\mathcal{O}(n(c^{2}+cK) + EcK)$.
\end{itemize}

\subsubsection*{\bf In-Depth Examination of CPF’s Complexity} The complexity of CPF is analyzed in the order of its framework. 
In {\bf structure-aware partition-wise filtering}, CPF first performs spectral graph coarsening, which has an approximate complexity of $\mathcal{O}(n+E)$ using Local Variation~\cite{rsa-Gcoarse-2,rsa-Gcoarse-3-Local-Variation}. 
It then performs partition-wise filtering, where computing the coefficient $\boldsymbol{\Theta}$ requires $\mathcal{O}(nK)$, propagating terms $\boldsymbol{L}^{k}\boldsymbol{X}$ takes $\mathcal{O}(EcK)$, and obtaining $\boldsymbol{Z}$ incurs $\mathcal{O}(ncK)$. 
For {\bf feature-aware partition-wise filtering}, CPF first clusters $\boldsymbol{Z}$ into $c$ partitions through $k$-means algorithm, which has a complexity of $\mathcal{O}(nc^{2})$~\cite{kmeans-complexity}, and then applies a class-aware filtering with the same complexity. 
Summing all terms, our CPF has a total complexity of $\mathcal{O}(n(c^{2}+cK) + EcK)$.

\subsubsection*{\bf Remark} As discussed above, the complexity of CPF scales with the addition $n + E$. 
While this is higher than graph-wise method, which scales solely with $E$, CPF remains markedly more efficient than node-wise filtering, which incurs a notably higher cost of multiplication $n \cdot E$. 
Moreover, on exceptionally large graphs, CPF can further improve its efficiency by precomputing the terms $\boldsymbol{L}^{k}\boldsymbol{X}$.

\begin{table*}[!th]
\caption{Node classification results on medium-to-large graphs. 
\#Improv. denotes the performance gain of CPF over the best baseline result. 
Along all the tables in this paper, boldface represents the first result, while underlined indicates the runner-up.
}
\vskip -0.05in
\label{table-node-classify-medium}
\centering
\setlength{\tabcolsep}{4.5pt}
\resizebox{\linewidth}{!}{
\begin{tabular}{cccccc|ccccc}
\hline
\multirow{2}{*}{\makecell[c]{Model \\ Type}} & \multirow{2}{*}{Method}              & \multicolumn{4}{c|}{homophilic graphs}                                                                                        & \multicolumn{5}{c}{heterophilic graphs}                                                                                                                       \\ \cline{3-11} 
                      &                                      & Cora                          & Cite.                         & Pubmed                        & Arxiv                         & Roman.                        & Amazon.                       & Ques.                         & Gamers                        & Genius                        \\ \hline
\multirow{4}{*}{\large \romannumeral1}    & H2GCN                                & $87.33_{\pm0.6}$              & $75.11_{\pm1.2}$              & $88.39_{\pm0.6}$              & $71.93_{\pm0.4}$              & $61.38_{\pm1.2}$              & $37.17_{\pm0.5}$              & $64.42_{\pm1.3}$              & $64.71_{\pm0.4}$              & $90.12_{\pm0.2}$              \\
                      & GLOGNN                               & $88.12_{\pm0.4}$              & $76.23_{\pm1.4}$              & $88.83_{\pm0.2}$              & $72.08_{\pm0.3}$              & $71.17_{\pm1.2}$              & $42.19_{\pm0.6}$              & $74.42_{\pm1.3}$              & $65.62_{\pm0.3}$              & $90.39_{\pm0.3}$              \\
                      & LINKX                                & $84.51_{\pm0.6}$              & $73.25_{\pm1.5}$              & $86.36_{\pm0.6}$              & $71.14_{\pm0.2}$              & $67.55_{\pm1.2}$              & $41.57_{\pm0.6}$              & $63.85_{\pm0.8}$              & $65.82_{\pm0.4}$              & $\underline{91.12_{\pm0.5}}$  \\
                      & OrderGNN                             & $87.55_{\pm0.2}$              & $75.46_{\pm1.2}$              & $88.31_{\pm0.3}$              & $71.90_{\pm0.5}$              & $71.69_{\pm1.6}$              & $40.93_{\pm0.5}$              & $70.82_{\pm1.0}$              & $66.09_{\pm0.3}$              & $89.45_{\pm0.4}$              \\
                      \hline
\multirow{5}{*}{\large \romannumeral2}   & GCN                                  & $86.48_{\pm0.4}$              & $75.23_{\pm1.0}$              & $87.29_{\pm0.2}$              & $71.77_{\pm0.1}$              & $72.33_{\pm1.6}$              & $42.09_{\pm0.6}$              & $75.17_{\pm0.8}$              & $63.29_{\pm0.5}$              & $86.73_{\pm0.5}$              \\
                      & GCNII                                & $86.77_{\pm0.2}$              & $\underline{76.57_{\pm1.5}}$ & $88.86_{\pm0.4}$              & $71.72_{\pm0.4}$              & $71.62_{\pm1.7}$              & $40.89_{\pm0.4}$              & $72.32_{\pm1.0}$              & $65.11_{\pm0.3}$              & $90.60_{\pm0.6}$              \\
                      & ChebNet                              & $86.83_{\pm0.7}$              & $74.39_{\pm1.3}$              & $85.92_{\pm0.5}$              & $71.52_{\pm0.3}$              & $64.44_{\pm1.5}$              & $38.81_{\pm0.7}$              & $70.42_{\pm1.2}$              & $63.62_{\pm0.4}$              & $87.42_{\pm0.2}$              \\
                      & ACMGCN                               & $87.21_{\pm0.4}$              & $76.03_{\pm1.4}$              & $87.37_{\pm0.4}$              & $71.70_{\pm0.3}$              & $66.48_{\pm1.2}$              & $39.53_{\pm0.9}$              & $67.84_{\pm0.5}$              & $64.73_{\pm0.3}$              & $83.45_{\pm0.7}$              \\
                      & Specformer                           & $88.19_{\pm0.6}$              & $75.87_{\pm1.5}$              & $88.74_{\pm0.2}$              & $71.88_{\pm0.2}$              & $71.69_{\pm1.4}$              & $42.06_{\pm0.8}$              & $70.75_{\pm1.2}$              & $65.80_{\pm0.2}$              & $89.39_{\pm0.6}$              \\
                      \hline
\multirow{6}{*}{\large \romannumeral3}  
                      & GPRGNN                               & $88.26_{\pm0.5}$              & $76.24_{\pm1.2}$              & $88.81_{\pm0.2}$              & $71.89_{\pm0.2}$              & $64.49_{\pm1.6}$              & $41.48_{\pm0.6}$              & $64.58_{\pm1.2}$              & $66.23_{\pm0.1}$              & $90.92_{\pm0.6}$              \\
                      & ChebNetII                            & $88.17_{\pm0.4}$              & $76.41_{\pm1.3}$              & $88.98_{\pm0.4}$              & $72.13_{\pm0.3}$              & $66.77_{\pm1.2}$              & $42.44_{\pm0.9}$              & $71.28_{\pm0.6}$              & $66.44_{\pm0.5}$              & $90.60_{\pm0.2}$              \\
                      & OptBasis                             & $88.35_{\pm0.6}$              & $76.22_{\pm1.4}$              & $89.38_{\pm0.3}$              & $72.10_{\pm0.2}$              & $64.28_{\pm1.8}$              & $41.63_{\pm0.8}$              & $69.60_{\pm1.2}$              & $66.81_{\pm0.4}$  & $90.97_{\pm0.5}$              \\
                      & JacobiConv                           & $\underline{88.53_{\pm0.8}}$  & $76.27_{\pm1.3}$              & $\underline{89.51_{\pm0.2}}$  & $71.87_{\pm0.3}$              & $70.10_{\pm1.7}$              & $42.18_{\pm0.4}$              & $72.16_{\pm1.3}$              & $64.17_{\pm0.3}$              & $89.32_{\pm0.5}$              \\
                      & AdaptKry                             & $88.23_{\pm0.7}$              & $76.54_{\pm1.2}$  & $88.38_{\pm0.6}$              & $72.33_{\pm0.3}$              & $71.40_{\pm1.3}$              & $42.31_{\pm1.1}$              & $72.55_{\pm1.0}$              & $66.27_{\pm0.3}$              & $90.55_{\pm0.3}$              \\
                      & UniFilter                            & $88.31_{\pm0.7}$              & $76.38_{\pm1.1}$              & $89.30_{\pm0.4}$              & $\underline{72.87_{\pm0.4}}$  & $71.22_{\pm1.5}$              & $41.37_{\pm0.6}$              & $73.83_{\pm0.8}$              & $65.75_{\pm0.4}$              & $90.66_{\pm0.2}$              \\ \hline
\multirow{3}{*}{\large \romannumeral4}   & NIGCN                      & $88.29_{\pm0.5}$              & $76.38_{\pm1.2}$              & $88.69_{\pm0.3}$              & $72.22_{\pm0.2}$              & $72.18_{\pm1.1}$  & $\underline{42.66_{\pm0.4}}$              & $\underline{75.68_{\pm0.8}}$  & $66.72_{\pm0.5}$              & $90.93_{\pm0.6}$              \\
                      & NODE-MOE$^{\S}$                      & $87.16_{\pm0.8}$              & $75.67_{\pm1.6}$              & $88.54_{\pm0.5}$              & $71.86_{\pm0.5}$              & $72.25_{\pm1.8}$              & $42.33_{\pm1.2}$              & $74.46_{\pm1.2}$              & $66.42_{\pm0.2}$              & $90.39_{\pm0.4}$              \\
                      & NFGNN                                & $88.06_{\pm0.4}$              & $76.22_{\pm1.4}$              & $88.43_{\pm0.4}$              & $72.15_{\pm0.3}$              & $\underline{72.46_{\pm1.2}}$  & $42.19_{\pm0.3}$              & $75.49_{\pm0.9}$  & $\underline{66.85_{\pm0.4}}$              & $90.87_{\pm0.5}$              \\ \hline
\multirow{2}{*}{}     & CPF (\textbf{Ours}) & \textbf{89.28}$\boldsymbol{_{\pm0.7}}$ & \textbf{77.55}$\boldsymbol{_{\pm0.9}}$              & \textbf{89.93}$\boldsymbol{_{\pm0.5}}$ & \textbf{75.46}$\boldsymbol{_{\pm0.3}}$ & \textbf{74.02}$\boldsymbol{_{\pm1.1}}$ & \textbf{46.12}$\boldsymbol{_{\pm0.7}}$ & \textbf{78.80}$\boldsymbol{_{\pm1.1}}$ & \textbf{69.77}$\boldsymbol{_{\pm0.3}}$ & \textbf{92.68}$\boldsymbol{_{\pm0.3}}$ \\
                      & \#Improv.                            & $0.75\%$                      & $0.98\%$                     & $0.42\%$                      & $2.59\%$                      & $1.56\%$                      & $3.46\%$                      & $3.12\%$                      & $2.92\%$                      & $1.56\%$                      \\ \hline
\end{tabular}}
$^{\S}$ Due to the lack of public code, we reimplement NODE-MOE based on its paper~\cite{nodewise-3}.
\vskip -0.02in
\end{table*}



\section{Empirical Studies}
\label{section-exp}

To validate the superiority and effectiveness of the CPF method, we conduct extensive experiments to answer the following questions:

\begin{itemize}[leftmargin=*,parsep=0pt,itemsep=0pt,topsep=1pt]
    \item {\bf RQ1:} How does CPF perform on benchmark node classification tasks compared to state-of-the-art methods?
    \item {\bf RQ2:} How do the two partition-wise filtering modules enhance CPF’s performance?
    \item {\bf RQ3:} How robust is CPF to variations in hyperparameters?
    \item {\bf RQ4:} Can CPF effectively generalize to real-world applications?
\end{itemize}

\subsection{Benchmark Node Classification Tasks (RQ1)}
\label{section-exp-node-classify}


\subsubsection{\bf Datasets and baselines}\

\subsubsection*{\bf Datasets} We select $13$ datasets with varied sizes and levels of heterophily. 
The homophilic datasets comprise citation networks (Cora, CiteSeer, PubMed)\cite{dataset1-cora} and large OGB graphs (ogbn-Arxiv, ogbn-Products, ogbn-Papers100M)\cite{dataset5-ogb}. 
For heterophilic datasets, we select three latest graphs (Roman-empire, Amazon-ratings, Questions)~\cite{dataset8-small-hetero} and four large-scale ones (Gamers, Genius, Snap-patent, Pokec)~\cite{dataset6-large-hetero}. 
(Due to the known data-leakage issues~\cite{dataset8-small-hetero}, we intentionally avoid the routine datasets in \cite{dataset3-pei}.) 

\subsubsection*{\bf Baselines and settings} We include $18$ advanced baseline methods tailored for both heterophilic and homophilic contexts, which can be grouped into four types as below:
\begin{itemize}[leftmargin=*,parsep=0pt,itemsep=0pt,topsep=1pt]
\item {\bf Non-filtering:} H2GCN~\cite{H2GCN}, GLOGNN~\cite{glognn++}, LINKX~\cite{dataset6-large-hetero}, OrderGNN~\cite{OrderedGNN}.
\item {\bf Non-decoupled Filtering:} GCN~\cite{GCN}, GCNII~\cite{gcnii}, ChebNet~\cite{ChebNet}, ACMGCN~\cite{ACMGCN}, Specformer~\cite{specformer}. 
\item {\bf Graph-wise Filtering:} GPRGNN~\cite{GPRGNN}, ChebNetII~\cite{ChebNetII}, OptBasis~\cite{OptBasisGNN}, JacobiConv~\cite{JacobiConv}, AdaptKry~\cite{decoupled-AdaptKry}, UniFilter~\cite{decoupled-UniFilter}.
\item {\bf Node-wise Filtering:} NIGCN~\cite{nodewise-1}, NODE-MOE~\cite{nodewise-3}, NFGNN~\cite{decoupled-NFGNN-nodewise}.
\end{itemize}
For the widely adopted baselines (GCN and ChebNet), we adopt consistent implementations drawn from prior research~\cite{BernNet-GNN-narrowbandresults-1,ChebNetII,chebnet2d,OptBasisGNN,JacobiConv,decoupled-FEGNN,decoupled-AdaptKry,decoupled-NFGNN-nodewise}. 
For the remaining baselines, we apply the hyperparameter tuning procedures outlined in their original papers.

\subsubsection*{\bf Implementation of CPF} For all experiments, we implement $\mathbf{T}_{k}$ using a Chebyshev basis~\cite{poly_chebyshev} and set the maximum polynomial degree to $K=10$ to maintain consistency with widely used filtering-based methods, ensuring a fair comparison. Moreover, we fix the coarsening ratio at $r=0.5$ across all experiments for uniformity. 
More specifics are available in Appendix~\ref{appendix:exp-details-node-classify}. 


\begin{table*}[!th]
\caption{Node classification results and total runtime (in hours) over $10$ full runs on exceptionally large graphs. 
CPF’s runtime includes graph coarsening time, which is also reported in brackets.
``OOM'' denotes an ``Out-Of-Memory'' failure. }
\vskip -0.05in
\label{table-node-classify-large}
\centering
\begin{tabular}{ccccccccc}
\hline
\multirow{2}{*}{Method} & \multicolumn{2}{c}{Products}            & \multicolumn{2}{c}{Papers100M}          & \multicolumn{2}{c}{Snap}                & \multicolumn{2}{c}{Pokec}               \\ \cline{2-9} 
                        & Test acc                      & Runtime & Test acc                      & Runtime & Test acc                      & Runtime & Test acc                      & Runtime \\ \hline
GCN                     & $76.37_{\pm0.2}$              & $12.2$   & OOM                           & -       & $46.66_{\pm0.1}$              & $19.8$   & $74.78_{\pm0.2}$              & $15.3$   \\
GPRGNN                  & $79.45_{\pm0.1}$              & $13.5$   & $66.13_{\pm0.2}$              & $113.5$  & $48.88_{\pm0.2}$              & $22.7$   & $79.55_{\pm0.3}$              & $16.7$   \\
ChebNetII               & $81.66_{\pm0.3}$              & $13.7$   & $\underline{67.11_{\pm0.2}}$  & $118.8$  & $51.74_{\pm0.2}$              & $23.0$   & $81.88_{\pm0.3}$              & $16.8$   \\
OptBasis                & $81.33_{\pm0.2}$              & $13.8$   & $67.03_{\pm0.3}$              & $121.7$  & $53.55_{\pm0.1}$              & $22.8$   & $82.09_{\pm0.3}$              & $16.8$   \\
AdaptKry                & $\underline{81.70_{\pm0.3}}$  & $14.5$   & $67.07_{\pm0.2}$              & $122.5$  & $55.92_{\pm0.2}$              & $23.6$   & $82.16_{\pm0.2}$              & $17.8$   \\
UniFilter               & $80.33_{\pm0.2}$              & $14.2$   & $66.79_{\pm0.3}$              & $117.8$  & $52.06_{\pm0.1}$              & $23.1$   & $82.23_{\pm0.3}$  & $16.6$   \\ \hline
NIGCN                   & $80.62_{\pm0.2}$              & $13.7$   & $66.57_{\pm0.2}$              & $118.5$  & $57.26_{\pm0.3}$  & $22.9$   & $\underline{82.33_{\pm0.4}}$              & $16.4$   \\
NODE-MOE$^{\S}$         & $80.57_{\pm0.3}$              & $15.1$   & $66.12_{\pm0.2}$              & $125.6$  & $56.36_{\pm0.4}$  & $25.7$   & $81.72_{\pm0.3}$              & $18.3$   \\
NFGNN                   & $81.11_{\pm0.2}$              & $14.1$   & $66.38_{\pm0.2}$              & $122.7$  & $\underline{57.83_{\pm0.3}}$  & $22.9$   & $82.16_{\pm0.3}$              & $17.1$   \\ \hline
CPF (\textbf{Ours})           & \textbf{83.87}$\boldsymbol{_{\pm0.2}}$ & $14.2 (0.9)$   & \textbf{68.87}$\boldsymbol{_{\pm0.3}}$ & $122.7 (7.3)$  & \textbf{64.70}$\boldsymbol{_{\pm0.3}}$ & $23.9 (1.8)$   & \textbf{85.95}$\boldsymbol{_{\pm0.2}}$ & $17.4 (1.3)$   \\
\#Improv.               & $2.17\%$   &   -        & $1.76\%$     &    -     & $6.87\%$    &    -      & $3.62\%$     &     -    \\ \hline
\end{tabular}
\vskip -0.02in
\end{table*}


\subsubsection{\bf Main results and discussions}
\label{section-exp-node-classify-main-results}\

This section provides an analysis of the results presented in Table~\ref{table-node-classify-medium} and~\ref{table-node-classify-large}. 
More exhaustive results are accessible in Appendix~\ref{appendix:additional-results-node-classification}. 

\subsubsection*{\bf Node-wise vs. Graph-wise} As shown in both tables, node-wise methods tend to outperform graph-wise methods on heterophilic datasets, while graph-wise methods prove more effective on homophilic datasets. 
This observation aligns with the discussions in Section~\ref{section-CPF-motivation}. 
In heterophilic graphs, node-wise methods excel by using node-specific filters, enabling them to handle diverse node patterns, whereas graph-wise methods lack an essential adaptability. 
In homophilic graphs, however, the consistency across nodes allows graph-wise filtering to perform adequately, whereas node-wise methods may introduce redundancy and a risk of overfitting.

\subsubsection*{\bf Comparative effectiveness of CPF} Our proposed CPF delivers notable improvements across both heterophilic and homophilic graphs. 
On heterophilic graphs, CPF outshines node-wise methods, with a maximum improvement of \textbf{6.87}\% on Snap over the nearest competitor, NFGNN. 
Likewise, on homophilic graphs, CPF outperforms graph-wise methods by up to \textbf{2.59}\% on ogbn-Arxiv compared to UniFiler, the closest contender. 
These improvements highlight the effectiveness of partition-wise filtering over both graph and node-wise paradigms.


\begin{table*}[!t]
\caption{Performance gains from the proposed structure-aware and feature-aware partition-wise filtering.}
\vskip -0.05in
\label{table-effectiveness-partition-filtering}
\centering
\begin{tabular}{cc|cccc|cccc}
\hline
Structure-aware  & Feature-aware  & Cora                                   & Cite.                                  & Arxiv                                  & Products                               & Roman.                                 & Amazon.                                & Gamers                                 & Genius                                 \\ \hline
                          &                         & $88.21_{\pm0.6}$                       & $76.18_{\pm1.3}$                       & $72.05_{\pm0.3}$                       & $81.48_{\pm0.3}$                       & $66.45_{\pm1.1}$                       & $42.34_{\pm0.7}$                       & $66.40_{\pm0.4}$                       & $90.63_{\pm0.5}$                       \\
                          & $\surd$                 & $88.58_{\pm0.7}$                       & $76.42_{\pm1.1}$                       & $72.86_{\pm0.3}$                       & $81.68_{\pm0.3}$                       & $67.53_{\pm1.1}$                       & $43.12_{\pm0.7}$                       & $66.83_{\pm0.3}$                       & $91.12_{\pm0.5}$                       \\
$\surd$                   &                         & $88.91_{\pm0.6}$                       & $77.07_{\pm1.1}$                       & $74.22_{\pm0.3}$                       & $82.10_{\pm0.3}$                       & $72.18_{\pm1.0}$                       & $44.52_{\pm0.6}$                       & $67.42_{\pm0.4}$                       & $91.65_{\pm0.4}$                       \\
$\surd$                   & $\surd$                 & \textbf{89.28}$\boldsymbol{_{\pm0.7}}$ & \textbf{77.55}$\boldsymbol{_{\pm0.9}}$ & \textbf{75.46}$\boldsymbol{_{\pm0.3}}$ & \textbf{83.87}$\boldsymbol{_{\pm0.2}}$ & \textbf{74.02}$\boldsymbol{_{\pm1.1}}$ & \textbf{46.12}$\boldsymbol{_{\pm0.7}}$ & \textbf{69.77}$\boldsymbol{_{\pm0.3}}$ & \textbf{92.68}$\boldsymbol{_{\pm0.3}}$ \\ \hline
\end{tabular}
\end{table*}


\subsubsection*{\bf Scalability and efficiency} Table~\ref{table-node-classify-large} presents a comparative analysis of CPF and filtering-based baselines, where all models adopt a decoupled architecture to enhance scalability and efficiency on large-scale graphs. 
We observe that CPF outperforms all baselines by substantial margins across all datasets while achieving efficiency levels on par with the best baselines. 
Although CPF needs a preprocessing step for graph coarsening, this step is performed only once and can be reused across multiple experiments. 
Given the low time complexity of the coarsening algorithm, CPF incurs minimal computational overhead, validating its scalability and efficiency.

\subsection{Unveiling the Impact of Partition-wise Filtering (RQ2)}
\label{section-exp-effectiveness-partion-filtering}

This section examines the effectiveness of the two partition-wise filtering steps: structure-aware and feature-aware, introduced in Section~\ref{section-CPF}. 
The results are provided in Table~\ref{table-effectiveness-partition-filtering}, and we offer an analysis of these findings below.

\subsubsection*{\bf Validation of individual and combined gains} As shown in Table~\ref{table-effectiveness-partition-filtering}, both structure-aware and feature-aware filtering enhance the overall model performance across all datasets, validating the individual efficacy of each step. 
More notably, we observe a surprising synergistic effect when both steps are combined in the model, namely the proposed CPF method. 
The CPF outperforms the individual filtering steps by a substantial margin, with performance improvements that far exceed the sum of the individual contributions. 
This finding underscores that the CPF model not only introduces two potent components but also amplifies their combined effect, resulting in a remarkable boost in performance.

\subsubsection*{\bf Relative significance of structure and feature filtering} The results demonstrate that structure-aware filtering contributes more significantly to performance gains than feature-aware filtering. 
This emphasizes the higher priority of the structural filtering aspect over the feature-based one, suggesting that further CPF research should primarily target advancements in structure-aware filtering.

\subsection{Ablation Studies (RQ3)}
\label{section-exp-ablation}

This section explores how CPF responds to hyperparameter variations, 
presenting ablation results in Figure~\ref{fig:ablation-r}, and Tables~\ref{table-node-classify-ablation-coarsening}, ~\ref{table-ablation-polynomial-basis}, and~\ref{table-ablation-maximum-degree}. 
Additional results are accessible in Appendix~\ref{appendix:additional-results-node-classification}.


\begin{table}[!th]
\caption{Ablation studies on coarsening algorithms. 
``Origin'' denotes the proposed CPF implementation using LV~\cite{rsa-Gcoarse-3-Local-Variation}.}
\vskip -0.05in
\label{table-node-classify-ablation-coarsening}
\centering
\begin{tabular}{ccccc}
\hline
Dataset & Origin (LV)                   & MGC~\cite{rsa-Gcoarse-4-MGC-SGC}                          & SGC~\cite{rsa-Gcoarse-4-MGC-SGC}                          & FGC~\cite{rsa-Gcoarse-8-FGC}                          \\ \hline
Cora    & $89.28_{\pm0.7}$              & $\underline{89.78_{\pm0.4}}$  & \textbf{90.12}$\boldsymbol{_{\pm0.3}}$ & $89.22_{\pm0.4}$              \\
Arxiv   & $\underline{75.46_{\pm0.3}}$  & $75.22_{\pm0.2}$              & $75.04_{\pm0.2}$              & \textbf{76.47}$\boldsymbol{_{\pm0.2}}$ \\
Roman.  & $74.02_{\pm1.1}$              & $\underline{74.76_{\pm1.2}}$  & $74.53_{\pm1.5}$              & \textbf{77.81}$\boldsymbol{_{\pm1.3}}$ \\
Genius  & $\underline{92.68_{\pm0.3}}$  & $92.37_{\pm0.3}$              & $\underline{92.60_{\pm0.2}}$  & \textbf{93.16}$\boldsymbol{_{\pm0.3}}$ \\ \hline
\end{tabular}
\vskip -0.05in
\end{table}


\subsubsection*{\bf Adaptability to diverse coarsening algorithms} Beyond the primary experiments with Local Variation (LV) algorithm~\cite{rsa-Gcoarse-3-Local-Variation}, we explore more coarsening methods, such as MGC~\cite{rsa-Gcoarse-4-MGC-SGC}, SGC~\cite{SGC} and FGC~\cite{rsa-Gcoarse-8-FGC}, to test CPF’s integration flexibility. 
Tables~\ref{table-node-classify-ablation-coarsening} and~\ref{table-additional-coarsening} reveal that the advanced methods consistently enhance CPF’s performance, highlighting its compatibility across coarsening algorithms. 
Moreover, FGC, a coarsening algorithm using both graph structure and node feature, shows considerable advancements over structure-only methods LV, MGC and SGC, indicating potential of improving CPF by combining richer information during coarsening.

\subsubsection*{\bf Impact of polynomial bases and degrees} Tables~\ref{table-ablation-polynomial-basis}, ~\ref{table-ablation-maximum-degree}, and~\ref{table-additional-polynomial-basis} illustrate that both polynomial bases and degrees evidently affect the outcomes.  
Chebyshev and Jacobian bases, known for their orthogonality and validated efficacy~\cite{ChebNetII,JacobiConv}, show notably strong performance. 
Further, increasing the degree generally improves accuracy, but only up to a certain point, after which gains plateau or even decline. 
This behavior aligns with the approximation theory, where higher-degree polynomials reduce approximation loss until diminishing returns set in~\cite{polyapprox_1,polyapprox_2}. 
The empirical results resonate with prior research on basis and degree selection in filtering-based GNNs~\cite{ChebNetII,JacobiConv,OptBasisGNN,decoupled-AdaptKry,decoupled-TFGNN}, showing CPF’s consistency in this domain.


\begin{table}[!h]
\caption{Ablation studies on polynomial bases, comparing three variants: Monomial, Bernstein, and Jacobian.}
\vskip -0.05in
\label{table-ablation-polynomial-basis}
\centering
\begin{tabular}{ccccc}
\hline
Basis         & Pubmed                        & Arxiv                         & Ques.                         & Gamers                        \\ \hline
Origin (Cheb)  & $\underline{89.93_{\pm0.5}}$  & \textbf{75.46}$\boldsymbol{_{\pm0.3}}$ & \textbf{78.80}$\boldsymbol{_{\pm1.1}}$ & \textbf{69.77}$\boldsymbol{_{\pm0.3}}$ \\ \hline
Mono          & $89.33_{\pm0.2}$              & $74.28_{\pm0.4}$              & $76.36_{\pm1.1}$              & $68.10_{\pm0.4}$              \\
Bern         & $89.51_{\pm0.2}$              & $74.65_{\pm0.2}$              & $76.02_{\pm1.0}$              & $67.63_{\pm0.4}$              \\
Jacobi          & \textbf{90.39}$\boldsymbol{_{\pm0.3}}$ & $\underline{75.10_{\pm0.2}}$  & $\underline{78.33_{\pm1.0}}$  & $\underline{69.19_{\pm0.3}}$  \\ \hline
\end{tabular}
\vskip -0.05in
\end{table}
\begin{table}[!h]
\caption{Ablation studies on maximum polynomial degree.}
\vskip -0.05in
\label{table-ablation-maximum-degree}
\centering
\setlength{\tabcolsep}{4pt}
\resizebox{\linewidth}{!}{
\begin{tabular}{cccccc}
\hline
Degree & $5$                & $10$               & $15$               & $20$               & $25$               \\ \hline
Arxiv  & $74.39_{\pm0.2}$ & $75.46_{\pm0.3}$ & $75.10_{\pm0.2}$ & $75.39_{\pm0.3}$ & $75.41_{\pm0.3}$ \\
Genius & $91.33_{\pm0.3}$ & $92.68_{\pm0.3}$ & $92.71_{\pm0.2}$ & $92.92_{\pm0.2}$ & $92.80_{\pm0.2}$ \\ \hline
\end{tabular}}
\vskip -0.05in
\end{table}



\begin{table*}[!th]
\caption{Graph anomaly detection results. }
\vskip -0.05in
\label{table-GAnoDet}
\centering
\begin{tabular}{cc|cc|cc|cc}
\hline
\multirow{2}{*}{Type}                               & Dateset                                           & \multicolumn{2}{c|}{YelpChi ($1\%$)}              & \multicolumn{2}{c|}{Amazon ($1\%$)}               & \multicolumn{2}{c}{T-Finance ($1\%$)}             \\
                                                      & Metric                                            & F1-macro                & AUROC                   & F1-macro                & AUROC                   & F1-macro                & AUROC                   \\ \hline
\multirow{3}{*}{GAD Models}                           & PC-GNN                                            & $60.55$                 & $75.29$                 & $82.62$                 & $\underline{91.61}$     & $83.40$                 & \textbf{91.85}    \\
                                                      & CARE-GNN                                          & $61.68$                 & $73.95$                 & $75.78$                 & $88.79$                 & $\underline{86.03}$     & $91.17$                 \\
                                                      & GDN                                               & $65.72$                 & $75.33$                 & $\underline{90.49}$     & \textbf{92.07}    & $77.38$                 & $89.42$                 \\ \hline
\multirow{2}{*}{\makecell[c]{GAD-specialized \\ Filtering-based GNNs}} & BWGNN                                             & \textbf{66.52}   & $\underline{77.23}$     & $90.28$                 & $89.19$                 & $85.56$                 & $91.38$                 \\
                                                      & GHRN                                              & $62.77$                 & $74.64$                 & $86.65$                 & $87.09$                 & $80.70$                 & $91.55$                 \\ \hline
\multirow{6}{*}{\makecell[c]{General-purpose \\ Filtering-based GNNs}}       & GCN                                               & $50.66$                 & $54.31$                 & $69.79$                 & $85.18$                 & $75.26$                 & $87.05$                 \\
                                                      & GPRGNN                                            & $60.45$                 & $67.44$                 & $83.71$                 & $85.28$                 & $77.53$                 & $85.69$                 \\
                                                      & OptBasis                                          & $62.03$                 & $68.32$                 & $86.12$                 & $85.02$                 & $79.28$                 & $86.22$                 \\
                                                      & AdaptKry                                          & $63.40$                 & $66.18$                 & $83.30$                 & $84.58$                 & $80.67$                 & $85.41$                 \\
                                                      & NIGCN                                          & $61.48$                 & $66.23$                 & $84.79$                 & $85.75$                 & $81.79$                 & $86.04$                 \\
                                                      & NFGNN                                             & $60.66$                 & $67.36$                 & $85.61$                 & $86.88$                 & $82.38$                 & $86.59$                 \\ \hline
\multirow{2}{*}{\textbf{Ours}}    & CPF         & $\underline{66.13}$     & \textbf{79.66}    &  \textbf{91.54}
                                                                    & $90.26$     & \textbf{87.39}   & $\underline{91.72}$     \\
                                                      & \#Improv.$^{\ddag}$ & $2.73\%$ & $11.34\%$ & $5.42\%$ & $3.38\%$ & $5.01\%$ & $4.67\%$ \\ \hline
\end{tabular}

$^{\ddag}$ Improvements are relative to general-purpose methods rather than GAD baselines.
\end{table*}
\begin{figure}[!th]
\centering
    \subfloat[Arxiv.]{
    \label{fig:r-pubmed}
    \includegraphics[width=0.45\linewidth]{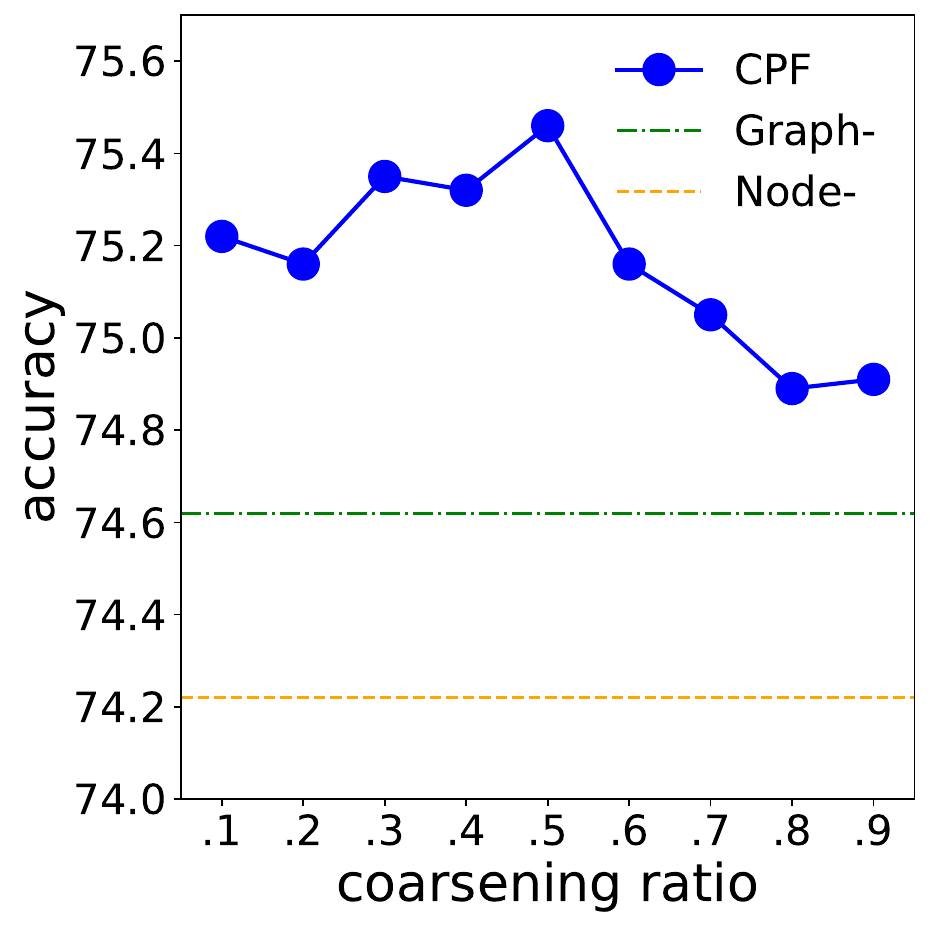}
    }
    \subfloat[Gamers.]{
    \label{fig:r-amazon}
    \includegraphics[width=0.45\linewidth]{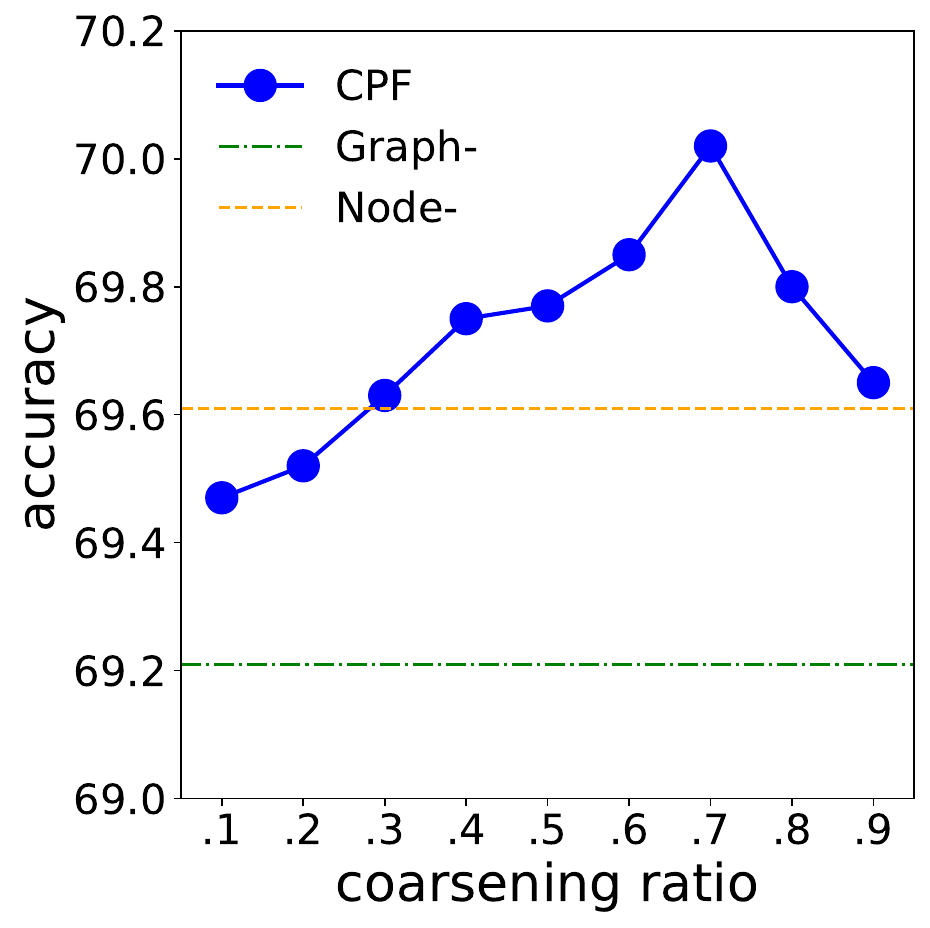}
    }
    \caption{The impact of coarsening ratio. Here, 
    ``Graph-'' and ``Node-'' represent graph-wise and node-wise filtering.}
    \label{fig:ablation-r}
\end{figure}


\subsubsection*{\bf Robustness of CPF to the coarsening ratio $r$} Figures~\ref{fig:ablation-r} and~\ref{fig:additional-ablation-r} reveal that CPF reaches peak performance when $r$ is in the mid-range of $\left(0, {(n-1)}/{n}\right)$, consistently outperforming both graph-wise and node-wise filtering across a broad range of $r$. 
The results confirm CPF's robustness to the $r$, suggesting a reduced reliance on extensive hyperparameter tuning in practical applications.

Moreover, the results align with the analysis in Section~\ref{section-CPF-motivation}. 
For heterophilic graphs, CPF achieves better performance at small $r$ (closer to node-wise filtering), while for homophilic graphs, it excels at larger $r$ (closer to graph-wise filtering). 
This further highlights the critical role of partition-wise filtering in combining these strategies to optimize GNN performance.

\subsection{Application Example: Graph Anomaly Detection (RQ4)}
\label{section-exp-graph-anomaly-detection}

To highlight the practical benefits of CPF, this section explores its application to the graph anomaly detection (GAD) task, a standard binary node classification problem (normal vs. abnormal)~\cite{GAnoDet-survey-1, GAnoDet-survey-2}.

\subsubsection{\bf Datasets and baselines}

\subsubsection*{\bf Datasets} We adopt three datasets (YelpChi, Amazon, and T-Finance) with a uniform low label-rate of $1\%$ following~\cite{GAnoDet-spectral-1-BWGNN}.

\subsubsection*{\bf Baselines and model implementations} We include $11$ baseline methods, organized into three types below:
\begin{itemize}[leftmargin=*,parsep=0pt,itemsep=0pt,topsep=1pt]
\item {\bf GAD models:} PC-GNN~\cite{GAnoDet-2-PC-GNN}, CARE-GNN~\cite{GAnoDet-1-CARE-GNN}, GDN~\cite{GAnoDet-3-GDN}.
\item {\bf GAD-specialized Filtering GNNs:} BWGNN~\cite{GAnoDet-spectral-1-BWGNN}, GHRN~\cite{GAnoDet-spectral-2-GHRN}.
\item {\bf General-purpose Filtering GNNs:} GCN~\cite{GCN}, GPRGNN~\cite{GPRGNN}, OptBasis~\cite{OptBasisGNN}, AdaptKry~\cite{decoupled-AdaptKry}, NIGCN~\cite{nodewise-1}, NFGNN~\cite{decoupled-NFGNN-nodewise}.
\end{itemize}
The implementations for the common baselines (PC-GNN, CARE-GNN, BWGNN, and GCN) are adopted from~\cite{GAnoDet-spectral-1-BWGNN}. 
For the feature transformation module in other general-purpose methods, we employ a two-layer MLP with $64$ hidden units, ensuring consistency with the GAD-specialized models. 
The remaining baselines adhere to their original configurations from the respective publications.

\subsubsection*{\bf Implementation of CPF} Our CPF is implemented using a two-layer MLP with 64 hidden units, aligning with the general-purpose methods. 
Hyperparameters are set following the process described in Section~\ref{section-exp-node-classify}, with further details provided in Appendix~\ref{appendix:exp-details-graph-anomaly-detection}.

\subsubsection{\bf Main results and discussions}
\label{section-exp-graph-anomaly-detection-main-results}\ 


\subsubsection*{\bf Enhanced utility in real-world application} 
In Table~\ref{table-GAnoDet}, the \#Improv. metric reveals that CPF achieves substantial improvements over general-purpose models, notably with up to $11.34\%$ increases on YelpChi. 
We conclude that although general-purpose methods yield favorable results in benchmark node classification tasks, they usually fall short in specific applications. 
In this regard, CPF, equipped with the partition-wise filtering, consistently delivers substantial enhancements in both standard and specific real-world tasks, confirming its effectiveness and practical utility.

\subsubsection*{\bf Comparable efficacy to GAD-specialized methods} Table~\ref{table-GAnoDet} further reveals that CPF achieves performance comparable to GAD-specialized filtering GNNs. 
Those GAD-specialized models (such as BWGNN and GHRN), while rooted in the same graph filtering concept, are designed with more specific features to drive performance improvements, thus outperforming general-purpose techniques. 
For instance, BWGNN~\cite{GAnoDet-spectral-1-BWGNN} mitigates the “right-shift” phenomenon using tailored beta wavelet filters, and GHRN~\cite{GAnoDet-spectral-2-GHRN} filters out high-frequency components to prune inter-class edges in heterophilic graphs. 
In contrast, CPF proposes an alternative and effective approach for GAD problems through an advanced filtering strategy, demonstrating impressive results. 
This result reflects a fresh direction in improving the practical performance of filtering-based GNNs via a more fundamental aspect of GNN paradigms.


\section{Related Works}
\label{section:related-works}

This section reviews prior work relevant to our study, covering filtering-based Graph Neural Networks, spectral graph coarsening techniques, and node classification on heterophilic graphs using GNNs. 
These areas provide the foundation and context for our proposed approach.

\subsection{Filtering-based Graph Neural Networks}
\label{section:related-works-filtering-gnn}

This paper explores a distinct branch of GNNs referred to as filtering-based GNNs, which apply graph filters to leverage the graph spectrum for processing graph-structured data. 
Filtering-based GNNs typically employ filters that can be characterized through matrix polynomials with parameterized coefficients. 
For example, GPRGNN~\cite{GPRGNN} introduces a monomial-based graph filter, interpreted as a generalized PageRank algorithm. 
ChebNet~\cite{ChebNet} uses Chebyshev polynomials to construct graph filters, with subsequent works~\cite{ChebNetII,chebnet2d} further refining this approach. 
JacobiConv~\cite{JacobiConv} unifies different methods by employing Jacobian polynomials.

A recent innovation in filtering-based GNNs lies in the deployment of graph filters, leading to the introduction of node-wise filtering. 
As opposed to the aforementioned graph-wise methods that apply a uniform filter across all nodes, node-wise methods, such as~\cite{decoupled-NFGNN-nodewise,nodewise-1,nodewise-3}, allocate distinct filters for each node. 

However, despite the fact that graph-wise filtering is a special case of node-wise filtering, the latter typically performs on par with the former in a wider range of tasks. 
In addition, lack of a unified framework that integrates both strategies limits the broader understanding of their respective benefits and impedes the progress toward more advanced methods.

\subsection{Graph Coarsening}
\label{section:related-works-graph-coarsening}
Graph coarsening, recognized as a graph summarization (also referred to as graph reduction) technique~\cite{Gcoarse_survey}, involves systematically merging nodes and/or edges of a graph to produce a simplified, coarser version, which has been extensively researched and applied across numerous real-world problems associated with graph-structured data~\cite{Gcoarse_survey-2,Gcoarse-infan-1,Gcoarse-infan-2,Gcoarse-co,Gcoarse-gl-1,Gcoarse-gl-2,Gcoarse-gl-3,Gcoarse-gl-4,Gcoarse-visualize}. 

Graph coarsening algorithms typically focus on minimizing spectral differences between the original and coarsened graphs, with various measures proposed to quantify these differences and guide the coarsening process. 
For instance, \cite{Gcoarse-1-algebraic-distance}~advocates coarsening via algebraic distance, ensuring that the spectral properties of the original and coarsened graphs remain consistent. 
\cite{rsa-Gcoarse-3-Local-Variation}~develops the restricted spectral approximation approach, which aims at preserving the spectral properties within a restricted eigenspace. 
\cite{rsa-Gcoarse-4-MGC-SGC}~produces coarsened graphs with minimizing the spectral distance regarding the perturbation of eigenvalues.

\subsection{Node Classification on Heterophilic Graphs utilizing GNNs}
\label{section:related-works-node-classify-heterophily}
In recent years, heterophilic graphs have garnered substantial interest within the graph learning field. 
In contrast to homophilic graphs, where connected nodes typically share the same label, heterophilic graphs connect nodes with dissimilar labels. 
This unique structure presents significant challenges for GNNs~\cite{heterophily-gnn-survey,heterophily-gnn-survey-2,heterophily-gnn-survey-3}, which are typically designed for homophilic settings. 
To address these challenges, a range of GNNs tailored to heterophily have emerged. 
H2GCN~\cite{H2GCN} introduces mechanisms tailored for embedding nodes in heterophilic settings, OrderGNN~\cite{OrderedGNN} reconfigures message-passing to better capture heterophilic relationships, and LRGNN~\cite{low-rank-gnn-2} employs a global label relationship matrix to boost performance in heterophilic scenarios.

\subsubsection*{\bf Addressing heterophily with graph filtering} Filtering-based GNNs have recently shown promise in overcoming challenges by creating dataset-specific filters that extend beyond conventional low-pass approaches. 
By adapting to the characteristics of heterophilic graphs, these GNNs achieve notable improvements in node classification under heterophily. 
Alongside the approaches summarized in Section~\ref{section:related-works-filtering-gnn}, ACMGCN~\cite{ACMGCN} incorporates an adaptive channel-mixing mechanism to combine various filter channels, boosting classification performance; 
AdaptKry~\cite{decoupled-AdaptKry} applies Krylov subspace theory to construct advanced filters, resulting in more effective filtering processes; 
TFGNN~\cite{decoupled-TFGNN} investigates the polynomial selection problem and proposes trigonometric graph filters, yielding more robust and powerful filtering-based GNNs.


\section{Conclusions}
\label{section-conclusion}

In this paper, we revisit existing filtering paradigms, graph-wise and node-wise, through the lens of the Contextual Stochastic Block Model, revealing their limitations in handling real-world graphs that blend homophily and heterophily. 
To address the drawbacks, we proposed CPF, a novel filtering paradigm that operates instead on node partitions. 
CPF integrates structure-aware partition-wise filtering, which generates node embeddings within partitions obtained via graph coarsening, and feature-aware partition-wise filtering, which refines embeddings via filtering on clusters derived from $k$-means clustering over features. 
Each phase of CPF is accompanied by fine-grained analysis, confirming its superiority over the counterparts. 
Through extensive experiments on benchmark node classification tasks and a real-world graph anomaly detection application, we validate CPF’s effectiveness and practical utility.

\subsubsection*{\bf Limitations and future work} CPF currently disentangles 
and sequentially combines 
the structure and feature filtering steps, but it lacks a deeper, more intuitive theoretical framework for this integration. 
Therefore, a promising direction for future work would be to bridge this gap, offering a more unified and grounded approach to filtering that draws stronger connections between structure and feature aspects.


\section*{Acknowledgment}
\label{section-ack}

The authors acknowledge funding from Research Grants Council (RGC) under grant \texttt{22303424} and GuangDong Basic and Applied Basic Research Foundation under grant \texttt{2025A1515010259}.

\clearpage
\bibliographystyle{ACM-Reference-Format}
\balance
\bibliography{references}


\begin{thebibliography}{91}


\ifx \showCODEN    \undefined \def \showCODEN     #1{\unskip}     \fi
\ifx \showISBNx    \undefined \def \showISBNx     #1{\unskip}     \fi
\ifx \showISBNxiii \undefined \def \showISBNxiii  #1{\unskip}     \fi
\ifx \showISSN     \undefined \def \showISSN      #1{\unskip}     \fi
\ifx \showLCCN     \undefined \def \showLCCN      #1{\unskip}     \fi
\ifx \shownote     \undefined \def \shownote      #1{#1}          \fi
\ifx \showarticletitle \undefined \def \showarticletitle #1{#1}   \fi
\ifx \showURL      \undefined \def \showURL       {\relax}        \fi
\providecommand\bibfield[2]{#2}
\providecommand\bibinfo[2]{#2}
\providecommand\natexlab[1]{#1}
\providecommand\showeprint[2][]{arXiv:#2}

\bibitem[Ben-David et~al\mbox{.}(2007)]%
        {kmeans-stability}
\bibfield{author}{\bibinfo{person}{Shai Ben-David}, \bibinfo{person}{D{\'a}vid P{\'a}l}, {and} \bibinfo{person}{Hans~Ulrich Simon}.} \bibinfo{year}{2007}\natexlab{}.
\newblock \showarticletitle{Stability of k-Means Clustering}. In \bibinfo{booktitle}{\emph{Learning Theory}}, \bibfield{editor}{\bibinfo{person}{Nader~H. Bshouty} {and} \bibinfo{person}{Claudio Gentile}} (Eds.). \bibinfo{publisher}{Springer Berlin Heidelberg}, \bibinfo{address}{Berlin, Heidelberg}, \bibinfo{pages}{20--34}.
\newblock
\showISBNx{978-3-540-72927-3}


\bibitem[Bianchi et~al\mbox{.}(2020)]%
        {ARMA-RationalGNN}
\bibfield{author}{\bibinfo{person}{Filippo~Maria Bianchi}, \bibinfo{person}{Daniele Grattarola}, \bibinfo{person}{Lorenzo Livi}, {and} \bibinfo{person}{Cesare Alippi}.} \bibinfo{year}{2020}\natexlab{}.
\newblock \showarticletitle{Graph Neural Networks With Convolutional ARMA Filters}.
\newblock \bibinfo{journal}{\emph{IEEE Transactions on Pattern Analysis and Machine Intelligence}} \bibinfo{volume}{44}, \bibinfo{number}{7} (\bibinfo{year}{2020}), \bibinfo{pages}{3496--3507}.
\newblock
\href{https://doi.org/10.1109/TPAMI.2021.3054830}{doi:\nolinkurl{10.1109/TPAMI.2021.3054830}}


\bibitem[Bo et~al\mbox{.}(2023)]%
        {specformer}
\bibfield{author}{\bibinfo{person}{Deyu Bo}, \bibinfo{person}{Chuan Shi}, \bibinfo{person}{Lele Wang}, {and} \bibinfo{person}{Renjie Liao}.} \bibinfo{year}{2023}\natexlab{}.
\newblock \showarticletitle{Specformer: Spectral Graph Neural Networks Meet Transformers}. In \bibinfo{booktitle}{\emph{The Eleventh International Conference on Learning Representations}}.
\newblock
\urldef\tempurl%
\url{https://openreview.net/forum?id=0pdSt3oyJa1}
\showURL{%
\tempurl}


\bibitem[Borisyuk et~al\mbox{.}(2024)]%
        {websearch_2}
\bibfield{author}{\bibinfo{person}{Fedor Borisyuk}, \bibinfo{person}{Shihai He}, \bibinfo{person}{Yunbo Ouyang}, \bibinfo{person}{Morteza Ramezani}, \bibinfo{person}{Peng Du}, \bibinfo{person}{Xiaochen Hou}, \bibinfo{person}{Chengming Jiang}, \bibinfo{person}{Nitin Pasumarthy}, \bibinfo{person}{Priya Bannur}, \bibinfo{person}{Birjodh Tiwana}, \bibinfo{person}{Ping Liu}, \bibinfo{person}{Siddharth Dangi}, \bibinfo{person}{Daqi Sun}, \bibinfo{person}{Zhoutao Pei}, \bibinfo{person}{Xiao Shi}, \bibinfo{person}{Sirou Zhu}, \bibinfo{person}{Qianqi Shen}, \bibinfo{person}{Kuang-Hsuan Lee}, \bibinfo{person}{David Stein}, \bibinfo{person}{Baolei Li}, \bibinfo{person}{Haichao Wei}, \bibinfo{person}{Amol Ghoting}, {and} \bibinfo{person}{Souvik Ghosh}.} \bibinfo{year}{2024}\natexlab{}.
\newblock \showarticletitle{LiGNN: Graph Neural Networks at LinkedIn} \emph{(\bibinfo{series}{KDD '24})}. \bibinfo{publisher}{Association for Computing Machinery}, \bibinfo{address}{New York, NY, USA}, \bibinfo{pages}{4793–4803}.
\newblock
\showISBNx{9798400704901}
\href{https://doi.org/10.1145/3637528.3671566}{doi:\nolinkurl{10.1145/3637528.3671566}}


\bibitem[Chen et~al\mbox{.}(2022)]%
        {Gcoarse_survey-2}
\bibfield{author}{\bibinfo{person}{Jie Chen}, \bibinfo{person}{Yousef Saad}, {and} \bibinfo{person}{Zechen Zhang}.} \bibinfo{year}{2022}\natexlab{}.
\newblock \showarticletitle{Graph coarsening: from scientific computing to machine learning}.
\newblock \bibinfo{journal}{\emph{SeMA Journal}} \bibinfo{volume}{79}, \bibinfo{number}{1} (\bibinfo{year}{2022}), \bibinfo{pages}{187--223}.
\newblock


\bibitem[Chen and Safro(2011)]%
        {Gcoarse-1-algebraic-distance}
\bibfield{author}{\bibinfo{person}{Jie Chen} {and} \bibinfo{person}{Ilya Safro}.} \bibinfo{year}{2011}\natexlab{}.
\newblock \showarticletitle{Algebraic Distance on Graphs}.
\newblock \bibinfo{journal}{\emph{SIAM Journal on Scientific Computing}} \bibinfo{volume}{33}, \bibinfo{number}{6} (\bibinfo{year}{2011}), \bibinfo{pages}{3468--3490}.
\newblock
\href{https://doi.org/10.1137/090775087}{doi:\nolinkurl{10.1137/090775087}}
\showeprint{https://doi.org/10.1137/090775087}


\bibitem[Chen et~al\mbox{.}(2020)]%
        {gcnii}
\bibfield{author}{\bibinfo{person}{Ming Chen}, \bibinfo{person}{Zhewei Wei}, \bibinfo{person}{Zengfeng Huang}, \bibinfo{person}{Bolin Ding}, {and} \bibinfo{person}{Yaliang Li}.} \bibinfo{year}{2020}\natexlab{}.
\newblock \showarticletitle{Simple and Deep Graph Convolutional Networks}. In \bibinfo{booktitle}{\emph{Proceedings of the 37th International Conference on Machine Learning}} \emph{(\bibinfo{series}{Proceedings of Machine Learning Research}, Vol.~\bibinfo{volume}{119})}. \bibinfo{publisher}{PMLR}, \bibinfo{pages}{1725--1735}.
\newblock
\urldef\tempurl%
\url{https://proceedings.mlr.press/v119/chen20v.html}
\showURL{%
\tempurl}


\bibitem[Chen et~al\mbox{.}(2023)]%
        {rsa-Gcoarse-6}
\bibfield{author}{\bibinfo{person}{Yifan Chen}, \bibinfo{person}{Rentian Yao}, \bibinfo{person}{Yun Yang}, {and} \bibinfo{person}{Jie Chen}.} \bibinfo{year}{2023}\natexlab{}.
\newblock \showarticletitle{A Gromov-{W}asserstein Geometric View of Spectrum-Preserving Graph Coarsening}. In \bibinfo{booktitle}{\emph{Proceedings of the 40th International Conference on Machine Learning}} \emph{(\bibinfo{series}{Proceedings of Machine Learning Research}, Vol.~\bibinfo{volume}{202})}, \bibfield{editor}{\bibinfo{person}{Andreas Krause}, \bibinfo{person}{Emma Brunskill}, \bibinfo{person}{Kyunghyun Cho}, \bibinfo{person}{Barbara Engelhardt}, \bibinfo{person}{Sivan Sabato}, {and} \bibinfo{person}{Jonathan Scarlett}} (Eds.). \bibinfo{publisher}{PMLR}, \bibinfo{pages}{5257--5281}.
\newblock
\urldef\tempurl%
\url{https://proceedings.mlr.press/v202/chen23ak.html}
\showURL{%
\tempurl}


\bibitem[Chen et~al\mbox{.}(2019)]%
        {social_2}
\bibfield{author}{\bibinfo{person}{Zhengdao Chen}, \bibinfo{person}{Lisha Li}, {and} \bibinfo{person}{Joan Bruna}.} \bibinfo{year}{2019}\natexlab{}.
\newblock \showarticletitle{Supervised Community Detection with Line Graph Neural Networks}. In \bibinfo{booktitle}{\emph{International Conference on Learning Representations}}.
\newblock
\urldef\tempurl%
\url{https://openreview.net/forum?id=H1g0Z3A9Fm}
\showURL{%
\tempurl}


\bibitem[Chien et~al\mbox{.}(2021)]%
        {GPRGNN}
\bibfield{author}{\bibinfo{person}{Eli Chien}, \bibinfo{person}{Jianhao Peng}, \bibinfo{person}{Pan Li}, {and} \bibinfo{person}{Olgica Milenkovic}.} \bibinfo{year}{2021}\natexlab{}.
\newblock \showarticletitle{Adaptive Universal Generalized PageRank Graph Neural Network}. In \bibinfo{booktitle}{\emph{International Conference on Learning Representations}}.
\newblock
\urldef\tempurl%
\url{https://openreview.net/forum?id=n6jl7fLxrP}
\showURL{%
\tempurl}


\bibitem[Chung(1997)]%
        {spectralgraphtheory}
\bibfield{author}{\bibinfo{person}{Fan Chung}.} \bibinfo{year}{1997}\natexlab{}.
\newblock \bibinfo{booktitle}{\emph{{Spectral Graph Theory}}}. Vol.~\bibinfo{volume}{92}.
\newblock \bibinfo{publisher}{CBMS Regional Conference Series in Mathematics}.
\newblock
\showISBNx{978-0-8218-0315-8}
\href{https://doi.org//10.1090/cbms/092}{doi:\nolinkurl{/10.1090/cbms/092}}


\bibitem[Defferrard et~al\mbox{.}(2016)]%
        {ChebNet}
\bibfield{author}{\bibinfo{person}{Michaël Defferrard}, \bibinfo{person}{Xavier Bresson}, {and} \bibinfo{person}{Pierre Vandergheynst}.} \bibinfo{year}{2016}\natexlab{}.
\newblock \showarticletitle{Convolutional Neural Networks on Graphs with Fast Localized Spectral Filtering}. In \bibinfo{booktitle}{\emph{Advances in Neural Information Processing Systems}}, \bibfield{editor}{\bibinfo{person}{D.~Lee}, \bibinfo{person}{M.~Sugiyama}, \bibinfo{person}{U.~Luxburg}, \bibinfo{person}{I.~Guyon}, {and} \bibinfo{person}{R.~Garnett}} (Eds.), Vol.~\bibinfo{volume}{29}. \bibinfo{publisher}{Curran Associates, Inc.}
\newblock


\bibitem[Deng et~al\mbox{.}(2020)]%
        {Gcoarse-gl-1}
\bibfield{author}{\bibinfo{person}{Chenhui Deng}, \bibinfo{person}{Zhiqiang Zhao}, \bibinfo{person}{Yongyu Wang}, \bibinfo{person}{Zhiru Zhang}, {and} \bibinfo{person}{Zhuo Feng}.} \bibinfo{year}{2020}\natexlab{}.
\newblock \showarticletitle{GraphZoom: A Multi-level Spectral Approach for Accurate and Scalable Graph Embedding}. In \bibinfo{booktitle}{\emph{International Conference on Learning Representations}}.
\newblock
\urldef\tempurl%
\url{https://openreview.net/forum?id=r1lGO0EKDH}
\showURL{%
\tempurl}


\bibitem[Deshpande et~al\mbox{.}(2018)]%
        {CSBM}
\bibfield{author}{\bibinfo{person}{Yash Deshpande}, \bibinfo{person}{Subhabrata Sen}, \bibinfo{person}{Andrea Montanari}, {and} \bibinfo{person}{Elchanan Mossel}.} \bibinfo{year}{2018}\natexlab{}.
\newblock \showarticletitle{Contextual Stochastic Block Models}. In \bibinfo{booktitle}{\emph{Advances in Neural Information Processing Systems}}, \bibfield{editor}{\bibinfo{person}{S.~Bengio}, \bibinfo{person}{H.~Wallach}, \bibinfo{person}{H.~Larochelle}, \bibinfo{person}{K.~Grauman}, \bibinfo{person}{N.~Cesa-Bianchi}, {and} \bibinfo{person}{R.~Garnett}} (Eds.), Vol.~\bibinfo{volume}{31}. \bibinfo{publisher}{Curran Associates, Inc.}
\newblock
\urldef\tempurl%
\url{https://proceedings.neurips.cc/paper_files/paper/2018/file/08fc80de8121419136e443a70489c123-Paper.pdf}
\showURL{%
\tempurl}


\bibitem[Dickens et~al\mbox{.}(2024)]%
        {Gcoarse-gl-4}
\bibfield{author}{\bibinfo{person}{Charles Dickens}, \bibinfo{person}{Edward Huang}, \bibinfo{person}{Aishwarya Reganti}, \bibinfo{person}{Jiong Zhu}, \bibinfo{person}{Karthik Subbian}, {and} \bibinfo{person}{Danai Koutra}.} \bibinfo{year}{2024}\natexlab{}.
\newblock \showarticletitle{Graph Coarsening via Convolution Matching for Scalable Graph Neural Network Training}. In \bibinfo{booktitle}{\emph{Companion Proceedings of the ACM Web Conference 2024}} (Singapore, Singapore) \emph{(\bibinfo{series}{WWW '24})}. \bibinfo{publisher}{Association for Computing Machinery}, \bibinfo{address}{New York, NY, USA}, \bibinfo{pages}{1502–1510}.
\newblock
\showISBNx{9798400701726}
\href{https://doi.org/10.1145/3589335.3651920}{doi:\nolinkurl{10.1145/3589335.3651920}}


\bibitem[Dong et~al\mbox{.}(2020)]%
        {GraphSignalProcessingforMachineLearningAReviewandNewPerspectives}
\bibfield{author}{\bibinfo{person}{Xiaowen Dong}, \bibinfo{person}{Dorina Thanou}, \bibinfo{person}{Laura Toni}, \bibinfo{person}{Michael Bronstein}, {and} \bibinfo{person}{Pascal Frossard}.} \bibinfo{year}{2020}\natexlab{}.
\newblock \showarticletitle{Graph Signal Processing for Machine Learning: A Review and New Perspectives}.
\newblock \bibinfo{journal}{\emph{IEEE Signal Processing Magazine}} \bibinfo{volume}{37}, \bibinfo{number}{6} (\bibinfo{year}{2020}), \bibinfo{pages}{117--127}.
\newblock
\href{https://doi.org/10.1109/MSP.2020.3014591}{doi:\nolinkurl{10.1109/MSP.2020.3014591}}


\bibitem[Dou et~al\mbox{.}(2020)]%
        {GAnoDet-1-CARE-GNN}
\bibfield{author}{\bibinfo{person}{Yingtong Dou}, \bibinfo{person}{Zhiwei Liu}, \bibinfo{person}{Li Sun}, \bibinfo{person}{Yutong Deng}, \bibinfo{person}{Hao Peng}, {and} \bibinfo{person}{Philip~S. Yu}.} \bibinfo{year}{2020}\natexlab{}.
\newblock \showarticletitle{Enhancing Graph Neural Network-based Fraud Detectors against Camouflaged Fraudsters}. In \bibinfo{booktitle}{\emph{Proceedings of the 29th ACM International Conference on Information \& Knowledge Management}} (Virtual Event, Ireland) \emph{(\bibinfo{series}{CIKM '20})}. \bibinfo{publisher}{Association for Computing Machinery}, \bibinfo{address}{New York, NY, USA}, \bibinfo{pages}{315–324}.
\newblock
\showISBNx{9781450368599}
\href{https://doi.org/10.1145/3340531.3411903}{doi:\nolinkurl{10.1145/3340531.3411903}}


\bibitem[Du et~al\mbox{.}(2022)]%
        {GBKGNN}
\bibfield{author}{\bibinfo{person}{Lun Du}, \bibinfo{person}{Xiaozhou Shi}, \bibinfo{person}{Qiang Fu}, \bibinfo{person}{Xiaojun Ma}, \bibinfo{person}{Hengyu Liu}, \bibinfo{person}{Shi Han}, {and} \bibinfo{person}{Dongmei Zhang}.} \bibinfo{year}{2022}\natexlab{}.
\newblock \showarticletitle{GBK-GNN: Gated Bi-Kernel Graph Neural Networks for Modeling Both Homophily and Heterophily}. In \bibinfo{booktitle}{\emph{Proceedings of the ACM Web Conference 2022}}. \bibinfo{pages}{1550--1558}.
\newblock
\urldef\tempurl%
\url{https://doi.org/10.1145/3485447.3512201}
\showURL{%
\tempurl}


\bibitem[Dunne and Shneiderman(2013)]%
        {Gcoarse-visualize}
\bibfield{author}{\bibinfo{person}{Cody Dunne} {and} \bibinfo{person}{Ben Shneiderman}.} \bibinfo{year}{2013}\natexlab{}.
\newblock \showarticletitle{Motif simplification: improving network visualization readability with fan, connector, and clique glyphs}. In \bibinfo{booktitle}{\emph{Proceedings of the SIGCHI Conference on Human Factors in Computing Systems}} (Paris, France) \emph{(\bibinfo{series}{CHI '13})}. \bibinfo{publisher}{Association for Computing Machinery}, \bibinfo{address}{New York, NY, USA}, \bibinfo{pages}{3247–3256}.
\newblock
\showISBNx{9781450318990}
\href{https://doi.org/10.1145/2470654.2466444}{doi:\nolinkurl{10.1145/2470654.2466444}}


\bibitem[Eliasof et~al\mbox{.}(2021)]%
        {pdegcn}
\bibfield{author}{\bibinfo{person}{Moshe Eliasof}, \bibinfo{person}{Eldad Haber}, {and} \bibinfo{person}{Eran Treister}.} \bibinfo{year}{2021}\natexlab{}.
\newblock \showarticletitle{{PDE}-{GCN}: Novel Architectures for Graph Neural Networks Motivated by Partial Differential Equations}. In \bibinfo{booktitle}{\emph{Advances in Neural Information Processing Systems}}, \bibfield{editor}{\bibinfo{person}{A.~Beygelzimer}, \bibinfo{person}{Y.~Dauphin}, \bibinfo{person}{P.~Liang}, {and} \bibinfo{person}{J.~Wortman Vaughan}} (Eds.).
\newblock
\urldef\tempurl%
\url{https://openreview.net/forum?id=wWtk6GxJB2x}
\showURL{%
\tempurl}


\bibitem[Englert et~al\mbox{.}(2014)]%
        {Gcoarse-co}
\bibfield{author}{\bibinfo{person}{Matthias Englert}, \bibinfo{person}{Anupam Gupta}, \bibinfo{person}{Robert Krauthgamer}, \bibinfo{person}{Harald R\"{a}cke}, \bibinfo{person}{Inbal Talgam-Cohen}, {and} \bibinfo{person}{Kunal Talwar}.} \bibinfo{year}{2014}\natexlab{}.
\newblock \showarticletitle{Vertex Sparsifiers: New Results from Old Techniques}.
\newblock \bibinfo{journal}{\emph{SIAM J. Comput.}} \bibinfo{volume}{43}, \bibinfo{number}{4} (\bibinfo{year}{2014}), \bibinfo{pages}{1239--1262}.
\newblock
\href{https://doi.org/10.1137/130908440}{doi:\nolinkurl{10.1137/130908440}}
\showeprint{https://doi.org/10.1137/130908440}


\bibitem[Fey and Lenssen(2019)]%
        {PyTorchGeometric}
\bibfield{author}{\bibinfo{person}{Matthias Fey} {and} \bibinfo{person}{Jan~Eric Lenssen}.} \bibinfo{year}{2019}\natexlab{}.
\newblock \bibinfo{title}{Fast Graph Representation Learning with PyTorch Geometric}.
\newblock
\href{https://doi.org/10.48550/ARXIV.1903.02428}{doi:\nolinkurl{10.48550/ARXIV.1903.02428}}


\bibitem[Gao et~al\mbox{.}(2023a)]%
        {GAnoDet-spectral-2-GHRN}
\bibfield{author}{\bibinfo{person}{Yuan Gao}, \bibinfo{person}{Xiang Wang}, \bibinfo{person}{Xiangnan He}, \bibinfo{person}{Zhenguang Liu}, \bibinfo{person}{Huamin Feng}, {and} \bibinfo{person}{Yongdong Zhang}.} \bibinfo{year}{2023}\natexlab{a}.
\newblock \showarticletitle{Addressing Heterophily in Graph Anomaly Detection: A Perspective of Graph Spectrum}. In \bibinfo{booktitle}{\emph{Proceedings of the ACM Web Conference 2023}} (Austin, TX, USA) \emph{(\bibinfo{series}{WWW '23})}. \bibinfo{publisher}{Association for Computing Machinery}, \bibinfo{address}{New York, NY, USA}, \bibinfo{pages}{1528–1538}.
\newblock
\showISBNx{9781450394161}
\href{https://doi.org/10.1145/3543507.3583268}{doi:\nolinkurl{10.1145/3543507.3583268}}


\bibitem[Gao et~al\mbox{.}(2023b)]%
        {GAnoDet-3-GDN}
\bibfield{author}{\bibinfo{person}{Yuan Gao}, \bibinfo{person}{Xiang Wang}, \bibinfo{person}{Xiangnan He}, \bibinfo{person}{Zhenguang Liu}, \bibinfo{person}{Huamin Feng}, {and} \bibinfo{person}{Yongdong Zhang}.} \bibinfo{year}{2023}\natexlab{b}.
\newblock \showarticletitle{Alleviating Structural Distribution Shift in Graph Anomaly Detection}. In \bibinfo{booktitle}{\emph{Proceedings of the Sixteenth ACM International Conference on Web Search and Data Mining}} (Singapore, Singapore) \emph{(\bibinfo{series}{WSDM '23})}. \bibinfo{publisher}{Association for Computing Machinery}, \bibinfo{address}{New York, NY, USA}, \bibinfo{pages}{357–365}.
\newblock
\showISBNx{9781450394079}
\href{https://doi.org/10.1145/3539597.3570377}{doi:\nolinkurl{10.1145/3539597.3570377}}


\bibitem[Gasteiger et~al\mbox{.}(2019)]%
        {decoupled-advantages-3-coupled-disadvantages-2-APPNP}
\bibfield{author}{\bibinfo{person}{Johannes Gasteiger}, \bibinfo{person}{Aleksandar Bojchevski}, {and} \bibinfo{person}{Stephan Günnemann}.} \bibinfo{year}{2019}\natexlab{}.
\newblock \showarticletitle{Predict then Propagate: Graph Neural Networks meet Personalized PageRank}. In \bibinfo{booktitle}{\emph{International Conference on Learning Representations}}.
\newblock
\urldef\tempurl%
\url{https://openreview.net/forum?id=H1gL-2A9Ym}
\showURL{%
\tempurl}


\bibitem[Gong et~al\mbox{.}(2024)]%
        {heterophily-gnn-survey-2}
\bibfield{author}{\bibinfo{person}{Chenghua Gong}, \bibinfo{person}{Yao Cheng}, \bibinfo{person}{Xiang Li}, \bibinfo{person}{Caihua Shan}, {and} \bibinfo{person}{Siqiang Luo}.} \bibinfo{year}{2024}\natexlab{}.
\newblock \bibinfo{title}{Learning from Graphs with Heterophily: Progress and Future}.
\newblock
\showeprint[arxiv]{2401.09769}~[cs.SI]
\urldef\tempurl%
\url{https://arxiv.org/abs/2401.09769}
\showURL{%
\tempurl}


\bibitem[Guo and Wei(2023)]%
        {OptBasisGNN}
\bibfield{author}{\bibinfo{person}{Yuhe Guo} {and} \bibinfo{person}{Zhewei Wei}.} \bibinfo{year}{2023}\natexlab{}.
\newblock \showarticletitle{Graph Neural Networks with Learnable and Optimal Polynomial Bases}. In \bibinfo{booktitle}{\emph{Proceedings of the 40th International Conference on Machine Learning}} \emph{(\bibinfo{series}{Proceedings of Machine Learning Research}, Vol.~\bibinfo{volume}{202})}, \bibfield{editor}{\bibinfo{person}{Andreas Krause}, \bibinfo{person}{Emma Brunskill}, \bibinfo{person}{Kyunghyun Cho}, \bibinfo{person}{Barbara Engelhardt}, \bibinfo{person}{Sivan Sabato}, {and} \bibinfo{person}{Jonathan Scarlett}} (Eds.). \bibinfo{publisher}{PMLR}, \bibinfo{pages}{12077--12097}.
\newblock
\urldef\tempurl%
\url{https://proceedings.mlr.press/v202/guo23i.html}
\showURL{%
\tempurl}


\bibitem[Han et~al\mbox{.}(2024)]%
        {nodewise-3}
\bibfield{author}{\bibinfo{person}{Haoyu Han}, \bibinfo{person}{Juanhui Li}, \bibinfo{person}{Wei Huang}, \bibinfo{person}{Xianfeng Tang}, \bibinfo{person}{Hanqing Lu}, \bibinfo{person}{Chen Luo}, \bibinfo{person}{Hui Liu}, {and} \bibinfo{person}{Jiliang Tang}.} \bibinfo{year}{2024}\natexlab{}.
\newblock \bibinfo{title}{Node-wise Filtering in Graph Neural Networks: A Mixture of Experts Approach}.
\newblock
\showeprint[arxiv]{2406.03464}~[cs.LG]
\urldef\tempurl%
\url{https://arxiv.org/abs/2406.03464}
\showURL{%
\tempurl}


\bibitem[He et~al\mbox{.}(2021)]%
        {BernNet-GNN-narrowbandresults-1}
\bibfield{author}{\bibinfo{person}{Mingguo He}, \bibinfo{person}{Zhewei Wei}, \bibinfo{person}{Zengfeng Huang}, {and} \bibinfo{person}{Hongteng Xu}.} \bibinfo{year}{2021}\natexlab{}.
\newblock \showarticletitle{BernNet: Learning Arbitrary Graph Spectral Filters via Bernstein Approximation}. In \bibinfo{booktitle}{\emph{Advances in Neural Information Processing Systems}}, \bibfield{editor}{\bibinfo{person}{A.~Beygelzimer}, \bibinfo{person}{Y.~Dauphin}, \bibinfo{person}{P.~Liang}, {and} \bibinfo{person}{J.~Wortman Vaughan}} (Eds.).
\newblock
\urldef\tempurl%
\url{https://openreview.net/forum?id=WigDnV-_Gq}
\showURL{%
\tempurl}


\bibitem[He et~al\mbox{.}(2022)]%
        {ChebNetII}
\bibfield{author}{\bibinfo{person}{Mingguo He}, \bibinfo{person}{Zhewei Wei}, {and} \bibinfo{person}{Ji-Rong Wen}.} \bibinfo{year}{2022}\natexlab{}.
\newblock \showarticletitle{Convolutional Neural Networks on Graphs with Chebyshev Approximation, Revisited}. In \bibinfo{booktitle}{\emph{Advances in Neural Information Processing Systems}}, \bibfield{editor}{\bibinfo{person}{Alice~H. Oh}, \bibinfo{person}{Alekh Agarwal}, \bibinfo{person}{Danielle Belgrave}, {and} \bibinfo{person}{Kyunghyun Cho}} (Eds.).
\newblock
\urldef\tempurl%
\url{https://openreview.net/forum?id=jxPJ4QA0KAb}
\showURL{%
\tempurl}


\bibitem[He et~al\mbox{.}(2020)]%
        {recsys_1}
\bibfield{author}{\bibinfo{person}{Xiangnan He}, \bibinfo{person}{Kuan Deng}, \bibinfo{person}{Xiang Wang}, \bibinfo{person}{Yan Li}, \bibinfo{person}{Yongdong Zhang}, {and} \bibinfo{person}{Meng Wang}.} \bibinfo{year}{2020}\natexlab{}.
\newblock \bibinfo{title}{LightGCN: Simplifying and Powering Graph Convolution Network for Recommendation}.
\newblock
\href{https://doi.org/10.48550/ARXIV.2002.02126}{doi:\nolinkurl{10.48550/ARXIV.2002.02126}}


\bibitem[Hu et~al\mbox{.}(2020)]%
        {dataset5-ogb}
\bibfield{author}{\bibinfo{person}{Weihua Hu}, \bibinfo{person}{Matthias Fey}, \bibinfo{person}{Marinka Zitnik}, \bibinfo{person}{Yuxiao Dong}, \bibinfo{person}{Hongyu Ren}, \bibinfo{person}{Bowen Liu}, \bibinfo{person}{Michele Catasta}, {and} \bibinfo{person}{Jure Leskovec}.} \bibinfo{year}{2020}\natexlab{}.
\newblock \showarticletitle{Open Graph Benchmark: Datasets for Machine Learning on Graphs}. In \bibinfo{booktitle}{\emph{Advances in Neural Information Processing Systems}}, Vol.~\bibinfo{volume}{33}. \bibinfo{pages}{22118--22133}.
\newblock
\urldef\tempurl%
\url{https://proceedings.neurips.cc/paper_files/paper/2020/file/fb60d411a5c5b72b2e7d3527cfc84fd0-Paper.pdf}
\showURL{%
\tempurl}


\bibitem[Huang et~al\mbox{.}(2024a)]%
        {decoupled-AdaptKry}
\bibfield{author}{\bibinfo{person}{Keke Huang}, \bibinfo{person}{Wencai Cao}, \bibinfo{person}{Hoang Ta}, \bibinfo{person}{Xiaokui Xiao}, {and} \bibinfo{person}{Pietro Li\`{o}}.} \bibinfo{year}{2024}\natexlab{a}.
\newblock \showarticletitle{Optimizing Polynomial Graph Filters: A Novel Adaptive Krylov Subspace Approach}. In \bibinfo{booktitle}{\emph{Proceedings of the ACM Web Conference 2024}} (Singapore, Singapore) \emph{(\bibinfo{series}{WWW '24})}. \bibinfo{publisher}{Association for Computing Machinery}, \bibinfo{address}{New York, NY, USA}, \bibinfo{pages}{1057–1068}.
\newblock
\showISBNx{9798400701719}
\href{https://doi.org/10.1145/3589334.3645705}{doi:\nolinkurl{10.1145/3589334.3645705}}


\bibitem[Huang et~al\mbox{.}(2023)]%
        {nodewise-1}
\bibfield{author}{\bibinfo{person}{Keke Huang}, \bibinfo{person}{Jing Tang}, \bibinfo{person}{Juncheng Liu}, \bibinfo{person}{Renchi Yang}, {and} \bibinfo{person}{Xiaokui Xiao}.} \bibinfo{year}{2023}\natexlab{}.
\newblock \showarticletitle{Node-wise Diffusion for Scalable Graph Learning}. In \bibinfo{booktitle}{\emph{Proceedings of the ACM Web Conference 2023}} (Austin, TX, USA) \emph{(\bibinfo{series}{WWW '23})}. \bibinfo{publisher}{Association for Computing Machinery}, \bibinfo{address}{New York, NY, USA}, \bibinfo{pages}{1723–1733}.
\newblock
\showISBNx{9781450394161}
\href{https://doi.org/10.1145/3543507.3583408}{doi:\nolinkurl{10.1145/3543507.3583408}}


\bibitem[Huang et~al\mbox{.}(2024b)]%
        {decoupled-UniFilter}
\bibfield{author}{\bibinfo{person}{Keke Huang}, \bibinfo{person}{Yu~Guang Wang}, \bibinfo{person}{Ming Li}, {and} \bibinfo{person}{Pietro Lio}.} \bibinfo{year}{2024}\natexlab{b}.
\newblock \showarticletitle{How Universal Polynomial Bases Enhance Spectral Graph Neural Networks: Heterophily, Over-smoothing, and Over-squashing}. In \bibinfo{booktitle}{\emph{Forty-first International Conference on Machine Learning}}.
\newblock
\urldef\tempurl%
\url{https://openreview.net/forum?id=Z2LH6Va7L2}
\showURL{%
\tempurl}


\bibitem[Huang et~al\mbox{.}(2021)]%
        {Gcoarse-gl-2}
\bibfield{author}{\bibinfo{person}{Zengfeng Huang}, \bibinfo{person}{Shengzhong Zhang}, \bibinfo{person}{Chong Xi}, \bibinfo{person}{Tang Liu}, {and} \bibinfo{person}{Min Zhou}.} \bibinfo{year}{2021}\natexlab{}.
\newblock \showarticletitle{Scaling Up Graph Neural Networks Via Graph Coarsening}. In \bibinfo{booktitle}{\emph{Proceedings of the 27th ACM SIGKDD Conference on Knowledge Discovery \& Data Mining}} (Virtual Event, Singapore) \emph{(\bibinfo{series}{KDD '21})}. \bibinfo{publisher}{Association for Computing Machinery}, \bibinfo{address}{New York, NY, USA}, \bibinfo{pages}{675–684}.
\newblock
\showISBNx{9781450383325}
\href{https://doi.org/10.1145/3447548.3467256}{doi:\nolinkurl{10.1145/3447548.3467256}}


\bibitem[Jain et~al\mbox{.}(1999)]%
        {clustering-survey}
\bibfield{author}{\bibinfo{person}{A.~K. Jain}, \bibinfo{person}{M.~N. Murty}, {and} \bibinfo{person}{P.~J. Flynn}.} \bibinfo{year}{1999}\natexlab{}.
\newblock \showarticletitle{Data clustering: a review}.
\newblock \bibinfo{journal}{\emph{ACM Comput. Surv.}} \bibinfo{volume}{31}, \bibinfo{number}{3} (\bibinfo{date}{Sept.} \bibinfo{year}{1999}), \bibinfo{pages}{264–323}.
\newblock
\showISSN{0360-0300}
\href{https://doi.org/10.1145/331499.331504}{doi:\nolinkurl{10.1145/331499.331504}}


\bibitem[Jin et~al\mbox{.}(2020)]%
        {rsa-Gcoarse-4-MGC-SGC}
\bibfield{author}{\bibinfo{person}{Yu Jin}, \bibinfo{person}{Andreas Loukas}, {and} \bibinfo{person}{Joseph JaJa}.} \bibinfo{year}{2020}\natexlab{}.
\newblock \showarticletitle{Graph Coarsening with Preserved Spectral Properties}. In \bibinfo{booktitle}{\emph{Proceedings of the Twenty Third International Conference on Artificial Intelligence and Statistics}} \emph{(\bibinfo{series}{Proceedings of Machine Learning Research}, Vol.~\bibinfo{volume}{108})}, \bibfield{editor}{\bibinfo{person}{Silvia Chiappa} {and} \bibinfo{person}{Roberto Calandra}} (Eds.). \bibinfo{publisher}{PMLR}, \bibinfo{pages}{4452--4462}.
\newblock
\urldef\tempurl%
\url{https://proceedings.mlr.press/v108/jin20a.html}
\showURL{%
\tempurl}


\bibitem[Joly and Keriven(2024)]%
        {rsa-Gcoarse-7}
\bibfield{author}{\bibinfo{person}{Antonin Joly} {and} \bibinfo{person}{Nicolas Keriven}.} \bibinfo{year}{2024}\natexlab{}.
\newblock \bibinfo{title}{Graph Coarsening with Message-Passing Guarantees}.
\newblock
\showeprint[arxiv]{2405.18127}~[cs.LG]
\urldef\tempurl%
\url{https://arxiv.org/abs/2405.18127}
\showURL{%
\tempurl}


\bibitem[Kingma and Ba(2014)]%
        {Adamoptimizer}
\bibfield{author}{\bibinfo{person}{Diederik~P. Kingma} {and} \bibinfo{person}{Jimmy Ba}.} \bibinfo{year}{2014}\natexlab{}.
\newblock \bibinfo{title}{Adam: A Method for Stochastic Optimization}.
\newblock
\href{https://doi.org/10.48550/ARXIV.1412.6980}{doi:\nolinkurl{10.48550/ARXIV.1412.6980}}


\bibitem[Kipf and Welling(2017)]%
        {GCN}
\bibfield{author}{\bibinfo{person}{Thomas~N. Kipf} {and} \bibinfo{person}{Max Welling}.} \bibinfo{year}{2017}\natexlab{}.
\newblock \showarticletitle{Semi-Supervised Classification with Graph Convolutional Networks}. In \bibinfo{booktitle}{\emph{International Conference on Learning Representations}}.
\newblock
\urldef\tempurl%
\url{https://openreview.net/forum?id=SJU4ayYgl}
\showURL{%
\tempurl}


\bibitem[Kumar et~al\mbox{.}(2023a)]%
        {rsa-Gcoarse-5}
\bibfield{author}{\bibinfo{person}{Manoj Kumar}, \bibinfo{person}{Anurag Sharma}, {and} \bibinfo{person}{Sandeep Kumar}.} \bibinfo{year}{2023}\natexlab{a}.
\newblock \showarticletitle{A Unified Framework for Optimization-Based Graph Coarsening}.
\newblock \bibinfo{journal}{\emph{Journal of Machine Learning Research}} \bibinfo{volume}{24}, \bibinfo{number}{118} (\bibinfo{year}{2023}), \bibinfo{pages}{1--50}.
\newblock
\urldef\tempurl%
\url{http://jmlr.org/papers/v24/22-1085.html}
\showURL{%
\tempurl}


\bibitem[Kumar et~al\mbox{.}(2023b)]%
        {rsa-Gcoarse-8-FGC}
\bibfield{author}{\bibinfo{person}{Manoj Kumar}, \bibinfo{person}{Anurag Sharma}, \bibinfo{person}{Shashwat Saxena}, {and} \bibinfo{person}{Sandeep Kumar}.} \bibinfo{year}{2023}\natexlab{b}.
\newblock \showarticletitle{Featured Graph Coarsening with Similarity Guarantees}. In \bibinfo{booktitle}{\emph{Proceedings of the 40th International Conference on Machine Learning}} \emph{(\bibinfo{series}{Proceedings of Machine Learning Research}, Vol.~\bibinfo{volume}{202})}, \bibfield{editor}{\bibinfo{person}{Andreas Krause}, \bibinfo{person}{Emma Brunskill}, \bibinfo{person}{Kyunghyun Cho}, \bibinfo{person}{Barbara Engelhardt}, \bibinfo{person}{Sivan Sabato}, {and} \bibinfo{person}{Jonathan Scarlett}} (Eds.). \bibinfo{publisher}{PMLR}, \bibinfo{pages}{17953--17975}.
\newblock
\urldef\tempurl%
\url{https://proceedings.mlr.press/v202/kumar23a.html}
\showURL{%
\tempurl}


\bibitem[Li et~al\mbox{.}(2024b)]%
        {decoupled-PCConv}
\bibfield{author}{\bibinfo{person}{Bingheng Li}, \bibinfo{person}{Erlin Pan}, {and} \bibinfo{person}{Zhao Kang}.} \bibinfo{year}{2024}\natexlab{b}.
\newblock \showarticletitle{PC-Conv: Unifying Homophily and Heterophily with Two-Fold Filtering}.
\newblock \bibinfo{journal}{\emph{Proceedings of the AAAI Conference on Artificial Intelligence}} \bibinfo{volume}{38}, \bibinfo{number}{12} (\bibinfo{date}{Mar.} \bibinfo{year}{2024}), \bibinfo{pages}{13437--13445}.
\newblock
\href{https://doi.org/10.1609/aaai.v38i12.29246}{doi:\nolinkurl{10.1609/aaai.v38i12.29246}}


\bibitem[Li et~al\mbox{.}(2025a)]%
        {ERGNN}
\bibfield{author}{\bibinfo{person}{Guoming Li}, \bibinfo{person}{Jian Yang}, {and} \bibinfo{person}{Shangsong Liang}.} \bibinfo{year}{2025}\natexlab{a}.
\newblock \showarticletitle{ERGNN: Spectral Graph Neural Network With Explicitly-Optimized Rational Graph Filters}. In \bibinfo{booktitle}{\emph{ICASSP 2025 - 2025 IEEE International Conference on Acoustics, Speech and Signal Processing (ICASSP)}}. \bibinfo{pages}{1--5}.
\newblock
\href{https://doi.org/10.1109/ICASSP49660.2025.10888930}{doi:\nolinkurl{10.1109/ICASSP49660.2025.10888930}}


\bibitem[Li et~al\mbox{.}(2024c)]%
        {chebnet2d}
\bibfield{author}{\bibinfo{person}{Guoming Li}, \bibinfo{person}{Jian Yang}, \bibinfo{person}{Shangsong Liang}, {and} \bibinfo{person}{Dongsheng Luo}.} \bibinfo{year}{2024}\natexlab{c}.
\newblock \bibinfo{title}{Spectral GNN via Two-dimensional (2-D) Graph Convolution}.
\newblock
\showeprint[arxiv]{2404.04559}~[cs.LG]


\bibitem[Li et~al\mbox{.}(2025b)]%
        {decoupled-TFGNN}
\bibfield{author}{\bibinfo{person}{Guoming Li}, \bibinfo{person}{Jian Yang}, \bibinfo{person}{Shangsong Liang}, {and} \bibinfo{person}{Dongsheng Luo}.} \bibinfo{year}{2025}\natexlab{b}.
\newblock \showarticletitle{Polynomial Selection in Spectral Graph Neural Networks: An Error-Sum of Function Slices Approach}. In \bibinfo{booktitle}{\emph{Proceedings of the ACM on Web Conference 2025}} (Sydney NSW, Australia) \emph{(\bibinfo{series}{WWW '25})}. \bibinfo{publisher}{Association for Computing Machinery}, \bibinfo{address}{New York, NY, USA}, \bibinfo{pages}{3276–3287}.
\newblock
\showISBNx{9798400712746}
\href{https://doi.org/10.1145/3696410.3714760}{doi:\nolinkurl{10.1145/3696410.3714760}}


\bibitem[Li et~al\mbox{.}(2024a)]%
        {nodewise-2}
\bibfield{author}{\bibinfo{person}{Xunkai Li}, \bibinfo{person}{Jingyuan Ma}, \bibinfo{person}{Zhengyu Wu}, \bibinfo{person}{Daohan Su}, \bibinfo{person}{Wentao Zhang}, \bibinfo{person}{Rong-Hua Li}, {and} \bibinfo{person}{Guoren Wang}.} \bibinfo{year}{2024}\natexlab{a}.
\newblock \showarticletitle{Rethinking Node-wise Propagation for Large-scale Graph Learning}. In \bibinfo{booktitle}{\emph{Proceedings of the ACM Web Conference 2024}} (Singapore, Singapore) \emph{(\bibinfo{series}{WWW '24})}. \bibinfo{publisher}{Association for Computing Machinery}, \bibinfo{address}{New York, NY, USA}, \bibinfo{pages}{560–569}.
\newblock
\showISBNx{9798400701719}
\href{https://doi.org/10.1145/3589334.3645450}{doi:\nolinkurl{10.1145/3589334.3645450}}


\bibitem[Li et~al\mbox{.}(2022)]%
        {glognn++}
\bibfield{author}{\bibinfo{person}{Xiang Li}, \bibinfo{person}{Renyu Zhu}, \bibinfo{person}{Yao Cheng}, \bibinfo{person}{Caihua Shan}, \bibinfo{person}{Siqiang Luo}, \bibinfo{person}{Dongsheng Li}, {and} \bibinfo{person}{Weining Qian}.} \bibinfo{year}{2022}\natexlab{}.
\newblock \showarticletitle{Finding Global Homophily in Graph Neural Networks When Meeting Heterophily}. In \bibinfo{booktitle}{\emph{Proceedings of the 39th International Conference on Machine Learning}} \emph{(\bibinfo{series}{Proceedings of Machine Learning Research}, Vol.~\bibinfo{volume}{162})}, \bibfield{editor}{\bibinfo{person}{Kamalika Chaudhuri}, \bibinfo{person}{Stefanie Jegelka}, \bibinfo{person}{Le~Song}, \bibinfo{person}{Csaba Szepesvari}, \bibinfo{person}{Gang Niu}, {and} \bibinfo{person}{Sivan Sabato}} (Eds.). \bibinfo{publisher}{PMLR}, \bibinfo{pages}{13242--13256}.
\newblock
\urldef\tempurl%
\url{https://proceedings.mlr.press/v162/li22ad.html}
\showURL{%
\tempurl}


\bibitem[Liang et~al\mbox{.}(2023)]%
        {low-rank-gnn-2}
\bibfield{author}{\bibinfo{person}{Langzhang Liang}, \bibinfo{person}{Xiangjing Hu}, \bibinfo{person}{Zenglin Xu}, \bibinfo{person}{Zixing Song}, {and} \bibinfo{person}{Irwin King}.} \bibinfo{year}{2023}\natexlab{}.
\newblock \showarticletitle{Predicting Global Label Relationship Matrix for Graph Neural Networks under Heterophily}. In \bibinfo{booktitle}{\emph{Advances in Neural Information Processing Systems}}, \bibfield{editor}{\bibinfo{person}{A.~Oh}, \bibinfo{person}{T.~Naumann}, \bibinfo{person}{A.~Globerson}, \bibinfo{person}{K.~Saenko}, \bibinfo{person}{M.~Hardt}, {and} \bibinfo{person}{S.~Levine}} (Eds.), Vol.~\bibinfo{volume}{36}. \bibinfo{publisher}{Curran Associates, Inc.}, \bibinfo{pages}{10909--10921}.
\newblock
\urldef\tempurl%
\url{https://proceedings.neurips.cc/paper_files/paper/2023/file/23aa2163dea287441ebebc1295d5b3fc-Paper-Conference.pdf}
\showURL{%
\tempurl}


\bibitem[Liao et~al\mbox{.}(2023)]%
        {decoupled-LD2}
\bibfield{author}{\bibinfo{person}{Ningyi Liao}, \bibinfo{person}{Siqiang Luo}, \bibinfo{person}{Xiang Li}, {and} \bibinfo{person}{Jieming Shi}.} \bibinfo{year}{2023}\natexlab{}.
\newblock \showarticletitle{{LD}2: Scalable Heterophilous Graph Neural Network with Decoupled Embeddings}. In \bibinfo{booktitle}{\emph{Thirty-seventh Conference on Neural Information Processing Systems}}.
\newblock
\urldef\tempurl%
\url{https://openreview.net/forum?id=7zkFc9TGKz}
\showURL{%
\tempurl}


\bibitem[Liao et~al\mbox{.}(2024)]%
        {decoupled-SCARA}
\bibfield{author}{\bibinfo{person}{Ningyi Liao}, \bibinfo{person}{Dingheng Mo}, \bibinfo{person}{Siqiang Luo}, \bibinfo{person}{Xiang Li}, {and} \bibinfo{person}{Pengcheng Yin}.} \bibinfo{year}{2024}\natexlab{}.
\newblock \showarticletitle{Scalable decoupling graph neural network with feature-oriented optimization}.
\newblock \bibinfo{journal}{\emph{The VLDB Journal}} \bibinfo{volume}{33}, \bibinfo{number}{3} (\bibinfo{year}{2024}), \bibinfo{pages}{667--683}.
\newblock


\bibitem[Lim et~al\mbox{.}(2021)]%
        {dataset6-large-hetero}
\bibfield{author}{\bibinfo{person}{Derek Lim}, \bibinfo{person}{Felix~Matthew Hohne}, \bibinfo{person}{Xiuyu Li}, \bibinfo{person}{Sijia~Linda Huang}, \bibinfo{person}{Vaishnavi Gupta}, \bibinfo{person}{Omkar~Prasad Bhalerao}, {and} \bibinfo{person}{Ser-Nam Lim}.} \bibinfo{year}{2021}\natexlab{}.
\newblock \showarticletitle{Large Scale Learning on Non-Homophilous Graphs: New Benchmarks and Strong Simple Methods}. In \bibinfo{booktitle}{\emph{Advances in Neural Information Processing Systems}}, \bibfield{editor}{\bibinfo{person}{A.~Beygelzimer}, \bibinfo{person}{Y.~Dauphin}, \bibinfo{person}{P.~Liang}, {and} \bibinfo{person}{J.~Wortman Vaughan}} (Eds.).
\newblock
\urldef\tempurl%
\url{https://openreview.net/forum?id=DfGu8WwT0d}
\showURL{%
\tempurl}


\bibitem[Liu et~al\mbox{.}(2021)]%
        {GAnoDet-2-PC-GNN}
\bibfield{author}{\bibinfo{person}{Yang Liu}, \bibinfo{person}{Xiang Ao}, \bibinfo{person}{Zidi Qin}, \bibinfo{person}{Jianfeng Chi}, \bibinfo{person}{Jinghua Feng}, \bibinfo{person}{Hao Yang}, {and} \bibinfo{person}{Qing He}.} \bibinfo{year}{2021}\natexlab{}.
\newblock \showarticletitle{Pick and Choose: A GNN-based Imbalanced Learning Approach for Fraud Detection}. In \bibinfo{booktitle}{\emph{Proceedings of the Web Conference 2021}} (Ljubljana, Slovenia) \emph{(\bibinfo{series}{WWW '21})}. \bibinfo{publisher}{Association for Computing Machinery}, \bibinfo{address}{New York, NY, USA}, \bibinfo{pages}{3168–3177}.
\newblock
\showISBNx{9781450383127}
\href{https://doi.org/10.1145/3442381.3449989}{doi:\nolinkurl{10.1145/3442381.3449989}}


\bibitem[Liu et~al\mbox{.}(2018)]%
        {Gcoarse_survey}
\bibfield{author}{\bibinfo{person}{Yike Liu}, \bibinfo{person}{Tara Safavi}, \bibinfo{person}{Abhilash Dighe}, {and} \bibinfo{person}{Danai Koutra}.} \bibinfo{year}{2018}\natexlab{}.
\newblock \showarticletitle{Graph Summarization Methods and Applications: A Survey}.
\newblock \bibinfo{journal}{\emph{ACM Comput. Surv.}} \bibinfo{volume}{51}, \bibinfo{number}{3}, Article \bibinfo{articleno}{62} (\bibinfo{date}{jun} \bibinfo{year}{2018}), \bibinfo{numpages}{34}~pages.
\newblock
\showISSN{0360-0300}
\href{https://doi.org/10.1145/3186727}{doi:\nolinkurl{10.1145/3186727}}


\bibitem[Loukas(2019)]%
        {rsa-Gcoarse-3-Local-Variation}
\bibfield{author}{\bibinfo{person}{Andreas Loukas}.} \bibinfo{year}{2019}\natexlab{}.
\newblock \showarticletitle{Graph Reduction with Spectral and Cut Guarantees}.
\newblock \bibinfo{journal}{\emph{Journal of Machine Learning Research}} \bibinfo{volume}{20}, \bibinfo{number}{116} (\bibinfo{year}{2019}), \bibinfo{pages}{1--42}.
\newblock
\urldef\tempurl%
\url{http://jmlr.org/papers/v20/18-680.html}
\showURL{%
\tempurl}


\bibitem[Loukas and Vandergheynst(2018)]%
        {rsa-Gcoarse-2}
\bibfield{author}{\bibinfo{person}{Andreas Loukas} {and} \bibinfo{person}{Pierre Vandergheynst}.} \bibinfo{year}{2018}\natexlab{}.
\newblock \showarticletitle{Spectrally Approximating Large Graphs with Smaller Graphs}. In \bibinfo{booktitle}{\emph{Proceedings of the 35th International Conference on Machine Learning}} \emph{(\bibinfo{series}{Proceedings of Machine Learning Research}, Vol.~\bibinfo{volume}{80})}, \bibfield{editor}{\bibinfo{person}{Jennifer Dy} {and} \bibinfo{person}{Andreas Krause}} (Eds.). \bibinfo{publisher}{PMLR}, \bibinfo{pages}{3237--3246}.
\newblock
\urldef\tempurl%
\url{https://proceedings.mlr.press/v80/loukas18a.html}
\showURL{%
\tempurl}


\bibitem[Luan et~al\mbox{.}(2024)]%
        {heterophily-gnn-survey-3}
\bibfield{author}{\bibinfo{person}{Sitao Luan}, \bibinfo{person}{Chenqing Hua}, \bibinfo{person}{Qincheng Lu}, \bibinfo{person}{Liheng Ma}, \bibinfo{person}{Lirong Wu}, \bibinfo{person}{Xinyu Wang}, \bibinfo{person}{Minkai Xu}, \bibinfo{person}{Xiao-Wen Chang}, \bibinfo{person}{Doina Precup}, \bibinfo{person}{Rex Ying}, \bibinfo{person}{Stan~Z. Li}, \bibinfo{person}{Jian Tang}, \bibinfo{person}{Guy Wolf}, {and} \bibinfo{person}{Stefanie Jegelka}.} \bibinfo{year}{2024}\natexlab{}.
\newblock \bibinfo{title}{The Heterophilic Graph Learning Handbook: Benchmarks, Models, Theoretical Analysis, Applications and Challenges}.
\newblock
\showeprint[arxiv]{2407.09618}~[cs.LG]
\urldef\tempurl%
\url{https://arxiv.org/abs/2407.09618}
\showURL{%
\tempurl}


\bibitem[Luan et~al\mbox{.}(2022)]%
        {ACMGCN}
\bibfield{author}{\bibinfo{person}{Sitao Luan}, \bibinfo{person}{Chenqing Hua}, \bibinfo{person}{Qincheng Lu}, \bibinfo{person}{Jiaqi Zhu}, \bibinfo{person}{Mingde Zhao}, \bibinfo{person}{Shuyuan Zhang}, \bibinfo{person}{Xiao-Wen Chang}, {and} \bibinfo{person}{Doina Precup}.} \bibinfo{year}{2022}\natexlab{}.
\newblock \showarticletitle{Revisiting Heterophily For Graph Neural Networks}. In \bibinfo{booktitle}{\emph{Advances in Neural Information Processing Systems}}, \bibfield{editor}{\bibinfo{person}{S.~Koyejo}, \bibinfo{person}{S.~Mohamed}, \bibinfo{person}{A.~Agarwal}, \bibinfo{person}{D.~Belgrave}, \bibinfo{person}{K.~Cho}, {and} \bibinfo{person}{A.~Oh}} (Eds.), Vol.~\bibinfo{volume}{35}. \bibinfo{publisher}{Curran Associates, Inc.}, \bibinfo{pages}{1362--1375}.
\newblock
\urldef\tempurl%
\url{https://proceedings.neurips.cc/paper_files/paper/2022/file/092359ce5cf60a80e882378944bf1be4-Paper-Conference.pdf}
\showURL{%
\tempurl}


\bibitem[Ma et~al\mbox{.}(2023)]%
        {GAnoDet-survey-1}
\bibfield{author}{\bibinfo{person}{Xiaoxiao Ma}, \bibinfo{person}{Jia Wu}, \bibinfo{person}{Shan Xue}, \bibinfo{person}{Jian Yang}, \bibinfo{person}{Chuan Zhou}, \bibinfo{person}{Quan~Z. Sheng}, \bibinfo{person}{Hui Xiong}, {and} \bibinfo{person}{Leman Akoglu}.} \bibinfo{year}{2023}\natexlab{}.
\newblock \showarticletitle{A Comprehensive Survey on Graph Anomaly Detection With Deep Learning}.
\newblock \bibinfo{journal}{\emph{IEEE Transactions on Knowledge and Data Engineering}} \bibinfo{volume}{35}, \bibinfo{number}{12} (\bibinfo{year}{2023}), \bibinfo{pages}{12012--12038}.
\newblock
\href{https://doi.org/10.1109/TKDE.2021.3118815}{doi:\nolinkurl{10.1109/TKDE.2021.3118815}}


\bibitem[MacQueen et~al\mbox{.}(1967)]%
        {kmeans}
\bibfield{author}{\bibinfo{person}{James MacQueen} {et~al\mbox{.}}} \bibinfo{year}{1967}\natexlab{}.
\newblock \showarticletitle{Some methods for classification and analysis of multivariate observations}. In \bibinfo{booktitle}{\emph{Proceedings of the fifth Berkeley symposium on mathematical statistics and probability}}, Vol.~\bibinfo{volume}{1}. Oakland, CA, USA, \bibinfo{pages}{281--297}.
\newblock


\bibitem[Manning(2009)]%
        {kmeans-complexity}
\bibfield{author}{\bibinfo{person}{Christopher~D Manning}.} \bibinfo{year}{2009}\natexlab{}.
\newblock \bibinfo{booktitle}{\emph{An introduction to information retrieval}}.
\newblock


\bibitem[Maskey et~al\mbox{.}(2023)]%
        {FLODE}
\bibfield{author}{\bibinfo{person}{Sohir Maskey}, \bibinfo{person}{Raffaele Paolino}, \bibinfo{person}{Aras Bacho}, {and} \bibinfo{person}{Gitta Kutyniok}.} \bibinfo{year}{2023}\natexlab{}.
\newblock \showarticletitle{A Fractional Graph Laplacian Approach to Oversmoothing}. In \bibinfo{booktitle}{\emph{Advances in Neural Information Processing Systems}}, \bibfield{editor}{\bibinfo{person}{A.~Oh}, \bibinfo{person}{T.~Naumann}, \bibinfo{person}{A.~Globerson}, \bibinfo{person}{K.~Saenko}, \bibinfo{person}{M.~Hardt}, {and} \bibinfo{person}{S.~Levine}} (Eds.), Vol.~\bibinfo{volume}{36}. \bibinfo{publisher}{Curran Associates, Inc.}, \bibinfo{pages}{13022--13063}.
\newblock
\urldef\tempurl%
\url{https://proceedings.neurips.cc/paper_files/paper/2023/file/2a514213ba899f2911723a38be8d4096-Paper-Conference.pdf}
\showURL{%
\tempurl}


\bibitem[Ohsaka et~al\mbox{.}(2017)]%
        {Gcoarse-infan-2}
\bibfield{author}{\bibinfo{person}{Naoto Ohsaka}, \bibinfo{person}{Tomohiro Sonobe}, \bibinfo{person}{Sumio Fujita}, {and} \bibinfo{person}{Ken-ichi Kawarabayashi}.} \bibinfo{year}{2017}\natexlab{}.
\newblock \showarticletitle{Coarsening Massive Influence Networks for Scalable Diffusion Analysis}. In \bibinfo{booktitle}{\emph{Proceedings of the 2017 ACM International Conference on Management of Data}} (Chicago, Illinois, USA) \emph{(\bibinfo{series}{SIGMOD '17})}. \bibinfo{publisher}{Association for Computing Machinery}, \bibinfo{address}{New York, NY, USA}, \bibinfo{pages}{635–650}.
\newblock
\showISBNx{9781450341974}
\href{https://doi.org/10.1145/3035918.3064045}{doi:\nolinkurl{10.1145/3035918.3064045}}


\bibitem[Ortega et~al\mbox{.}(2018)]%
        {GraphSignalProcessingOverviewChallengesandApplications}
\bibfield{author}{\bibinfo{person}{Antonio Ortega}, \bibinfo{person}{Pascal Frossard}, \bibinfo{person}{Jelena Kovačević}, \bibinfo{person}{José M.~F. Moura}, {and} \bibinfo{person}{Pierre Vandergheynst}.} \bibinfo{year}{2018}\natexlab{}.
\newblock \showarticletitle{Graph Signal Processing: Overview, Challenges, and Applications}.
\newblock \bibinfo{journal}{\emph{Proc. IEEE}} \bibinfo{volume}{106}, \bibinfo{number}{5} (\bibinfo{year}{2018}), \bibinfo{pages}{808--828}.
\newblock
\href{https://doi.org/10.1109/JPROC.2018.2820126}{doi:\nolinkurl{10.1109/JPROC.2018.2820126}}


\bibitem[Pei et~al\mbox{.}(2020)]%
        {dataset3-pei}
\bibfield{author}{\bibinfo{person}{Hongbin Pei}, \bibinfo{person}{Bingzhe Wei}, \bibinfo{person}{Kevin Chen-Chuan Chang}, \bibinfo{person}{Yu Lei}, {and} \bibinfo{person}{Bo Yang}.} \bibinfo{year}{2020}\natexlab{}.
\newblock \showarticletitle{Geom-GCN: Geometric Graph Convolutional Networks}. In \bibinfo{booktitle}{\emph{International Conference on Learning Representations}}.
\newblock
\urldef\tempurl%
\url{https://openreview.net/forum?id=S1e2agrFvS}
\showURL{%
\tempurl}


\bibitem[Phillips(2003)]%
        {polyapprox_2}
\bibfield{author}{\bibinfo{person}{George~M Phillips}.} \bibinfo{year}{2003}\natexlab{}.
\newblock \bibinfo{booktitle}{\emph{Interpolation and approximation by polynomials}}. Vol.~\bibinfo{volume}{14}.
\newblock \bibinfo{publisher}{Springer New York}.
\newblock
\showISBNx{978-0-387-00215-6}
\href{https://doi.org/10.1007/b97417}{doi:\nolinkurl{10.1007/b97417}}


\bibitem[Platonov et~al\mbox{.}(2023)]%
        {dataset8-small-hetero}
\bibfield{author}{\bibinfo{person}{Oleg Platonov}, \bibinfo{person}{Denis Kuznedelev}, \bibinfo{person}{Michael Diskin}, \bibinfo{person}{Artem Babenko}, {and} \bibinfo{person}{Liudmila Prokhorenkova}.} \bibinfo{year}{2023}\natexlab{}.
\newblock \showarticletitle{A critical look at the evaluation of {GNN}s under heterophily: Are we really making progress?}. In \bibinfo{booktitle}{\emph{The Eleventh International Conference on Learning Representations}}.
\newblock
\urldef\tempurl%
\url{https://openreview.net/forum?id=tJbbQfw-5wv}
\showURL{%
\tempurl}


\bibitem[Purohit et~al\mbox{.}(2014)]%
        {Gcoarse-infan-1}
\bibfield{author}{\bibinfo{person}{Manish Purohit}, \bibinfo{person}{B.~Aditya Prakash}, \bibinfo{person}{Chanhyun Kang}, \bibinfo{person}{Yao Zhang}, {and} \bibinfo{person}{V.S. Subrahmanian}.} \bibinfo{year}{2014}\natexlab{}.
\newblock \showarticletitle{Fast influence-based coarsening for large networks} \emph{(\bibinfo{series}{KDD '14})}. \bibinfo{publisher}{Association for Computing Machinery}, \bibinfo{address}{New York, NY, USA}, \bibinfo{pages}{1296–1305}.
\newblock
\showISBNx{9781450329569}
\href{https://doi.org/10.1145/2623330.2623701}{doi:\nolinkurl{10.1145/2623330.2623701}}


\bibitem[Qiao et~al\mbox{.}(2024)]%
        {GAnoDet-survey-2}
\bibfield{author}{\bibinfo{person}{Hezhe Qiao}, \bibinfo{person}{Hanghang Tong}, \bibinfo{person}{Bo An}, \bibinfo{person}{Irwin King}, \bibinfo{person}{Charu Aggarwal}, {and} \bibinfo{person}{Guansong Pang}.} \bibinfo{year}{2024}\natexlab{}.
\newblock \bibinfo{title}{Deep Graph Anomaly Detection: A Survey and New Perspectives}.
\newblock
\showeprint[arxiv]{2409.09957}~[cs.LG]
\urldef\tempurl%
\url{https://arxiv.org/abs/2409.09957}
\showURL{%
\tempurl}


\bibitem[Rivlin(2020)]%
        {poly_chebyshev}
\bibfield{author}{\bibinfo{person}{Theodore~J Rivlin}.} \bibinfo{year}{2020}\natexlab{}.
\newblock \bibinfo{booktitle}{\emph{Chebyshev polynomials}}.
\newblock \bibinfo{publisher}{Courier Dover Publications}.
\newblock


\bibitem[Sandryhaila and Moura(2013a)]%
        {DiscreteSignalProcessingonGraphs}
\bibfield{author}{\bibinfo{person}{Aliaksei Sandryhaila} {and} \bibinfo{person}{José M.~F. Moura}.} \bibinfo{year}{2013}\natexlab{a}.
\newblock \showarticletitle{Discrete Signal Processing on Graphs}.
\newblock \bibinfo{journal}{\emph{IEEE Transactions on Signal Processing}} \bibinfo{volume}{61}, \bibinfo{number}{7} (\bibinfo{year}{2013}), \bibinfo{pages}{1644--1656}.
\newblock
\href{https://doi.org/10.1109/TSP.2013.2238935}{doi:\nolinkurl{10.1109/TSP.2013.2238935}}


\bibitem[Sandryhaila and Moura(2013b)]%
        {DiscretesignalprocessingongraphsGraphfilters}
\bibfield{author}{\bibinfo{person}{Aliaksei Sandryhaila} {and} \bibinfo{person}{José M.~F. Moura}.} \bibinfo{year}{2013}\natexlab{b}.
\newblock \showarticletitle{Discrete signal processing on graphs: Graph filters}. In \bibinfo{booktitle}{\emph{2013 IEEE International Conference on Acoustics, Speech and Signal Processing}}. \bibinfo{pages}{6163--6166}.
\newblock
\href{https://doi.org/10.1109/ICASSP.2013.6638849}{doi:\nolinkurl{10.1109/ICASSP.2013.6638849}}


\bibitem[Sandryhaila and Moura(2013c)]%
        {DiscretesignalprocessingongraphsGraphfouriertransform}
\bibfield{author}{\bibinfo{person}{Aliaksei Sandryhaila} {and} \bibinfo{person}{José M.~F. Moura}.} \bibinfo{year}{2013}\natexlab{c}.
\newblock \showarticletitle{Discrete signal processing on graphs: Graph fourier transform}. In \bibinfo{booktitle}{\emph{2013 IEEE International Conference on Acoustics, Speech and Signal Processing}}. \bibinfo{pages}{6167--6170}.
\newblock
\href{https://doi.org/10.1109/ICASSP.2013.6638850}{doi:\nolinkurl{10.1109/ICASSP.2013.6638850}}


\bibitem[Smyth(1998)]%
        {polyapprox_1}
\bibfield{author}{\bibinfo{person}{Gordon~K Smyth}.} \bibinfo{year}{1998}\natexlab{}.
\newblock \showarticletitle{Polynomial approximation}.
\newblock \bibinfo{journal}{\emph{Encyclopedia of Biostatistics}}  \bibinfo{volume}{13} (\bibinfo{year}{1998}).
\newblock


\bibitem[Song et~al\mbox{.}(2023)]%
        {OrderedGNN}
\bibfield{author}{\bibinfo{person}{Yunchong Song}, \bibinfo{person}{Chenghu Zhou}, \bibinfo{person}{Xinbing Wang}, {and} \bibinfo{person}{Zhouhan Lin}.} \bibinfo{year}{2023}\natexlab{}.
\newblock \showarticletitle{Ordered {GNN}: Ordering Message Passing to Deal with Heterophily and Over-smoothing}. In \bibinfo{booktitle}{\emph{The Eleventh International Conference on Learning Representations}}.
\newblock
\urldef\tempurl%
\url{https://openreview.net/forum?id=wKPmPBHSnT6}
\showURL{%
\tempurl}


\bibitem[Strang(2006)]%
        {linearalgebra}
\bibfield{author}{\bibinfo{person}{Gilbert Strang}.} \bibinfo{year}{2006}\natexlab{}.
\newblock \bibinfo{booktitle}{\emph{Linear algebra and its applications.}}
\newblock \bibinfo{publisher}{Belmont, CA: Thomson, Brooks/Cole}.
\newblock


\bibitem[Sun et~al\mbox{.}(2023)]%
        {decoupled-FEGNN}
\bibfield{author}{\bibinfo{person}{Jiaqi Sun}, \bibinfo{person}{Lin Zhang}, \bibinfo{person}{Guangyi Chen}, \bibinfo{person}{Peng Xu}, \bibinfo{person}{Kun Zhang}, {and} \bibinfo{person}{Yujiu Yang}.} \bibinfo{year}{2023}\natexlab{}.
\newblock \showarticletitle{Feature Expansion for Graph Neural Networks}. In \bibinfo{booktitle}{\emph{Proceedings of the 40th International Conference on Machine Learning}} \emph{(\bibinfo{series}{Proceedings of Machine Learning Research}, Vol.~\bibinfo{volume}{202})}. \bibinfo{publisher}{PMLR}, \bibinfo{pages}{33156--33176}.
\newblock


\bibitem[Tang et~al\mbox{.}(2022)]%
        {GAnoDet-spectral-1-BWGNN}
\bibfield{author}{\bibinfo{person}{Jianheng Tang}, \bibinfo{person}{Jiajin Li}, \bibinfo{person}{Ziqi Gao}, {and} \bibinfo{person}{Jia Li}.} \bibinfo{year}{2022}\natexlab{}.
\newblock \showarticletitle{Rethinking Graph Neural Networks for Anomaly Detection}. In \bibinfo{booktitle}{\emph{Proceedings of the 39th International Conference on Machine Learning}} \emph{(\bibinfo{series}{Proceedings of Machine Learning Research}, Vol.~\bibinfo{volume}{162})}, \bibfield{editor}{\bibinfo{person}{Kamalika Chaudhuri}, \bibinfo{person}{Stefanie Jegelka}, \bibinfo{person}{Le~Song}, \bibinfo{person}{Csaba Szepesvari}, \bibinfo{person}{Gang Niu}, {and} \bibinfo{person}{Sivan Sabato}} (Eds.). \bibinfo{publisher}{PMLR}, \bibinfo{pages}{21076--21089}.
\newblock
\urldef\tempurl%
\url{https://proceedings.mlr.press/v162/tang22b.html}
\showURL{%
\tempurl}


\bibitem[Wang and Zhang(2022)]%
        {JacobiConv}
\bibfield{author}{\bibinfo{person}{Xiyuan Wang} {and} \bibinfo{person}{Muhan Zhang}.} \bibinfo{year}{2022}\natexlab{}.
\newblock \showarticletitle{How Powerful are Spectral Graph Neural Networks}. In \bibinfo{booktitle}{\emph{Proceedings of the 39th International Conference on Machine Learning}} \emph{(\bibinfo{series}{Proceedings of Machine Learning Research}, Vol.~\bibinfo{volume}{162})}, \bibfield{editor}{\bibinfo{person}{Kamalika Chaudhuri}, \bibinfo{person}{Stefanie Jegelka}, \bibinfo{person}{Le~Song}, \bibinfo{person}{Csaba Szepesvari}, \bibinfo{person}{Gang Niu}, {and} \bibinfo{person}{Sivan Sabato}} (Eds.). \bibinfo{publisher}{PMLR}, \bibinfo{pages}{23341--23362}.
\newblock
\urldef\tempurl%
\url{https://proceedings.mlr.press/v162/wang22am.html}
\showURL{%
\tempurl}


\bibitem[Wu et~al\mbox{.}(2019)]%
        {SGC}
\bibfield{author}{\bibinfo{person}{Felix Wu}, \bibinfo{person}{Amauri Souza}, \bibinfo{person}{Tianyi Zhang}, \bibinfo{person}{Christopher Fifty}, \bibinfo{person}{Tao Yu}, {and} \bibinfo{person}{Kilian Weinberger}.} \bibinfo{year}{2019}\natexlab{}.
\newblock \showarticletitle{Simplifying Graph Convolutional Networks}. In \bibinfo{booktitle}{\emph{Proceedings of the 36th International Conference on Machine Learning}}, Vol.~\bibinfo{volume}{97}. \bibinfo{publisher}{PMLR}, \bibinfo{pages}{6861--6871}.
\newblock
\urldef\tempurl%
\url{https://proceedings.mlr.press/v97/wu19e.html}
\showURL{%
\tempurl}


\bibitem[Wu et~al\mbox{.}(2022)]%
        {nodeformer}
\bibfield{author}{\bibinfo{person}{Qitian Wu}, \bibinfo{person}{Wentao Zhao}, \bibinfo{person}{Zenan Li}, \bibinfo{person}{David Wipf}, {and} \bibinfo{person}{Junchi Yan}.} \bibinfo{year}{2022}\natexlab{}.
\newblock \showarticletitle{NodeFormer: A Scalable Graph Structure Learning Transformer for Node Classification}. In \bibinfo{booktitle}{\emph{Advances in Neural Information Processing Systems}}, \bibfield{editor}{\bibinfo{person}{Alice~H. Oh}, \bibinfo{person}{Alekh Agarwal}, \bibinfo{person}{Danielle Belgrave}, {and} \bibinfo{person}{Kyunghyun Cho}} (Eds.).
\newblock
\urldef\tempurl%
\url{https://openreview.net/forum?id=sMezXGG5So}
\showURL{%
\tempurl}


\bibitem[Wu et~al\mbox{.}(2021)]%
        {comprehensivegnn}
\bibfield{author}{\bibinfo{person}{Zonghan Wu}, \bibinfo{person}{Shirui Pan}, \bibinfo{person}{Fengwen Chen}, \bibinfo{person}{Guodong Long}, \bibinfo{person}{Chengqi Zhang}, {and} \bibinfo{person}{Philip~S. Yu}.} \bibinfo{year}{2021}\natexlab{}.
\newblock \showarticletitle{A Comprehensive Survey on Graph Neural Networks}.
\newblock \bibinfo{journal}{\emph{IEEE Transactions on Neural Networks and Learning Systems}} \bibinfo{volume}{32}, \bibinfo{number}{1} (\bibinfo{year}{2021}), \bibinfo{pages}{4--24}.
\newblock
\href{https://doi.org/10.1109/TNNLS.2020.2978386}{doi:\nolinkurl{10.1109/TNNLS.2020.2978386}}


\bibitem[Xia et~al\mbox{.}(2021)]%
        {recsys_2}
\bibfield{author}{\bibinfo{person}{Lianghao Xia}, \bibinfo{person}{Yong Xu}, \bibinfo{person}{Chao Huang}, \bibinfo{person}{Peng Dai}, {and} \bibinfo{person}{Liefeng Bo}.} \bibinfo{year}{2021}\natexlab{}.
\newblock \showarticletitle{Graph Meta Network for Multi-Behavior Recommendation}. In \bibinfo{booktitle}{\emph{The 44th International ACM SIGIR Conference on Research and Development in Information Retrieval}}. \bibinfo{pages}{757–766}.
\newblock
\href{https://doi.org/10.1145/3404835.3462972}{doi:\nolinkurl{10.1145/3404835.3462972}}


\bibitem[Yang et~al\mbox{.}(2016)]%
        {dataset1-cora}
\bibfield{author}{\bibinfo{person}{Zhilin Yang}, \bibinfo{person}{William~W. Cohen}, {and} \bibinfo{person}{Ruslan Salakhutdinov}.} \bibinfo{year}{2016}\natexlab{}.
\newblock \bibinfo{title}{Revisiting Semi-Supervised Learning with Graph Embeddings}.
\newblock
\href{https://doi.org/10.48550/ARXIV.1603.08861}{doi:\nolinkurl{10.48550/ARXIV.1603.08861}}


\bibitem[Zhang et~al\mbox{.}(2019)]%
        {websearch_1}
\bibfield{author}{\bibinfo{person}{Yuan Zhang}, \bibinfo{person}{Dong Wang}, {and} \bibinfo{person}{Yan Zhang}.} \bibinfo{year}{2019}\natexlab{}.
\newblock \showarticletitle{Neural IR Meets Graph Embedding: A Ranking Model for Product Search}. In \bibinfo{booktitle}{\emph{The World Wide Web Conference}} (San Francisco, CA, USA) \emph{(\bibinfo{series}{WWW '19})}. \bibinfo{publisher}{Association for Computing Machinery}, \bibinfo{address}{New York, NY, USA}, \bibinfo{pages}{2390–2400}.
\newblock
\showISBNx{9781450366748}
\href{https://doi.org/10.1145/3308558.3313468}{doi:\nolinkurl{10.1145/3308558.3313468}}


\bibitem[Zheng et~al\mbox{.}(2022)]%
        {Gcoarse-gl-3}
\bibfield{author}{\bibinfo{person}{Da Zheng}, \bibinfo{person}{Xiang Song}, \bibinfo{person}{Chengru Yang}, \bibinfo{person}{Dominique LaSalle}, {and} \bibinfo{person}{George Karypis}.} \bibinfo{year}{2022}\natexlab{}.
\newblock \showarticletitle{Distributed Hybrid CPU and GPU training for Graph Neural Networks on Billion-Scale Heterogeneous Graphs}. In \bibinfo{booktitle}{\emph{Proceedings of the 28th ACM SIGKDD Conference on Knowledge Discovery and Data Mining}} (Washington DC, USA) \emph{(\bibinfo{series}{KDD '22})}. \bibinfo{publisher}{Association for Computing Machinery}, \bibinfo{address}{New York, NY, USA}, \bibinfo{pages}{4582–4591}.
\newblock
\showISBNx{9781450393850}
\href{https://doi.org/10.1145/3534678.3539177}{doi:\nolinkurl{10.1145/3534678.3539177}}


\bibitem[Zheng et~al\mbox{.}(2023)]%
        {decoupled-NFGNN-nodewise}
\bibfield{author}{\bibinfo{person}{Shuai Zheng}, \bibinfo{person}{Zhenfeng Zhu}, \bibinfo{person}{Zhizhe Liu}, \bibinfo{person}{Youru Li}, {and} \bibinfo{person}{Yao Zhao}.} \bibinfo{year}{2023}\natexlab{}.
\newblock \showarticletitle{Node-Oriented Spectral Filtering for Graph Neural Networks}.
\newblock \bibinfo{journal}{\emph{IEEE Transactions on Pattern Analysis and Machine Intelligence}} \bibinfo{volume}{46}, \bibinfo{number}{1} (\bibinfo{year}{2023}), \bibinfo{pages}{388--402}.
\newblock
\href{https://doi.org/10.1109/TPAMI.2023.3324937}{doi:\nolinkurl{10.1109/TPAMI.2023.3324937}}


\bibitem[Zheng et~al\mbox{.}(2024)]%
        {heterophily-gnn-survey}
\bibfield{author}{\bibinfo{person}{Xin Zheng}, \bibinfo{person}{Yi Wang}, \bibinfo{person}{Yixin Liu}, \bibinfo{person}{Ming Li}, \bibinfo{person}{Miao Zhang}, \bibinfo{person}{Di Jin}, \bibinfo{person}{Philip~S. Yu}, {and} \bibinfo{person}{Shirui Pan}.} \bibinfo{year}{2024}\natexlab{}.
\newblock \bibinfo{title}{Graph Neural Networks for Graphs with Heterophily: A Survey}.
\newblock
\showeprint[arxiv]{2202.07082}~[cs.LG]
\urldef\tempurl%
\url{https://arxiv.org/abs/2202.07082}
\showURL{%
\tempurl}


\bibitem[Zhou et~al\mbox{.}(2022)]%
        {gnn-survey}
\bibfield{author}{\bibinfo{person}{Yu Zhou}, \bibinfo{person}{Haixia Zheng}, \bibinfo{person}{Xin Huang}, \bibinfo{person}{Shufeng Hao}, \bibinfo{person}{Dengao Li}, {and} \bibinfo{person}{Jumin Zhao}.} \bibinfo{year}{2022}\natexlab{}.
\newblock \showarticletitle{Graph Neural Networks: Taxonomy, Advances, and Trends}.
\newblock \bibinfo{journal}{\emph{ACM Transactions on Intelligent Systems and Technology}} \bibinfo{volume}{13}, \bibinfo{number}{1}, Article \bibinfo{articleno}{15} (\bibinfo{date}{jan} \bibinfo{year}{2022}), \bibinfo{numpages}{54}~pages.
\newblock
\showISSN{2157-6904}
\href{https://doi.org/10.1145/3495161}{doi:\nolinkurl{10.1145/3495161}}


\bibitem[Zhu et~al\mbox{.}(2020)]%
        {H2GCN}
\bibfield{author}{\bibinfo{person}{Jiong Zhu}, \bibinfo{person}{Yujun Yan}, \bibinfo{person}{Lingxiao Zhao}, \bibinfo{person}{Mark Heimann}, \bibinfo{person}{Leman Akoglu}, {and} \bibinfo{person}{Danai Koutra}.} \bibinfo{year}{2020}\natexlab{}.
\newblock \showarticletitle{Beyond Homophily in Graph Neural Networks: Current Limitations and Effective Designs}. In \bibinfo{booktitle}{\emph{Advances in Neural Information Processing Systems}}, Vol.~\bibinfo{volume}{33}. \bibinfo{publisher}{Curran Associates, Inc.}, \bibinfo{pages}{7793--7804}.
\newblock
\urldef\tempurl%
\url{https://proceedings.neurips.cc/paper_files/paper/2020/file/58ae23d878a47004366189884c2f8440-Paper.pdf}
\showURL{%
\tempurl}


\end{thebibliography}


\clearpage
\appendix
\onecolumn

\begin{center}
\textbf{\fontsize{20pt}{15pt}\selectfont Appendix}
\end{center}


\section{Proof of Proposition~\ref{proposition:optimal-is-hybrid}}
\label{appendix:proposition:optimal-is-hybrid}

Proposition~\ref{proposition:optimal-is-hybrid}: \textit{
Consider a binary-class graph $\mathcal{G}$ generated from the distribution $CSBM(n, \boldsymbol{\mu}, \boldsymbol{\nu}, (p_0, q_0), (p_1, q_1), P)$, as defined in~\cite{nodewise-3}. Here, $\boldsymbol{\mu}$ and $\boldsymbol{\nu}$ define Gaussian distributions for the generation of random node features, while $(p_0, q_0)$, $(p_1, q_1)$, and $P$ control the graph structure’s statistical properties.}

\textit{
Define $\mathcal{V}_{\text{homo}}$ and $\mathcal{V}_{\text{hete}}$ as the homophilic and heterophilic node subsets, respectively. A minimal-parameter solution for achieving linear separability across all nodes is attainable using a uniform (graph-wise) filter for $\mathcal{V}_{\text{homo}}$ and node-specific filters for $\mathcal{V}_{\text{hete}}$.
}
\begin{proof}
\label{proof:proposition:optimal-is-hybrid}
First, according to Theorem 1 in~\cite{nodewise-3}, the following statements hold:

\begin{itemize}[leftmargin=25pt,parsep=2pt,itemsep=2pt,topsep=2pt]
    \item In a binary-class homophilic graph $\mathcal{G}$, a uniform filter suffices to achieve linear separability.
    \item In a binary-class heterophilic graph $\mathcal{G}$, employing distinct filters for each node guarantees linear separability.
\end{itemize}

Thus, we examine the subsets $\mathcal{V}_{\text{homo}}$ and $\mathcal{V}_{\text{hete}}$ separately. 
Let $\boldsymbol{Y}$ represent the true node labels,

\subsubsection*{\bf Homophilic Subset $\mathcal{V}_{\text{homo}}$} We define a filter with coefficient $\theta$ that achieves linear separability on $\mathcal{V}_{\text{homo}}$, leading to

\begin{equation}
\label{eq:proof:proposition:optimal-is-hybrid-1}
\boldsymbol{Y}_{\mathcal{V}_{\text{homo}}} = \boldsymbol{Z}_{\mathcal{V}_{\text{homo}}} = \sum_{p\in\mathcal{V}_{\text{homo}}} 
\frac{\mathrm{diag}(\boldsymbol{\delta}_{p})}{|\mathcal{V}_{\text{homo}}|} \sum_{k}\theta_{k}\mathbf{T}_{k}(\boldsymbol{L})\boldsymbol{X}
\end{equation}

\subsubsection*{\bf Heterophilic Subset $\mathcal{V}_{\text{hete}}$} Analogously, we define a filter coefficient set $\{\theta^{\ast}_{1}, \theta^{\ast}_{2},...,\theta^{\ast}_{|\mathcal{V}_{\text{hete}}|}\}$ that guarantees linear separability on $\mathcal{V}_{\text{hete}}$, resulting in 

\begin{equation}
\label{eq:proof:proposition:optimal-is-hybrid-2}
\boldsymbol{Y}_{q} = \boldsymbol{Z}_{q} = 
\frac{\mathrm{diag}(\boldsymbol{\delta}_{q})}{|\mathcal{V}_{\text{hete}}|} \sum_{k}\theta^{\ast}_{qk}\mathbf{T}_{k}(\boldsymbol{L})\boldsymbol{X}\ ,\ q = 1,2,...,|\mathcal{V}_{\text{hete}}|\ .
\end{equation}

Thus, by integrating Eq.~\ref{eq:proof:proposition:optimal-is-hybrid-1} and Eq.~\ref{eq:proof:proposition:optimal-is-hybrid-2}, we derive

\begin{align}
\label{eq:proof:proposition:optimal-is-hybrid-3}
\boldsymbol{Y} & = \boldsymbol{Y}_{\mathcal{V}_{\text{homo}}} + \sum_{q\in\mathcal{V}_{\text{hete}}}\boldsymbol{Y}_{q} = \boldsymbol{Z}_{\mathcal{V}_{\text{homo}}} + \sum_{q\in\mathcal{V}_{\text{hete}}}\boldsymbol{Z}_{q} \notag \\
& = \sum_{k} \left( \sum_{p\in\mathcal{V}_{\text{homo}}}\frac{\theta_{k} \cdot \mathrm{diag}(\boldsymbol{\delta}_{p})}{|\mathcal{V}_{\text{homo}}|} + \sum_{q\in\mathcal{V}_{\text{hete}}} \frac{\theta^{\ast}_{qk} \cdot \mathrm{diag}(\boldsymbol{\delta}_{q})}{|\mathcal{V}_{\text{hete}}|} \right) \mathbf{T}_{k}(\boldsymbol{L})\boldsymbol{X}\ ,
\end{align}

thus completing the proof.

\end{proof}


\section{Proof of Theorem~\ref{theorem:message-passing-guarantee}}
\label{appendix:theorem:message-passing-guarantee}

Theorem~\ref{theorem:message-passing-guarantee}: \textit{
Let $\boldsymbol{\Delta}=\boldsymbol{L}-\boldsymbol{I}$ and $\boldsymbol{\Pi}=\boldsymbol{C}^{+}\boldsymbol{C}$. 
For the $k$-th order term of the polynomial-based operator, denoted as $\boldsymbol{\Delta}^{k}$, the filtered signal $\boldsymbol{\Delta}^{k}\boldsymbol{x}$ satisfies the following inequality:
\begin{equation}
\lVert \boldsymbol{\Delta}^{k}\boldsymbol{x} - \boldsymbol{\Pi}\boldsymbol{\Delta}\boldsymbol{\Pi}\boldsymbol{\Delta}^{k-1}\boldsymbol{x} \rVert_{\boldsymbol{L}}  \leq \epsilon_{\boldsymbol{L}, \boldsymbol{C}, \mathcal{R}} \Vert \boldsymbol{x} \Vert_{\boldsymbol{L}} \left( \lVert \boldsymbol{\Delta} \rVert_{\boldsymbol{L}} + \lVert \boldsymbol{\Pi}\boldsymbol{\Delta} \rVert_{\boldsymbol{L}} \right)\ .
\end{equation}
Furthermore, when the coarsening algorithm achieves a sufficiently small RSA constant $\epsilon_{\boldsymbol{L}, \boldsymbol{C}, \mathcal{R}} \rightarrow 0$, the following approximation result holds:
\begin{equation}
\lVert \boldsymbol{\Delta}^{k}\boldsymbol{x} - (\boldsymbol{\Pi}\boldsymbol{\Delta}\boldsymbol{\Pi})^{k}\boldsymbol{x} \rVert_{\boldsymbol{L}} \rightarrow 0.
\end{equation}
}
\begin{proof}
\label{proof:message-passing-guarantee}
The proof is carried out in two logical stages:

\subsubsection*{\bf Proof of the inequality} Referring to the theorem in~\cite{rsa-Gcoarse-7} (see the section 3), the coarsening algorithm that guarantees a low RSA constant satisfies the following inequality:
\begin{equation}
\label{eq:message-passing-origin}
\lVert \boldsymbol{\Delta}\boldsymbol{x} - \boldsymbol{\Pi}\boldsymbol{\Delta}\boldsymbol{\Pi}\boldsymbol{x} \rVert_{\boldsymbol{L}}  \leq \epsilon_{\boldsymbol{L}, \boldsymbol{C}, \mathcal{R}} \Vert x\Vert_{\boldsymbol{L}} \left( \lVert \boldsymbol{\Delta} \rVert_{\boldsymbol{L}} + \lVert \boldsymbol{\Pi}\boldsymbol{\Delta} \rVert_{\boldsymbol{L}} \right)\ .
\end{equation}
Notice that $\boldsymbol{x}$ is assumed to be in $\mathbb{R}^{n}$, we replace $\boldsymbol{x}$ in Eq.~\eqref{eq:message-passing-origin} with $\boldsymbol{\Delta}^{k-1}\boldsymbol{x}$, leading to the following derivation:
\begin{equation}
\label{eq:message-passing-derivation-1}
\lVert \boldsymbol{\Delta}^{k}\boldsymbol{x} - \boldsymbol{\Pi}\boldsymbol{\Delta}\boldsymbol{\Pi}\boldsymbol{\Delta}^{k-1}\boldsymbol{x} \rVert_{\boldsymbol{L}}  \leq \epsilon_{\boldsymbol{L}, \boldsymbol{C}, \mathcal{R}} \Vert \boldsymbol{\Delta}^{k-1}\boldsymbol{x} \Vert_{\boldsymbol{L}} \left( \lVert \boldsymbol{\Delta} \rVert_{\boldsymbol{L}} + \lVert \boldsymbol{\Pi}\boldsymbol{\Delta} \rVert_{\boldsymbol{L}} \right)\ .
\end{equation}
Let $\lambda_{\text{max}}=\max\vert\lambda(\boldsymbol{\Delta}^{k-1})\vert$ denote the maximum absolute eigenvalue of $\boldsymbol{\Delta}^{k-1}$. 
From the spectral graph theory~\cite{spectralgraphtheory}, we know that $\lambda(\boldsymbol{\Delta}^{k-1})=\lambda^{k-1}(\boldsymbol{\Delta})$, with $-1 \leq \lambda(\boldsymbol{\Delta}) < 1$, resulting in the inequality below:
\begin{equation}
\label{eq:message-passing-derivation-2}
0 < \lambda_{\text{max}}=\max\vert\lambda^{k-1}(\boldsymbol{\Delta})\vert \leq 1\ .
\end{equation}
Thus, by recalling the Cauchy-Schwarz inequality~\cite{linearalgebra}, we can further derive that:
\begin{equation}
\label{eq:message-passing-derivation-3}
\Vert \boldsymbol{\Delta}^{k-1}\boldsymbol{x} \Vert_{\boldsymbol{L}} \leq \lambda_{\text{max}} \Vert \boldsymbol{x} \Vert_{\boldsymbol{L}} \leq \Vert \boldsymbol{x} \Vert_{\boldsymbol{L}}\ .
\end{equation}
Incorporating Eq.~\eqref{eq:message-passing-derivation-1} and Eq.~\eqref{eq:message-passing-derivation-3} yields the following formulation:
\begin{equation}
\label{eq:message-passing-derivation-4}
\lVert \boldsymbol{\Delta}^{k}\boldsymbol{x} - \boldsymbol{\Pi}\boldsymbol{\Delta}\boldsymbol{\Pi}\boldsymbol{\Delta}^{k-1}\boldsymbol{x} \rVert_{\boldsymbol{L}}  \leq \epsilon_{\boldsymbol{L}, \boldsymbol{C}, \mathcal{R}} \Vert \boldsymbol{x} \Vert_{\boldsymbol{L}} \left( \lVert \boldsymbol{\Delta} \rVert_{\boldsymbol{L}} + \lVert \boldsymbol{\Pi}\boldsymbol{\Delta} \rVert_{\boldsymbol{L}} \right)\ .
\end{equation}

\subsubsection*{\bf Proof of the approximation relation} We prove this approximation relation through mathematical induction. 
Specifically, 

(\rom{1})~For $k=1$, given that $\epsilon_{\boldsymbol{L}, \boldsymbol{C}, \mathcal{R}} \rightarrow 0$, we obtain the following:
\begin{equation}
\label{eq:message-passing-derivation-5}
\lVert \boldsymbol{\Delta}\boldsymbol{x} - (\boldsymbol{\Pi}\boldsymbol{\Delta}\boldsymbol{\Pi})\boldsymbol{x} \rVert_{\boldsymbol{L}} \rightarrow 0\ ,
\end{equation}
indicating that the approximation holds for $k=1$. 

(\rom{2})~Assume the approximation holds for $k=n$ where $n\in\mathbb{Z}$ and $n > 1$, yielding the following result:
\begin{equation}
\label{eq:message-passing-derivation-6}
\lVert \boldsymbol{\Delta}^{n}\boldsymbol{x} - (\boldsymbol{\Pi}\boldsymbol{\Delta}\boldsymbol{\Pi})^{n}\boldsymbol{x} \rVert_{\boldsymbol{L}} \rightarrow 0.
\end{equation}
Building on this, we proceed to prove that the approximation holds for $k=n+1$. 
In particular, we have the following relation:
\begin{equation}
\label{eq:message-passing-derivation-7}
\lVert \boldsymbol{\Delta}^{n+1}\boldsymbol{x} - (\boldsymbol{\Pi}\boldsymbol{\Delta}\boldsymbol{\Pi})^{n+1}\boldsymbol{x} \rVert_{\boldsymbol{L}} \overset{\text{Eq.~\eqref{eq:message-passing-derivation-6}}}{\rightarrow} \lVert \boldsymbol{\Delta}^{n+1}\boldsymbol{x} - \boldsymbol{\Pi}\boldsymbol{\Delta}\boldsymbol{\Pi}\boldsymbol{\Delta}^{n}\boldsymbol{x} \rVert_{\boldsymbol{L}}\ .
\end{equation}
Notice that the right-hand side of the equation above satisfies the established inequality:
\begin{equation}
\label{eq:message-passing-derivation-8}
\lVert \boldsymbol{\Delta}^{n+1}\boldsymbol{x} - \boldsymbol{\Pi}\boldsymbol{\Delta}\boldsymbol{\Pi}\boldsymbol{\Delta}^{n}\boldsymbol{x} \rVert_{\boldsymbol{L}} \leq \epsilon_{\boldsymbol{L}, \boldsymbol{C}, \mathcal{R}} \Vert x\Vert_{\boldsymbol{L}} \left( \lVert \boldsymbol{\Delta} \rVert_{\boldsymbol{L}} + \lVert \boldsymbol{\Pi}\boldsymbol{\Delta} \rVert_{\boldsymbol{L}} \right) \rightarrow 0\ ,
\end{equation}
confirming that the approximation holds for $k=n+1$. 

Therefore, based on the above induction, we establish that the approximation relation holds for $k=1,2,...$, thereby completing the proof.
\end{proof}


\section{Step-by-Step Derivation from Eq.~\ref{eq:partion-filtering} to Eq.~\ref{eq:partion-filtering-unified}}
\label{appendix:derivation}

We begin by deriving Eq.~\ref{eq:partion-filtering} as follows:

\begin{equation}
\label{eq:appendix:derivation-1}
\boldsymbol{Z}_{\mathcal{V}_{i}} = \sum_{k=0}^{K} \sum_{m\in\mathcal{V}_{i}} \frac{\theta_{ik} \mathrm{diag}(\boldsymbol{\delta}_{m})}{|\mathcal{V}_{i}|} \mathbf{T}_{k}(\boldsymbol{L})\boldsymbol{X}\ .
\end{equation}

Since $\boldsymbol{Z}$ is given by the sum over all partitions as $\boldsymbol{Z} = \sum_{i=1}^{n^{\prime}} \boldsymbol{Z}_{\mathcal{V}_{i}}$, we obtain:

\begin{align}
\label{eq:appendix:derivation-2}
\boldsymbol{Z} & = \sum_{i=1}^{n^{\prime}} \sum_{k=0}^{K} \sum_{m\in\mathcal{V}_{i}} \frac{\theta_{ik} \mathrm{diag}(\boldsymbol{\delta}_{m})}{|\mathcal{V}_{i}|} \mathbf{T}_{k}(\boldsymbol{L})\boldsymbol{X} \notag \\
& = \sum_{k=0}^{K} \underbrace{\left( \sum_{i=1}^{n^{\prime}} \sum_{m\in\mathcal{V}_{i}} \frac{\theta_{ik} \mathrm{diag}(\boldsymbol{\delta}_{m})}{|\mathcal{V}_{i}|} \right)}_{\text{formulation $k$}} \mathbf{T}_{k}(\boldsymbol{L})\boldsymbol{X}\ .
\end{align}

By introducing the matrix $\boldsymbol{\Theta} \in \mathbb{R}^{n^{\prime} \times (K+1)}$, where $\boldsymbol{\Theta}_{ik} = \theta_{ik} / |\mathcal{V}_{i}|$, we rewrite the formulation $k$ in Eq.~\ref{eq:appendix:derivation-2} as:

\begin{equation}
\label{eq:appendix:derivation-3}
\sum_{i=1}^{n^{\prime}} \sum_{m\in\mathcal{V}_{i}} \frac{\theta_{ik} \mathrm{diag}(\boldsymbol{\delta}_{m})}{|\mathcal{V}_{i}|} = \mathrm{diag}(\boldsymbol{C}^{+}\boldsymbol{\Theta_{:k+1}})\ .
\end{equation}

This directly leads to Eq.~\ref{eq:partion-filtering-unified}, thus completing the derivation.


\section{Proof of Proposition~\ref{proposition:unifying}}
\label{appendix:proposition:unifying}

Proposition~\ref{proposition:unifying}: \textit{
Recall that the coarsening ratio $r$ spans the range $\left[0,\frac{n-1}{n}\right]$.
When the coarsening ratio $r=\frac{n-1}{n}$, Eq.~\ref{eq:partion-filtering-unified} performs graph-wise filtering; when $r=0$, Eq.~\ref{eq:partion-filtering-unified} performs node-wise filtering.
}
\begin{proof}
\label{proof:proposition:unifying}
The proof proceeds in two straightforward parts:

\subsubsection*{\bf Eq.~\ref{eq:partion-filtering-unified} performs graph-wise filtering} When setting $r=\frac{n-1}{n}$, the coarsening algorithm results in a single partition $n^{\prime}=1$, i.e., the entire set of nodes in $\mathcal{G}$ is grouped into a single cluster $\mathcal{V}=\mathcal{V}_{\mathcal{G}}$. 
As a result, Eq.~\ref{eq:partion-filtering-unified} applies the filtering operation on this single partition using one filter, producing the output $\boldsymbol{Z}$, which directly corresponds to the node embeddings for the full graph, making Eq.~\ref{eq:partion-filtering-unified} equivalent to graph-wise filtering.

\subsubsection*{\bf Eq.~\ref{eq:partion-filtering-unified} performs node-wise filtering} When setting $r=0$, the coarsening algorithm produces $n^{\prime}=n$ disjoint node partitions, denoted as $\mathcal{V}=\{\mathcal{V}_{1},\mathcal{V}_{2},...,\mathcal{V}_{n}\}$, where each $\mathcal{V}_{i}$, $i=1,2,...,n$, contains exactly one unique node from $\mathcal{G}$. 
Hence, the filtering operation becomes a collection of $n$ independent filters, where each corresponding to a specific node, making Eq.~\ref{eq:partion-filtering-unified} equivalent to node-wise filtering.
\end{proof}

Furthermore, Notice that the coarsening ratio $r$ varies within the range $\left[0,\frac{n-1}{n}\right]$. 
In this context, graph-wise filtering at $r=\frac{n-1}{n}$ and node-wise filtering at $r=0$ represent the two extremes of the Eq.~\ref{eq:partion-filtering-unified}.


\section{Proof of Proposition~\ref{proposition:feature-adjustment}}
\label{appendix:proposition:feature-adjustment}

Proposition~\ref{proposition:feature-adjustment}: \textit{
Let $\boldsymbol{z}_{1}, \boldsymbol{z}_{2}, \boldsymbol{z}_{3} \in \mathbb{R}^c$ and define $d_{12} = \|\boldsymbol{z}_{1} - \boldsymbol{z}_{2}\|$ and $d_{13} = \|\boldsymbol{z}_{1} - \boldsymbol{z}_{3}\|$. Assume that $d_{12} < d_{13}$. Then:
\begin{enumerate}[leftmargin=25pt,parsep=2pt,itemsep=2pt,topsep=2pt]
    \item There exists $\boldsymbol{W} \in \mathbb{R}^{c \times c}$, such that $\|\boldsymbol{W}\boldsymbol{z}_{1} - \boldsymbol{W}\boldsymbol{z}_{2}\| > \|\boldsymbol{W}\boldsymbol{z}_{1} - \boldsymbol{W}\boldsymbol{z}_{3}\|$
    \item There exists $\boldsymbol{W}_1$, $\boldsymbol{W}_2 \in \mathbb{R}^{c \times c}$, such that $\|\boldsymbol{W}_1\boldsymbol{z}_{1} - \boldsymbol{W}_1\boldsymbol{z}_{2}\| > \|\boldsymbol{W}_1\boldsymbol{z}_{1} - \boldsymbol{W}_2\boldsymbol{z}_{3}\|$
\end{enumerate}
}
\begin{proof}
\label{proof:proposition:feature-adjustment}
We prove each part separately.

\subsubsection*{\bf For statement (1)} 
Since $d_{12} < d_{13}$, define
$\boldsymbol{u} = \boldsymbol{z}_{1} - \boldsymbol{z}_{2}$ and $\boldsymbol{v} = \boldsymbol{z}_{1} - \boldsymbol{z}_{3}$.  
Then
$\|\boldsymbol{u}\| = d_{12}$ and $\|\boldsymbol{v}\| = d_{13}$, so that
$\|\boldsymbol{u}\| < \|\boldsymbol{v}\|$. 
The idea is to define a linear transformation that scales the direction $\boldsymbol{u}$ differently from the direction $\boldsymbol{v}$. Choose positive real numbers $\alpha$ and $\beta$ such that
$\alpha \|\boldsymbol{u}\| > \beta \|\boldsymbol{v}\|$.  
For example, one may take $\alpha = 1$ and choose any $\beta$ satisfying
$0 < \beta < \frac{\|\boldsymbol{u}\|}{\|\boldsymbol{v}\|}$. 
Since $\boldsymbol{u}$ and $\boldsymbol{v}$ are vectors in $\mathbb{R}^c$, extend the set $\{\boldsymbol{u}, \boldsymbol{v}\}$ to a basis of $\mathbb{R}^c$; let this basis be
$\{\boldsymbol{u}, \boldsymbol{v}, \boldsymbol{w}_3, \boldsymbol{w}_4, \dots, \boldsymbol{w}_c\}$.  
Define the linear transformation $W : \mathbb{R}^c \to \mathbb{R}^c$ by specifying its action on the basis elements as follows:
$\boldsymbol{W}(\boldsymbol{u}) = \alpha\, \boldsymbol{u}$, $\boldsymbol{W}(\boldsymbol{v}) = \beta\, \boldsymbol{v}$, and $\boldsymbol{W}(\boldsymbol{w}_j) = \boldsymbol{w}_j$ for $j = 3, \dots, c$.  
By linearity, this uniquely defines a $c \times c$ matrix $W$.

Notice that
$\boldsymbol{W}(\boldsymbol{z}_{1} - \boldsymbol{z}_{2}) = \boldsymbol{W}(\boldsymbol{u}) = \alpha\, \boldsymbol{u}$, and $\boldsymbol{W}(\boldsymbol{z}_{1} - \boldsymbol{z}_{3}) = \boldsymbol{W}(\boldsymbol{v}) = \beta\, \boldsymbol{v}$.  
Taking norms, we obtain:
$\|W\boldsymbol{z}_{1} - W\boldsymbol{z}_{2}\| = \|\alpha\, \boldsymbol{u}\| = \alpha \|\boldsymbol{u}\|$, and  
$\|W\boldsymbol{z}_{1} - W\boldsymbol{z}_{3}\| = \|\beta\, \boldsymbol{v}\| = \beta \|\boldsymbol{v}\|$.  
By our choice of $\alpha$ and $\beta$, we have
$\alpha \|\boldsymbol{u}\| > \beta \|\boldsymbol{v}\|$, so that
$\|W\boldsymbol{z}_{1} - W\boldsymbol{z}_{2}\| > \|W\boldsymbol{z}_{1} - W\boldsymbol{z}_{3}\|$.

\subsubsection*{\bf For statement (2)} 
We wish to find matrices $\boldsymbol{W}_1$ and $\boldsymbol{W}_2$ such that
$\|\boldsymbol{W}_1\boldsymbol{z}_{1} - \boldsymbol{W}_1\boldsymbol{z}_{2}\| > \|\boldsymbol{W}_1\boldsymbol{z}_{1} - \boldsymbol{W}_2\boldsymbol{z}_{3}\|$.  
A simple approach is to restrict the second distance to be very small. For instance, choose
$\boldsymbol{W}_1 = I$,  
the $c \times c$ identity matrix. Then,
$\|\boldsymbol{W}_1\boldsymbol{z}_{1} - \boldsymbol{W}_1\boldsymbol{z}_{2}\| = \|\boldsymbol{z}_{1} - \boldsymbol{z}_{2}\| = d_{12}$.

Now, we need to choose a matrix $\boldsymbol{W}_2$ such that
$\|\boldsymbol{z}_{1} - \boldsymbol{W}_2\boldsymbol{z}_{3}\| < d_{12}$. 
Since $\boldsymbol{z}_{1}$ and $\boldsymbol{z}_{3}$ are given, if we choose $\boldsymbol{W}_2$ to satisfy
$\boldsymbol{W}_2 \boldsymbol{z}_{3} = \boldsymbol{z}_{1}$,  
then
$\|\boldsymbol{z}_{1} - \boldsymbol{W}_2\boldsymbol{z}_{3}\| = \|\boldsymbol{z}_{1} - \boldsymbol{z}_{1}\| = 0 < d_{12}$.  
To show such a $c \times c$ matrix $\boldsymbol{W}_2$ exists, note that if $\boldsymbol{z}_{3}$ is nonzero (which must be the case since $d_{13} = \|\boldsymbol{z}_{1} - \boldsymbol{z}_{3}\| > d_{12} \ge 0$), then there are infinitely many linear maps sending $\boldsymbol{z}_{3}$ to $\boldsymbol{z}_{1}$.  
For example, choose a basis $\{\boldsymbol{z}_{3}, \boldsymbol{u}_2, \dots, \boldsymbol{u}_c\}$ of $\mathbb{R}^c$ and define $\boldsymbol{W}_2$ by
$\boldsymbol{W}_2(\boldsymbol{z}_{3}) = \boldsymbol{z}_{1}$, $\boldsymbol{W}_2(\boldsymbol{u}_j) = 0$ for $j = 2, \dots, c$.  
Then extend $\boldsymbol{W}_2$ linearly to all of $\mathbb{R}^c$. 
With this choice of $\boldsymbol{W}_2$, we have
$\|\boldsymbol{W}_1\boldsymbol{z}_{1} - \boldsymbol{W}_2\boldsymbol{z}_{3}\| = \|\boldsymbol{z}_{1} - \boldsymbol{z}_{1}\| = 0$,  
and hence,
$\|\boldsymbol{W}_1\boldsymbol{z}_{1} - \boldsymbol{W}_1\boldsymbol{z}_{2}\| = d_{12} > 0 = \|\boldsymbol{W}_1\boldsymbol{z}_{1} - \boldsymbol{W}_2\boldsymbol{z}_{3}\|$.

\subsubsection*{\bf Summary} Based on the preceding discussion, we have constructed matrices $\boldsymbol{W}_1$ and $\boldsymbol{W}_2$ satisfying the required inequality. 
Moreover, this result can be generalized to a collection of linear transformations $\{\boldsymbol{W}_1, \boldsymbol{W}_2, ..., \boldsymbol{W}_c\}$, allowing us to demonstrate the desired correction over the entire matrix $\boldsymbol{Z}$.
\end{proof}


\section{Experimental Details}
\label{appendix:exp-details}

This section elaborates on the comprehensive details of the experiments outlined in Sections~\ref{section-exp-node-classify} and~\ref{section-exp-graph-anomaly-detection}. 
Experiments are carried out on an NVIDIA Tesla V100 (32GB) GPU, utilizing an Ubuntu 20.04 OS and CUDA version 11.8.

\subsection{Experimental details for node classification}
\label{appendix:exp-details-node-classify}

\subsubsection*{\bf Dataset statistics} The statistics of the $13$ datasets used in Section~\ref{section-exp-node-classify} are provided in Tables~\ref{table-datasets-statistics-medium-to-large} and~\ref{table-datasets-statistics-exlarge}. 

\begin{table*}[!ht]
  \caption{Statistics for medium-to-large datasets, with \# $\mathcal{H}$ indicating the edge homophily measure as defined in~\cite{H2GCN}.} 
  \vskip -0.05in
  \label{table-datasets-statistics-medium-to-large}
  \centering
  \begin{tabular}{lccccccccc}
  \hline
    & Cora & CiteSeer & PubMed & Ogbn-arxiv & Roman-empire & Amazon-ratings & Questions & Gamers & Genius \\ \hline
    \# Nodes & 2708 & 3327 & 19,717 &  169,343  &  22,662  &  24,492  &  48,921 & 168,114  &  \textbf{421,961}  \\
    \# Edges &  5278 & 4552 & 44,324 & 1,157,799 & 32,927 & 93,050 & 153,540 & \textbf{6,797,557} & 922,868 \\
    \# Features & 1433 & 3703 & 500 & 128 & 300 & 300 & 301 & 7 & 12 \\
    \# Classes & 7 & 6 & 5 & 40 & 18 & 5 & 2 & 2 & 2 \\
    \# $\mathcal{H}$~\cite{H2GCN} &  0.81 &  0.74 &  0.80 &  0.65 &  0.05 &  0.38 & 0.84 &  0.55 &  0.62 \\
    \hline
  \end{tabular}
\end{table*}
\begin{table}[!ht]
  \caption{Statistics for exceptionally large datasets. 
  \# $\mathcal{H}$ for Ogbn-papers100M is unavailable due to runtime exceedance.} 
  \vskip -0.05in
  \label{table-datasets-statistics-exlarge}
  \centering
  \setlength{\tabcolsep}{3pt}
  \renewcommand\arraystretch{1.05}
  \begin{tabular}{lcccc}
  \hline
    & Products & Papers100M & Snap-patents & Pokec \\ \hline
    \# Nodes &  2,449,029 & \textbf{111,059,956} & 2,923,922 &  1,632,803 \\
    \# Edges &   61,859,140 & \textbf{1,615,685,872} & 13,975,788 & 30,622,564 \\
    \# Features & 100 & 128 & 269 & 65 \\
    \# Classes & 47 & 172 & 5 & 2 \\
    \# $\mathcal{H}$ &  0.81 &  - & 0.07 &  0.45  \\
    \hline
  \end{tabular}
\end{table}

\subsubsection*{\bf Baseline implementations} We provide GitHub links to official implementations for all baseline models referenced in this paper, including additional baselines evaluated in Section~\ref{appendix:additional-results-node-classification}. 

For widely adopted baselines GCN, ChebNet, ARMA and APPNP, we adopt consistent implementations drawn from prior research~\cite{BernNet-GNN-narrowbandresults-1,ChebNetII,chebnet2d,OptBasisGNN,JacobiConv,decoupled-FEGNN,decoupled-AdaptKry,decoupled-NFGNN-nodewise}. 
For the remaining baselines, we resort to the officially released code, accessible via the provided URLs as below.

\begin{itemize}[leftmargin=25pt,parsep=2pt,itemsep=2pt,topsep=2pt]
\item H2GCN: \url{https://github.com/GemsLab/H2GCN}
\item GloGNN: \url{https://github.com/RecklessRonan/GloGNN}
\item Nodeformer: \url{https://github.com/qitianwu/NodeFormer}
\item PDEGCN: \url{https://openreview.net/forum?id=wWtk6GxJB2x}
\item GBKGNN: \url{https://github.com/Xzh0u/GBK-GNN}
\item LINKX: \url{https://github.com/CUAI/Non-Homophily-Large-Scale}
\item OrderGNN: \url{https://github.com/lumia-group/orderedgnn}
\item LRGNN: \url{https://github.com/Jinx-byebye/LRGNN}
\item GCN: \url{https://github.com/ivam-he/ChebNetII}
\item SGC: \url{https://github.com/ivam-he/ChebNetII}
\item GCNII: \url{https://github.com/chennnM/GCNII}
\item ChebNet: \url{https://github.com/ivam-he/ChebNetII}
\item ARMA: \url{https://github.com/ivam-he/ChebNetII}
\item ACMGCN: \url{https://github.com/SitaoLuan/ACM-GNN}
\item Specformer: \url{https://github.com/DSL-Lab/Specformer}
\item FLODE: \url{https://github.com/rpaolino/flode}
\item APPNP: \url{https://github.com/ivam-he/ChebNetII}
\item GPRGNN: \url{https://github.com/jianhao2016/GPRGNN}
\item BernNet: \url{https://github.com/ivam-he/BernNet}
\item ChebNetII: \url{https://github.com/ivam-he/ChebNetII}
\item OptBasis: \url{https://github.com/yuziGuo/FarOptBasis}
\item NIGCN: \url{https://github.com/kkhuang81/NIGCN}
\item JacobiConv: \url{https://github.com/GraphPKU/JacobiConv}
\item FEGNN: \url{https://github.com/sajqavril/Feature-Extension-Graph-Neural-Networks}
\item AdaptKry: \url{https://github.com/kkhuang81/AdaptKry}
\item PCNet: \url{https://github.com/uestclbh/PC-Conv}
\item UniFilter: \url{https://github.com/kkhuang81/UniFilter}
\item NFGNN: \url{https://github.com/SsGood/NFGNN}
\end{itemize}

\subsubsection*{\bf Implementation of CPF} As introduced in Section~\ref{section-CPF-decoupled-gnn}, CPF is implemented in two distinct configurations to accommodate graphs of varying sizes. 
For medium-to-large graphs detailed in Table~\ref{table-node-classify-medium}, we utilize the architecture represented by Eq.~\eqref{eq:decoupled-CPF-medium}. 
For exceptionally large graphs listed in Table~\ref{table-node-classify-large}, we employ the architecture shown in Eq.~\eqref{eq:decoupled-CPF-large}.

The MLP architecture within CPF is dataset-specific. 
For medium-sized graphs (Cora, Citeseer, Pubmed, Roman-empire, Amazon-ratings, and Questions), we use a two-layer MLP with $64$ hidden units. 
In contrast, larger datasets are assigned three-layer MLPs with varying hidden units: $128$ for Gamers and Genius, $256$ for Snap-patents and Pokec, $512$ for Ogbn-arxiv, and $1024$ for Ogbn-papers100M.

For all main experiments, we implement $\mathbf{T}_{k}$ using the Chebyshev basis~\cite{poly_chebyshev}, setting the maximum polynomial degree to $K=10$ and the coarsening ratio to $r=0.5$.

The remaining hyperparameters are tuned via grid search, optimizing weight decay over $\{5e-2,5e-4,0\}$, dropout over $\{0,0.5\}$, and learning rate over $\{0.5,0.1,0.05,0.01,0.005,0.001\}$.

\subsubsection*{\bf Model training and testing} We adhere to the dataset split protocols prevalent in the literature. 
Specifically, for Cora, Citeseer, and Pubmed, we utilize the well-established $60\%/20\%/20\%$ train/val/test split, as detailed in various studies~\cite{BernNet-GNN-narrowbandresults-1,ChebNetII,chebnet2d,JacobiConv,OptBasisGNN,decoupled-FEGNN,decoupled-PCConv,decoupled-UniFilter,decoupled-AdaptKry}. 
For the Roman-empire, Amazon-ratings, and Questions datasets, we implement a $50\%/25\%/25\%$ train/val/test split, consistent with the original dataset protocols~\cite{dataset8-small-hetero}. 
The same $50\%/25\%/25\%$ train/val/test split is applied to the Gamers, Genius, Snap-patents, and Pokec datasets, as specified in~\cite{dataset6-large-hetero}. 
Lastly, for Ogbn-arxiv, Ogbn-products, and Ogbn-papers100M, we adopt the fixed splits established in the original OGB dataset paper~\cite{dataset5-ogb}.

Models are trained for a maximum of $1,000$ epochs, implementing early stopping after $200$ epochs if validation accuracy shows no improvement. 
A mini-batch training strategy with $20000$ nodes per batch is applied to the experiments on exceptionally large graphs. 
The optimization process employs the Adam optimizer~\cite{Adamoptimizer}. 
For each dataset, we create $10$ random node splits, conducting $10$ random initializations of each baseline on these splits. 
The final results reported for each baseline on a given dataset represent the average from $100$ evaluations.

\subsection{Experimental details for graph anomaly detection}
\label{appendix:exp-details-graph-anomaly-detection}

\subsubsection*{\bf Dataset statistics} Table~\ref{table-datasets-statistics-GAnodet} presents the statistics of datasets used in Section~\ref{section-exp-graph-anomaly-detection}. 

\begin{table}[!ht]
  \caption{Statistics of datasets utilized for graph anomaly detection. 
  \# Anomaly represents the rate of abnormal nodes.}
  \vskip -0.05in
  \label{table-datasets-statistics-GAnodet}
  \centering
  \begin{tabular}{lccc}
  \hline
    & YelpChi & Amazon & T-Finance \\ \hline
    \# Nodes &  45,954  & 11,944 & 39,357  \\
    \# Edges &  3,846,979 & 4,398,392 & 21,222,543  \\
    \# Features & 32 & 25 & 10 \\
    \# Anomaly & 14.53\% & 6.87\% &  4.58\% \\
    \hline
  \end{tabular}
  \vskip -0.05in
\end{table}

\subsubsection*{\bf Baseline implementations} We provide GitHub links to the official implementations of all baseline models referenced in this paper. Specifically, for general-purpose filtering-based GNNs, including GPRGNN, AdaptKry, OptBasis, NIGCN, and NFGNN—each proposed as a uniform decoupled GNN architecture—we implement these baselines following CPF's variant for medium-to-large graphs as shown in Eq.~\eqref{eq:decoupled-CPF-medium}. 
The maximum polynomial degree for each model is set to $10$, and we use a two-layer MLP with $64$ hidden units for feature transformation, aligned with the BWGNN parameter settings~\cite{GAnoDet-spectral-1-BWGNN}. 
The GCN baseline is implemented in a two-layer format with $64$ hidden units to match established standards. 
For other baselines, we rely on their official implementations, which are linked below. 
All models are rebuilt and evaluated in PyG~\cite{PyTorchGeometric} framework to maintain experimental fairness.

\begin{itemize}[leftmargin=25pt,parsep=2pt,itemsep=2pt,topsep=2pt]
\item PC-GNN: \url{https://github.com/PonderLY/PC-GNN}
\item CARE-GNN: \url{https://github.com/YingtongDou/CARE-GNN}
\item GDN: \url{https://github.com/blacksingular/wsdm_GDN}
\item BWGNN: \url{https://github.com/squareRoot3/Rethinking-Anomaly-Detection}
\item GHRN: \url{https://github.com/blacksingular/GHRN}
\item GPRGNN: \url{https://github.com/jianhao2016/GPRGNN}
\item NIGCN: \url{https://github.com/kkhuang81/NIGCN}
\item OptBasis: \url{https://github.com/yuziGuo/FarOptBasis}
\item AdaptKry: \url{https://github.com/kkhuang81/AdaptKry}
\item NFGNN: \url{https://github.com/SsGood/NFGNN}
\end{itemize}

\subsubsection*{\bf Implementation of CPF} Our CPF employs a decoupled structure in line with standard filtering-based GNNs, with a maximum polynomial degree of $K=10$ and a two-layer MLP containing $64$ hidden units for feature transformation. 
Consistent with previous setups, we utilize the Chebyshev basis for $\mathbf{T}_{k}$ and apply a coarsening ratio $r=0.5$, as elaborated in Appendix~\ref{appendix:exp-details-node-classify}.

\subsubsection*{\bf Model training and testing} We follow the training protocol outlined in BWGNN's paper~\cite{GAnoDet-spectral-1-BWGNN}, maintaining a validation-to-test set ratio of $1:2$. 
All baselines are trained over $100$ epochs without early stopping, using the Adam optimizer. 
Test results are reported based on models achieving the highest Macro-F1 score on the validation set, averaged across $10$ random initializations.


\section{Additional Results}
\label{appendix:additional-results}

This section presents additional results that reinforce the experiments discussed in the main text, providing further validation of our conclusions.

\subsection{Additional results of node classification}
\label{appendix:additional-results-node-classification}

We incorporate advanced baselines into our comparisons to validate the effectiveness of our CPF method. 
A summary of the results can be found in Tables~\ref{table-node-classify-medium-full} and~\ref{table-node-classify-large-full}. 

The results presented in both tables demonstrate that CPF consistently outperforms each of the advanced baselines by substantial margins while maintaining comparable efficiency. In contrast, these baselines only show promising performance on a limited subset of the datasets. This further underscores the superior efficacy of CPF over contemporary counterparts.

\begin{table*}[!t]
\caption{Additional node classification results on medium-to-large graphs, with more baselines being included.}
\vskip -0.05in
\label{table-node-classify-medium-full}
\centering
\setlength{\tabcolsep}{4pt}
\resizebox{\linewidth}{!}{
\begin{tabular}{cccccc|ccccc}
\hline
\multirow{2}{*}{\makecell[c]{Model \\ Type}} & \multirow{2}{*}{Method}              & \multicolumn{4}{c|}{homophilic graphs}                                                                                        & \multicolumn{5}{c}{heterophilic graphs}                                                                                                                       \\ \cline{3-11} 
                      &                                      & Cora                          & Cite.                         & Pubmed                        & Arxiv                         & Roman.                        & Amazon.                       & Ques.                         & Gamers                        & Genius                        \\ \hline
\multirow{8}{*}{\large \romannumeral1}    & H2GCN                                & $87.33_{\pm0.6}$              & $75.11_{\pm1.2}$              & $88.39_{\pm0.6}$              & $71.93_{\pm0.4}$              & $61.38_{\pm1.2}$              & $37.17_{\pm0.5}$              & $64.42_{\pm1.3}$              & $64.71_{\pm0.4}$              & $90.12_{\pm0.2}$              \\
                      & GLOGNN                               & $88.12_{\pm0.4}$              & $76.23_{\pm1.4}$              & $88.83_{\pm0.2}$              & $72.08_{\pm0.3}$              & $71.17_{\pm1.2}$              & $42.19_{\pm0.6}$              & $74.42_{\pm1.3}$              & $65.62_{\pm0.3}$              & $90.39_{\pm0.3}$              \\
                      & Nodeformer~\cite{nodeformer}              & $87.44_{\pm0.4}$              & $76.11_{\pm1.1}$              & $87.62_{\pm0.3}$              & $72.31_{\pm0.3}$              & $64.33_{\pm1.3}$              & $38.17_{\pm0.4}$              & $70.22_{\pm1.4}$              & $66.02_{\pm0.3}$              & $88.44_{\pm0.3}$              \\
                      & PDEGCN~\cite{pdegcn}                  & $87.62_{\pm0.5}$              & $75.83_{\pm1.3}$              & $88.43_{\pm0.3}$              & $71.88_{\pm0.4}$              & $60.27_{\pm1.1}$              & $36.41_{\pm0.7}$              & $67.53_{\pm1.1}$              & $62.15_{\pm0.3}$              & $88.71_{\pm0.5}$              \\
                      & GBKGNN~\cite{GBKGNN}                               & $87.38_{\pm0.3}$              & $75.68_{\pm1.2}$              & $88.28_{\pm0.4}$              & $71.76_{\pm0.3}$              & $61.46_{\pm1.4}$              & $35.17_{\pm0.7}$              & $66.28_{\pm0.9}$              & $63.44_{\pm0.4}$              & $88.45_{\pm0.3}$              \\
                      & LINKX                                & $84.51_{\pm0.6}$              & $73.25_{\pm1.5}$              & $86.36_{\pm0.6}$              & $71.14_{\pm0.2}$              & $67.55_{\pm1.2}$              & $41.57_{\pm0.6}$              & $63.85_{\pm0.8}$              & $65.82_{\pm0.4}$              & $\underline{91.12_{\pm0.5}}$  \\
                      & OrderGNN                             & $87.55_{\pm0.2}$              & $75.46_{\pm1.2}$              & $88.31_{\pm0.3}$              & $71.90_{\pm0.5}$              & $71.69_{\pm1.6}$              & $40.93_{\pm0.5}$              & $70.82_{\pm1.0}$              & $66.09_{\pm0.3}$              & $89.45_{\pm0.4}$              \\
                      & LRGNN~\cite{low-rank-gnn-2}                                & $87.48_{\pm0.3}$              & $75.29_{\pm1.0}$              & $88.65_{\pm0.4}$              & $71.69_{\pm0.3}$              & $72.35_{\pm1.4}$              & $42.56_{\pm0.4}$  & $71.82_{\pm1.1}$              & $66.29_{\pm0.5}$              & $90.38_{\pm0.7}$              \\ \hline
\multirow{7}{*}{\large \romannumeral2}   & GCN                                  & $86.48_{\pm0.4}$              & $75.23_{\pm1.0}$              & $87.29_{\pm0.2}$              & $71.77_{\pm0.1}$              & $72.33_{\pm1.6}$              & $42.09_{\pm0.6}$              & $75.17_{\pm0.8}$              & $63.29_{\pm0.5}$              & $86.73_{\pm0.5}$              \\
                      & GCNII                                & $86.77_{\pm0.2}$              & $\underline{76.57_{\pm1.5}}$ & $88.86_{\pm0.4}$              & $71.72_{\pm0.4}$              & $71.62_{\pm1.7}$              & $40.89_{\pm0.4}$              & $72.32_{\pm1.0}$              & $65.11_{\pm0.3}$              & $90.60_{\pm0.6}$              \\
                      & ChebNet                              & $86.83_{\pm0.7}$              & $74.39_{\pm1.3}$              & $85.92_{\pm0.5}$              & $71.52_{\pm0.3}$              & $64.44_{\pm1.5}$              & $38.81_{\pm0.7}$              & $70.42_{\pm1.2}$              & $63.62_{\pm0.4}$              & $87.42_{\pm0.2}$              \\
                      & ARMA~\cite{ARMA-RationalGNN}                    & $86.49_{\pm0.6}$              & $75.43_{\pm1.4}$              & $87.59_{\pm0.1}$              & $71.63_{\pm0.4}$              & $63.17_{\pm1.4}$              & $37.06_{\pm0.5}$              & $69.53_{\pm0.9}$              & $62.61_{\pm0.4}$              & $86.41_{\pm0.8}$              \\
                      & ACMGCN                               & $87.21_{\pm0.4}$              & $76.03_{\pm1.4}$              & $87.37_{\pm0.4}$              & $71.70_{\pm0.3}$              & $66.48_{\pm1.2}$              & $39.53_{\pm0.9}$              & $67.84_{\pm0.5}$              & $64.73_{\pm0.3}$              & $83.45_{\pm0.7}$              \\
                      & Specformer                           & $88.19_{\pm0.6}$              & $75.87_{\pm1.5}$              & $88.74_{\pm0.2}$              & $71.88_{\pm0.2}$              & $71.69_{\pm1.4}$              & $42.06_{\pm0.8}$              & $70.75_{\pm1.2}$              & $65.80_{\pm0.2}$              & $89.39_{\pm0.6}$              \\
                      & FLODE~\cite{FLODE}                                & $87.08_{\pm0.6}$              & $74.52_{\pm1.2}$              & $88.45_{\pm0.3}$              & $71.72_{\pm0.3}$              & $68.32_{\pm1.3}$              & $39.35_{\pm1.1}$              & $67.47_{\pm1.4}$              & $64.56_{\pm0.2}$              & $90.16_{\pm0.6}$              \\
                      \hline
\multirow{10}{*}{\large \romannumeral3}  
                      & APPNP~\cite{decoupled-advantages-3-coupled-disadvantages-2-APPNP}                                & $87.33_{\pm0.7}$              & $75.26_{\pm1.0}$              & $88.49_{\pm0.2}$              & $71.82_{\pm0.2}$              & $62.72_{\pm1.0}$              & $36.91_{\pm1.0}$              & $62.38_{\pm1.4}$              & $65.11_{\pm0.2}$              & $88.73_{\pm0.4}$              \\
                      & GPRGNN                               & $88.26_{\pm0.5}$              & $76.24_{\pm1.2}$              & $88.81_{\pm0.2}$              & $71.89_{\pm0.2}$              & $64.49_{\pm1.6}$              & $41.48_{\pm0.6}$              & $64.58_{\pm1.2}$              & $66.23_{\pm0.1}$              & $90.92_{\pm0.6}$              \\
                      & BernNet~\cite{BernNet-GNN-narrowbandresults-1}                              & $87.57_{\pm0.4}$              & $75.81_{\pm1.8}$              & $88.48_{\pm0.3}$              & $71.72_{\pm0.3}$              & $65.44_{\pm1.4}$              & $40.74_{\pm0.7}$              & $65.53_{\pm1.6}$              & $65.74_{\pm0.3}$              & $89.75_{\pm0.3}$              \\
                      & ChebNetII                            & $88.17_{\pm0.4}$              & $76.41_{\pm1.3}$              & $88.98_{\pm0.4}$              & $72.13_{\pm0.3}$              & $66.77_{\pm1.2}$              & $42.44_{\pm0.9}$              & $71.28_{\pm0.6}$              & $66.44_{\pm0.5}$              & $90.60_{\pm0.2}$              \\
                      & OptBasis                             & $88.35_{\pm0.6}$              & $76.22_{\pm1.4}$              & $89.38_{\pm0.3}$              & $72.10_{\pm0.2}$              & $64.28_{\pm1.8}$              & $41.63_{\pm0.8}$              & $69.60_{\pm1.2}$              & $66.81_{\pm0.4}$  & $90.97_{\pm0.5}$              \\
                      & JacobiConv                           & $\underline{88.53_{\pm0.8}}$  & $76.27_{\pm1.3}$              & $\underline{89.51_{\pm0.2}}$  & $71.87_{\pm0.3}$              & $70.10_{\pm1.7}$              & $42.18_{\pm0.4}$              & $72.16_{\pm1.3}$              & $64.17_{\pm0.3}$              & $89.32_{\pm0.5}$              \\
                      & FEGNN~\cite{decoupled-FEGNN}                                & $87.38_{\pm0.8}$              & $76.08_{\pm1.0}$              & $88.16_{\pm0.5}$              & $71.48_{\pm0.3}$              & $66.43_{\pm1.2}$              & $39.72_{\pm0.5}$              & $69.91_{\pm1.1}$              & $65.12_{\pm0.3}$              & $90.06_{\pm0.4}$              \\
                      & AdaptKry                             & $88.23_{\pm0.7}$              & $76.54_{\pm1.2}$  & $88.38_{\pm0.6}$              & $72.33_{\pm0.3}$              & $71.40_{\pm1.3}$              & $42.31_{\pm1.1}$              & $72.55_{\pm1.0}$              & $66.27_{\pm0.3}$              & $90.55_{\pm0.3}$              \\
                      & PCConv~\cite{decoupled-PCConv}                  & $88.18_{\pm0.6}$              & $75.47_{\pm1.2}$              & $88.69_{\pm0.2}$              & $72.03_{\pm0.2}$              & $68.13_{\pm1.4}$              & $40.06_{\pm1.0}$              & $70.44_{\pm1.1}$              & $65.19_{\pm0.3}$              & $90.19_{\pm0.2}$              \\
                      & UniFilter                            & $88.31_{\pm0.7}$              & $76.38_{\pm1.1}$              & $89.30_{\pm0.4}$              & $\underline{72.87_{\pm0.4}}$  & $71.22_{\pm1.5}$              & $41.37_{\pm0.6}$              & $73.83_{\pm0.8}$              & $65.75_{\pm0.4}$              & $90.66_{\pm0.2}$              \\ \hline
\multirow{3}{*}{\large \romannumeral4}   & NIGCN                      & $88.29_{\pm0.5}$              & $76.38_{\pm1.2}$              & $88.69_{\pm0.3}$              & $72.22_{\pm0.2}$              & $72.18_{\pm1.1}$  & $\underline{42.66_{\pm0.4}}$              & $\underline{75.68_{\pm0.8}}$  & $66.72_{\pm0.5}$              & $90.93_{\pm0.6}$              \\
                      & NODE-MOE$^{\S}$                      & $87.16_{\pm0.8}$              & $75.67_{\pm1.6}$              & $88.54_{\pm0.5}$              & $71.86_{\pm0.5}$              & $72.25_{\pm1.8}$              & $42.33_{\pm1.2}$              & $74.46_{\pm1.2}$              & $66.42_{\pm0.2}$              & $90.39_{\pm0.4}$              \\
                      & NFGNN                                & $88.06_{\pm0.4}$              & $76.22_{\pm1.4}$              & $88.43_{\pm0.4}$              & $72.15_{\pm0.3}$              & $\underline{72.46_{\pm1.2}}$  & $42.19_{\pm0.3}$              & $75.49_{\pm0.9}$  & $\underline{66.85_{\pm0.4}}$              & $90.87_{\pm0.5}$              \\ \hline
\multirow{2}{*}{}     & CPF (\textbf{Ours}) & \textbf{89.28}$\boldsymbol{_{\pm0.7}}$ & \textbf{77.55}$\boldsymbol{_{\pm0.9}}$              & \textbf{89.93}$\boldsymbol{_{\pm0.5}}$ & \textbf{75.46}$\boldsymbol{_{\pm0.3}}$ & \textbf{74.02}$\boldsymbol{_{\pm1.1}}$ & \textbf{46.12}$\boldsymbol{_{\pm0.7}}$ & \textbf{78.80}$\boldsymbol{_{\pm1.1}}$ & \textbf{69.77}$\boldsymbol{_{\pm0.3}}$ & \textbf{92.68}$\boldsymbol{_{\pm0.3}}$ \\
                      & \#Improv.                            & $0.75\%$                      & $0.98\%$                     & $0.42\%$                      & $2.59\%$                      & $1.56\%$                      & $3.46\%$                      & $3.12\%$                      & $2.92\%$                      & $1.56\%$                      \\ \hline
\end{tabular}}
\end{table*}

\subsection{Additional results of ablation studies}
\label{appendix:additional-results-ablation}

We provide a set of detailed ablation studies as supplementary material to enhance the results presented in Section~\ref{section-exp-ablation}. 
These studies offer deeper insights into our CPF methodology. 
The summarized results can be found in Tables~\ref{table-additional-coarsening} and~\ref{table-additional-polynomial-basis}, as well as in Figure~\ref{fig:additional-ablation-r}.

\subsubsection*{\bf Additional results on coarsening ratio $r$} 
\begin{figure*}[!th]
\centering
    \subfloat[Cora.]{
    \label{fig:additional-r-cora}
    \includegraphics[width=0.225\linewidth]{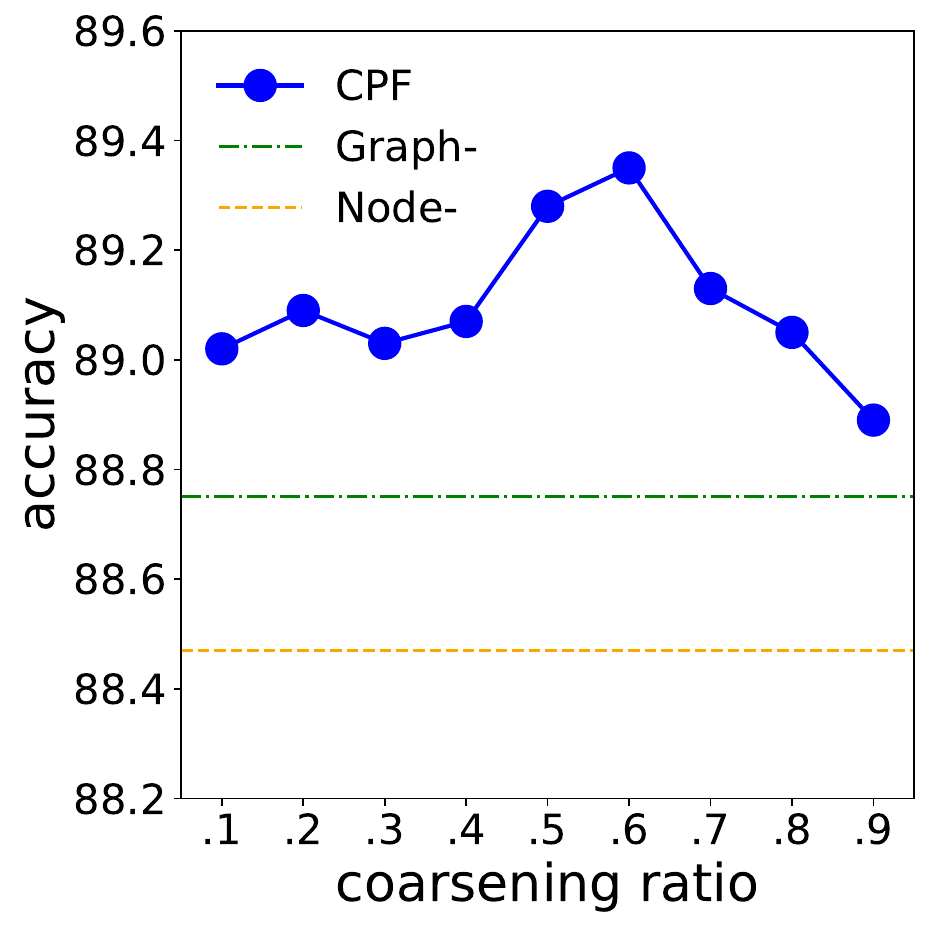}
    }
    \subfloat[Cite.]{
    \label{fig:additional-r-citeseer}
    \includegraphics[width=0.225\linewidth]{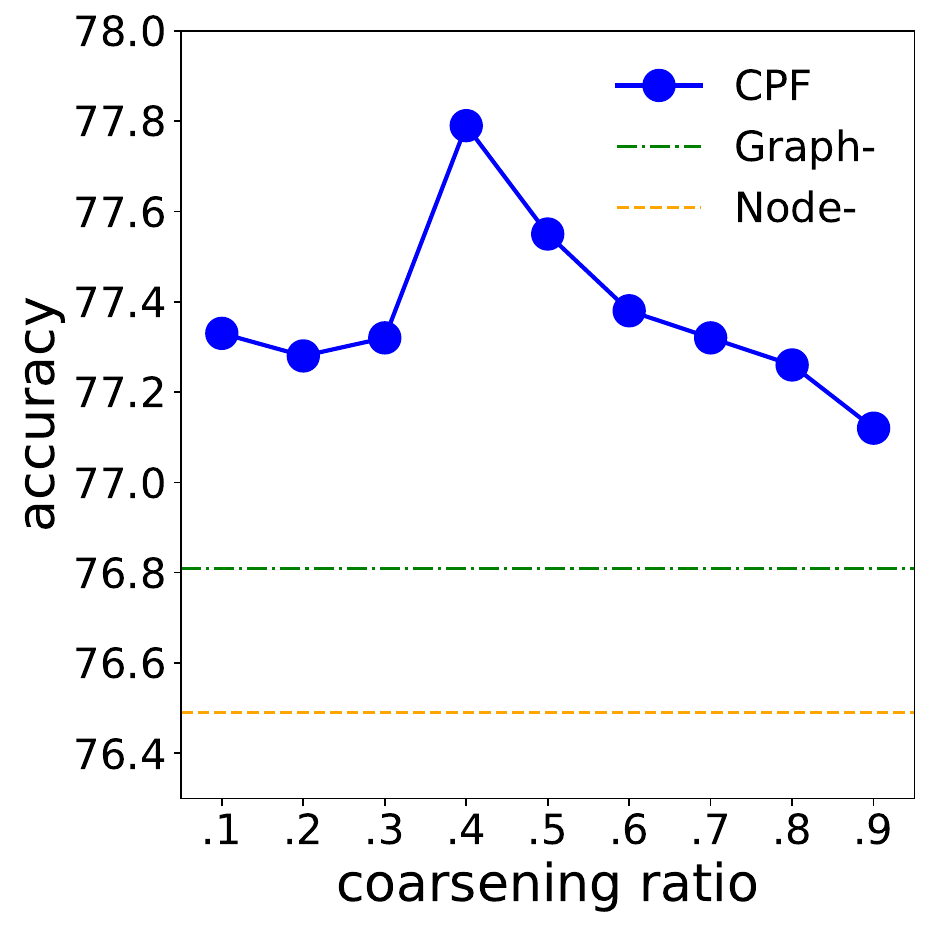}
    }
    \subfloat[Pubmed.]{
    \label{fig:additional-r-pubmed}
    \includegraphics[width=0.225\linewidth]{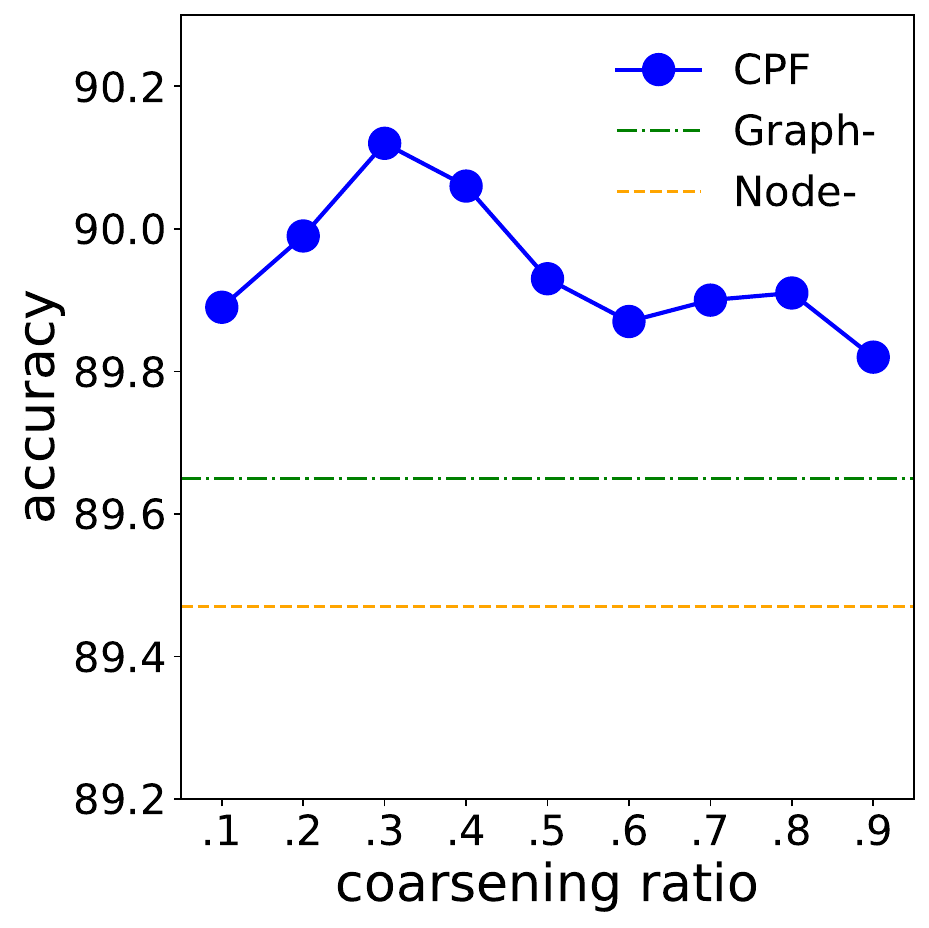}
    }
    \subfloat[Arxiv.]{
    \label{fig:additional-r-arxiv}
    \includegraphics[width=0.225\linewidth]{figure/arxiv.pdf}
    }\\
    \subfloat[Roman.]{
    \label{fig:additional-r-roman}
    \includegraphics[width=0.225\linewidth]{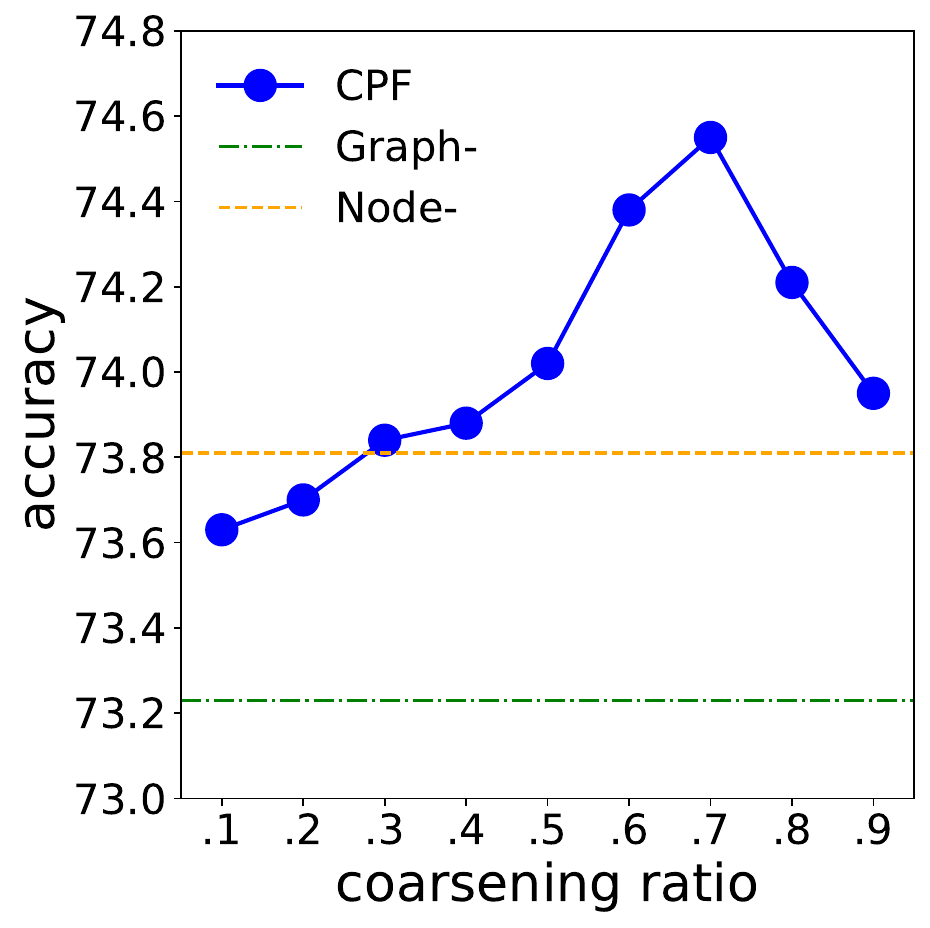}
    }
    \subfloat[Amazon.]{
    \label{fig:additional-r-amazon}
    \includegraphics[width=0.225\linewidth]{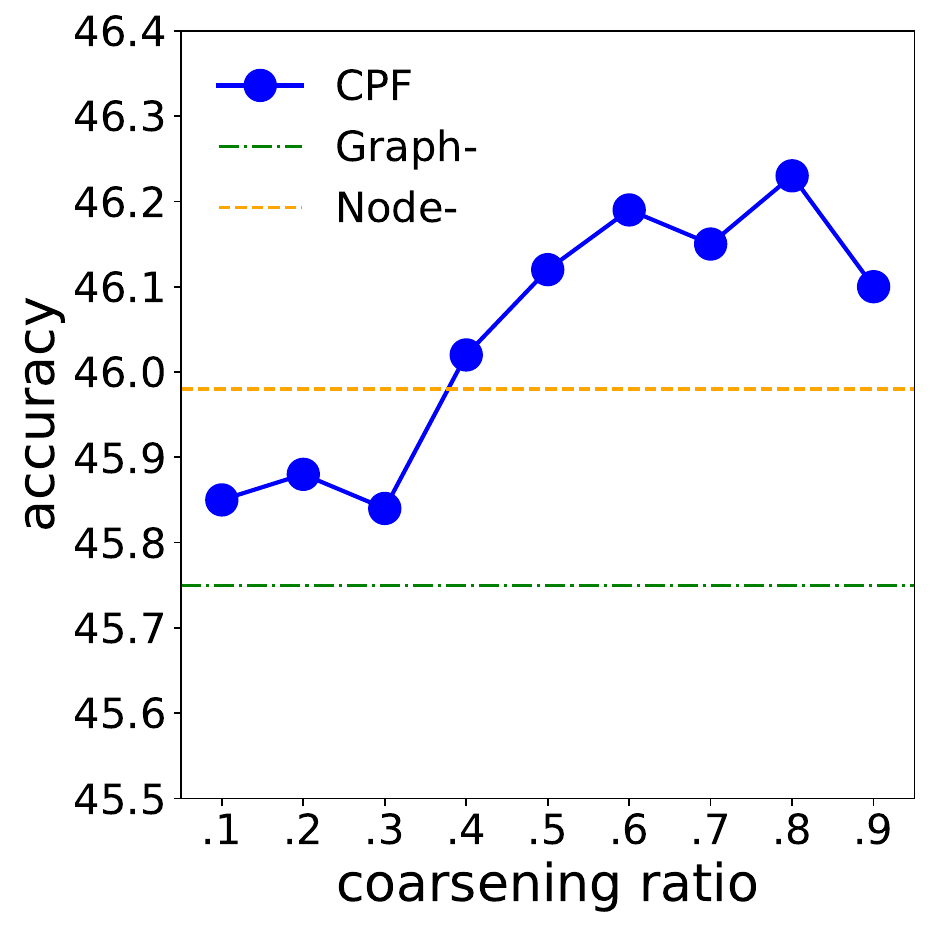}
    }
    \subfloat[Gamers.]{
    \label{fig:additional-r-gamers}
    \includegraphics[width=0.225\linewidth]{figure/gamers.pdf}
    }
    \subfloat[Genius.]{
    \label{fig:additional-r-genius}
    \includegraphics[width=0.225\linewidth]{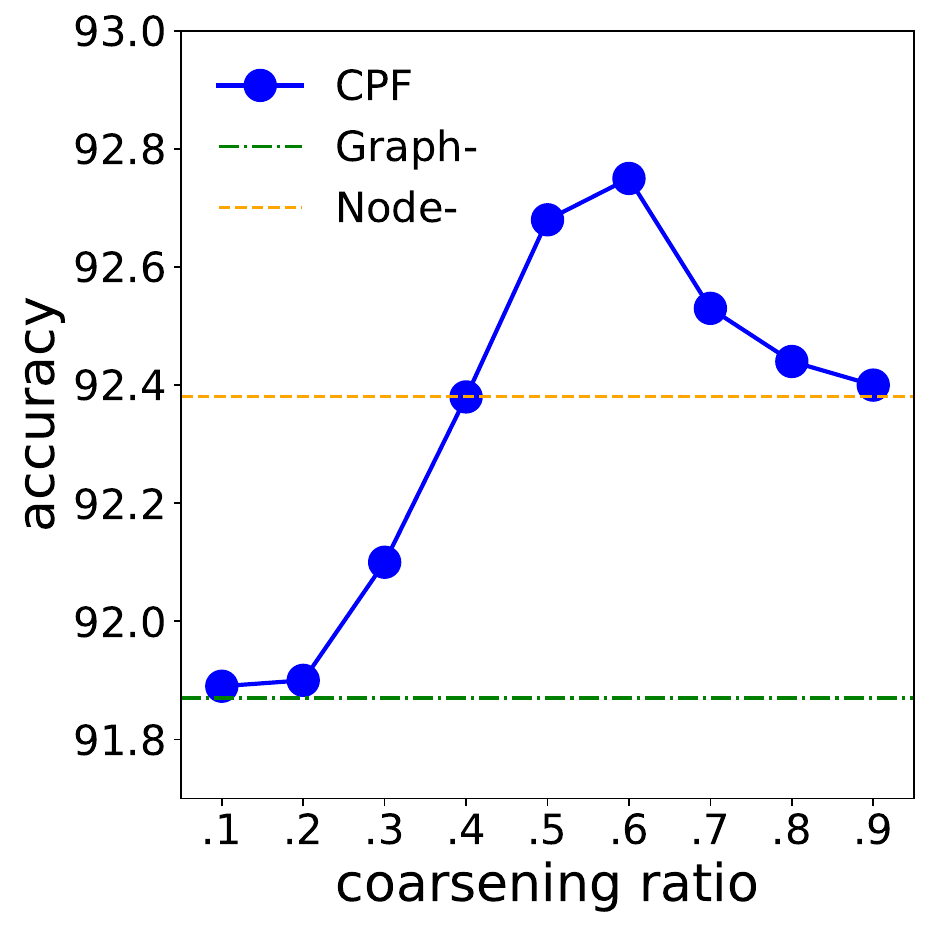}
    }
    \caption{Additional ablation studies of coarsening ratio $r$ evaluated on graphs with diverse sizes and heterophily.}
    \label{fig:additional-ablation-r}
\end{figure*}

Figure~\ref{fig:additional-ablation-r} provides additional ablation results on the coarsening ratio $r$, evaluated across graphs of diverse sizes and heterophily characteristics. 
The results clearly demonstrate that each dataset may require a different optimal value for $r$, with notable performance gains achieved at these optimal selections. 
Generally, this optimum is situated in the middle of the interval $\left(0,\frac{n-1}{n}\right)$, aligned with the partition-wise filtering strategy, whereas the extreme strategies, graph-wise and node-wise, tend to result in comparably poor outcomes. 
This observation corroborates our analysis in Section~\ref{section-CPF}, highlighting that CPF represents a more sophisticated strategy, capable of performing in a "mixture-wise" manner that combines the strengths of both strategies.

\subsubsection*{\bf Additional results for coarsening algorithms} Table~\ref{table-additional-coarsening} illustrates that employing advanced coarsening algorithms leads to consistent performance improvements in CPF. 
Among these, the FGC algorithm stands out as the only method that incorporates node features beyond structural elements, achieving the best overall results.

\subsubsection*{\bf Additional results on polynomial bases} Table~\ref{table-additional-polynomial-basis} includes additional datasets to assess the effects of various polynomial bases. 
The results indicate that the orthogonal polynomial bases, specifically Chebyshev and Jacobian, yield superior performance in CPF. 
These findings align with previous research on the integration of diverse bases in filtering-based GNNs, suggesting that the use of orthogonal bases facilitates improved convergence during GNN training~\cite{ChebNetII,JacobiConv,OptBasisGNN}.

\begin{table*}[!t]
\caption{Additional node classification and runtime results on exceptionally large graphs, with more baselines being included.}
\vskip -0.05in
\label{table-node-classify-large-full}
\centering
\begin{tabular}{ccccccccc}
\hline
\multirow{2}{*}{Method} & \multicolumn{2}{c}{Products}            & \multicolumn{2}{c}{Papers100M}          & \multicolumn{2}{c}{Snap}                & \multicolumn{2}{c}{Pokec}               \\ \cline{2-9} 
                        & Test acc                      & Runtime & Test acc                      & Runtime & Test acc                      & Runtime & Test acc                      & Runtime \\ \hline
GCN                     & $76.37_{\pm0.2}$              & $12.2$   & OOM                           & -       & $46.66_{\pm0.1}$              & $19.8$   & $74.78_{\pm0.2}$              & $15.3$   \\
SGC~\cite{SGC}                     & $75.16_{\pm0.2}$              & $9.1$   & $64.02_{\pm0.2}$              & $104.3$  & $31.11_{\pm0.2}$              & $16.2$   & $60.29_{\pm0.1}$              & $11.8$   \\
APPNP~\cite{decoupled-advantages-3-coupled-disadvantages-2-APPNP}                   & $76.83_{\pm0.3}$              & $12.7$   & $62.88_{\pm0.1}$              & $112.2$  & $34.26_{\pm0.2}$              & $18.9$   & $62.53_{\pm0.1}$              & $15.4$   \\
GPRGNN                  & $79.45_{\pm0.1}$              & $13.5$   & $66.13_{\pm0.2}$              & $113.5$  & $48.88_{\pm0.2}$              & $22.7$   & $79.55_{\pm0.3}$              & $16.7$   \\
BernNet~\cite{BernNet-GNN-narrowbandresults-1}                 & $79.82_{\pm0.2}$              & $14.8$   & $66.08_{\pm0.2}$              & $119.8$  & $47.48_{\pm0.3}$              & $23.1$   & $80.55_{\pm0.2}$              & $17.2$   \\
ChebNetII               & $81.66_{\pm0.3}$              & $13.7$   & $\underline{67.11_{\pm0.2}}$  & $118.8$  & $51.74_{\pm0.2}$              & $23.0$   & $81.88_{\pm0.3}$              & $16.8$   \\
JacobiConv~\cite{JacobiConv}              & $79.35_{\pm0.2}$              & $10.7$   & $65.45_{\pm0.2}$              & $108.8$  & $50.66_{\pm0.2}$              & $17.6$   & $73.83_{\pm0.2}$              & $12.9$   \\
FEGNN~\cite{decoupled-FEGNN}                   & $78.77_{\pm0.3}$              & $14.2$   & $65.19_{\pm0.3}$              & $115.3$  & $46.87_{\pm0.1}$              & $21.1$   & $68.67_{\pm0.3}$              & $16.8$   \\
OptBasis                & $81.33_{\pm0.2}$              & $13.8$   & $67.03_{\pm0.3}$              & $121.7$  & $53.55_{\pm0.1}$              & $22.8$   & $82.09_{\pm0.3}$              & $16.8$   \\
AdaptKry                & $\underline{81.70_{\pm0.3}}$  & $14.5$   & $67.07_{\pm0.2}$              & $122.5$  & $55.92_{\pm0.2}$              & $23.6$   & $82.16_{\pm0.2}$              & $17.8$   \\
PCConv~\cite{decoupled-PCConv}                  & $80.47_{\pm0.2}$              & $14.2$   & $66.19_{\pm0.2}$              & $114.3$  & $51.77_{\pm0.3}$              & $22.5$   & $80.69_{\pm0.2}$              & $16.9$   \\
UniFilter               & $80.33_{\pm0.2}$              & $14.2$   & $66.79_{\pm0.3}$              & $117.8$  & $52.06_{\pm0.1}$              & $23.1$   & $82.23_{\pm0.3}$  & $16.6$   \\ \hline
NIGCN                   & $80.62_{\pm0.2}$              & $13.7$   & $66.57_{\pm0.2}$              & $118.5$  & $57.26_{\pm0.3}$  & $22.9$   & $\underline{82.33_{\pm0.4}}$              & $16.4$   \\
NODE-MOE$^{\S}$         & $80.57_{\pm0.3}$              & $15.1$   & $66.12_{\pm0.2}$              & $125.6$  & $56.36_{\pm0.4}$  & $25.7$   & $81.72_{\pm0.3}$              & $18.3$   \\
NFGNN                   & $81.11_{\pm0.2}$              & $14.1$   & $66.38_{\pm0.2}$              & $122.7$  & $\underline{57.83_{\pm0.3}}$  & $22.9$   & $82.16_{\pm0.3}$              & $17.1$   \\ \hline
CPF (\textbf{Ours})           & \textbf{83.87}$\boldsymbol{_{\pm0.2}}$ & $14.2 (0.9)$   & \textbf{68.87}$\boldsymbol{_{\pm0.3}}$ & $122.7 (7.3)$  & \textbf{64.70}$\boldsymbol{_{\pm0.3}}$ & $23.9 (1.8)$   & \textbf{85.95}$\boldsymbol{_{\pm0.2}}$ & $17.4 (1.3)$   \\
\#Improv.               & $2.17\%$   &   -        & $1.76\%$     &    -     & $6.87\%$    &    -      & $3.62\%$     &     -    \\ \hline
\end{tabular}
\end{table*}
\begin{table*}[!t]
\caption{Additional ablation studies of coarsening algorithms.}
\vskip -0.05in
\label{table-additional-coarsening}
\centering
\renewcommand\arraystretch{1.05}
\begin{tabular}{cccccccccc}
\hline
Dataset                            & Cora                          & Cite.                         & Pubmed                        & Arxiv                         & Roman.                        & Amazon.                       & Ques.                         & Gamers                        & Genius                        \\ \hline
Origin (LV)                        & $89.28_{\pm0.7}$              & $77.55_{\pm0.9}$              & $\underline{89.93_{\pm0.5}}$  & $\underline{75.46_{\pm0.3}}$  & $74.02_{\pm1.1}$              & $46.12_{\pm0.7}$              & $78.80_{\pm1.1}$              & \textbf{69.77}$\boldsymbol{_{\pm0.3}}$ & $\underline{92.68_{\pm0.3}}$  \\ \hline
MGC                                & $\underline{89.78_{\pm0.4}}$  & \textbf{78.78}$\boldsymbol{_{\pm1.3}}$ & $90.11_{\pm0.5}$              & $75.22_{\pm0.2}$              & $\underline{74.76_{\pm1.2}}$  & $\underline{47.52_{\pm0.5}}$  & $78.39_{\pm1.1}$              & $68.12_{\pm0.3}$              & $92.37_{\pm0.3}$              \\
SGC                                & \textbf{90.12}$\boldsymbol{_{\pm0.3}}$ & $78.01_{\pm1.5}$              & \textbf{90.88}$\boldsymbol{_{\pm0.3}}$ & $75.04_{\pm0.2}$              & $74.53_{\pm1.5}$              & $47.17_{\pm0.5}$              & \textbf{80.38}$\boldsymbol{_{\pm1.1}}$ & $68.66_{\pm0.3}$              & $\underline{92.60_{\pm0.2}}$  \\
FGC                                & $89.22_{\pm0.4}$              & $\underline{78.56_{\pm1.1}}$  & $90.26_{\pm0.2}$              & \textbf{76.47}$\boldsymbol{_{\pm0.2}}$ & \textbf{77.81}$\boldsymbol{_{\pm1.3}}$ & \textbf{53.28}$\boldsymbol{_{\pm0.4}}$ & $\underline{79.52_{\pm1.1}}$  & $\underline{69.28_{\pm0.4}}$  & \textbf{93.16}$\boldsymbol{_{\pm0.3}}$ \\ \hline
\end{tabular}
\end{table*}
\begin{table*}[!t]
\caption{Additional ablation studies of polynomial bases, with more datasets being included.}
\vskip -0.05in
\label{table-additional-polynomial-basis}
\centering
\begin{tabular}{cccccccccc}
\hline
Polynomial basis         & Cora                          & Cite.                         & Pubmed                        & Arxiv                         & Roman.                        & Amazon.                       & Ques.                         & Gamers                        & Genius                        \\ \hline
Origin (Chebyshev) & $\underline{89.28_{\pm0.7}}$  & $77.55_{\pm0.9}$              & $\underline{89.93_{\pm0.5}}$  & \textbf{75.46}$\boldsymbol{_{\pm0.3}}$ & $\underline{74.02_{\pm1.1}}$  & $\underline{46.12_{\pm0.7}}$  & \textbf{78.80}$\boldsymbol{_{\pm1.1}}$ & \textbf{69.77}$\boldsymbol{_{\pm0.3}}$ & \textbf{92.68}$\boldsymbol{_{\pm0.3}}$ \\ \hline
Monomial          & $88.79_{\pm0.4}$              & \textbf{78.65}$\boldsymbol{_{\pm1.3}}$ &  $89.33_{\pm0.2}$            & $74.28_{\pm0.4}$              & $72.22_{\pm1.1}$              & $45.38_{\pm0.6}$              & $76.36_{\pm1.1}$              & $68.10_{\pm0.4}$              & $90.73_{\pm0.3}$              \\
Bernstein          & $88.82_{\pm0.3}$              & $\underline{78.17_{\pm1.3}}$  & $89.51_{\pm0.2}$          & $74.65_{\pm0.2}$              & $72.45_{\pm1.3}$              & $45.55_{\pm0.6}$              & $76.02_{\pm1.0}$              & $67.63_{\pm0.4}$              & $91.23_{\pm0.3}$              \\
Jacobian        & \textbf{89.77}$\boldsymbol{_{\pm0.3}}$ & $77.95_{\pm1.1}$              &  \textbf{90.39}$\boldsymbol{_{\pm0.3}}$  & $\underline{75.10_{\pm0.2}}$  & \textbf{74.57}$\boldsymbol{_{\pm1.0}}$ & \textbf{47.71}$\boldsymbol{_{\pm0.4}}$ & $\underline{78.33_{\pm1.0}}$  & $\underline{69.19_{\pm0.3}}$  & $\underline{92.30_{\pm0.2}}$  \\ \hline
\end{tabular}
\end{table*}


\end{document}